\documentclass[11pt]{article}

\usepackage[final]{acl}

\newif\ifshowrev

\newcommand{\rev}[1]{%
  \ifshowrev
    \textcolor{blue}{#1}%
  \else
    #1%
  \fi
}

\newcommand{\revbg}[1]{%
  \ifshowrev
    #1 
  \else
    #1%
  \fi
}

\newif\iflink
\linktrue

\usepackage{times}
\usepackage{latexsym}
\usepackage{booktabs}
\usepackage{hyperref}
\usepackage{diagbox}

\usepackage[T1]{fontenc}

\usepackage[utf8]{inputenc}

\usepackage{microtype}

\usepackage{inconsolata}

\usepackage{graphicx}
\usepackage[nonumberlist, nostyles, nogroupskip]{glossaries}
\glsdisablehyper

\newcommand{\gradiend}{\textsc{Gradiend}}

\newcommand{\bert}{GermanBERT}
\newcommand{\modernbert}{ModernGBERT}
\newcommand{\gbert}{GBERT}
\newcommand{\eurobert}{EuroBERT}
\newcommand{\gpttwo}{GermanGPT-2}
\newcommand{\llama}{LLaMA} 

\newcommand{\cor}{\text{Cor}}
\newcommand{\enc}{enc}
\newcommand{\dec}{dec}

\newcommand{\nominative}{\ensuremath{\textsc{Nom}}}
\newcommand{\accusative}{\ensuremath{\textsc{Acc}}}
\newcommand{\dative}{\ensuremath{\textsc{Dat}}}
\newcommand{\genitive}{\ensuremath{\textsc{Gen}}}

\newcommand{\male}{\ensuremath{\textsc{Masc}}}
\newcommand{\female}{\ensuremath{\textsc{Fem}}}
\newcommand{\neutral}{\ensuremath{\textsc{Neut}}}

\newcommand{\gradiendsymb}{\ensuremath{G}}

\newcommand{\gradcNgMF}{\ensuremath{\gradiendsymb_{\nominative}^{\female,\male}}}

\newcommand{\gradcNDgF}{\ensuremath{\gradiendsymb_{\nominative,\dative}^{\female}}}
\newcommand{\gradcNGgF}{\ensuremath{\gradiendsymb_{\nominative,\genitive}^{\female}}}
\newcommand{\gradcDAgF}{\ensuremath{\gradiendsymb_{\accusative,\dative}^{\female}}}
\newcommand{\gradcADgF}{\ensuremath{\gradiendsymb_{\accusative,\dative}^{\female}}}
\newcommand{\gradcGAgF}{\ensuremath{\gradiendsymb_{\accusative,\genitive}^{\female}}}

\newcommand{\gradcNDgM}{\ensuremath{\gradiendsymb_{\nominative,\dative}^{\male}}}
\newcommand{\gradcDgMF}{\ensuremath{\gradiendsymb_{\dative}^{\female,\male}}}
\newcommand{\gradcDgFN}{\ensuremath{\gradiendsymb_{\dative}^{\female,\neutral}}}
\newcommand{\gradcNGgM}{\ensuremath{\gradiendsymb_{\nominative,\genitive}^{\male}}}

\newcommand{\gradcGgFM}{\ensuremath{\gradiendsymb_{\genitive}^{\female,\male}}}
\newcommand{\gradcGgMF}{\ensuremath{\gradiendsymb_{\genitive}^{\female,\male}}}
\newcommand{\gradcGgFN}{\ensuremath{\gradiendsymb_{\genitive}^{\female,\neutral}}}

\newcommand{\gradcNgFN}{\ensuremath{\gradiendsymb_{\nominative}^{\female,\neutral}}}
\newcommand{\gradcAgFN}{\ensuremath{\gradiendsymb_{\accusative}^{\female,\neutral}}}

\newcommand{\gradcNDgN}{\ensuremath{\gradiendsymb_{\nominative,\dative}^{\neutral}}}
\newcommand{\gradcADgN}{\ensuremath{\gradiendsymb_{\accusative,\dative}^{\neutral}}}
\newcommand{\gradcNGgN}{\ensuremath{\gradiendsymb_{\nominative,\genitive}^{\neutral}}}
\newcommand{\gradcAGgN}{\ensuremath{\gradiendsymb_{\accusative,\genitive}^{\neutral}}}
\newcommand{\gradcGAgN}{\ensuremath{\gradiendsymb_{\accusative,\genitive}^{\neutral}}}

\newcommand{\gradcGDgM}{\ensuremath{\gradiendsymb_{\genitive,\dative}^{\male}}}
\newcommand{\gradcGDgN}{\ensuremath{\gradiendsymb_{\genitive,\dative}^{\neutral}}}

\newcommand{\datasymbol}{\ensuremath{D}}
\newcommand{\dataNM}{\ensuremath{\datasymbol_\nominative^\male}}
\newcommand{\dataNF}{\ensuremath{\datasymbol_\nominative^\female}}
\newcommand{\dataNN}{\ensuremath{\datasymbol_\nominative^\neutral}}
\newcommand{\dataDM}{\ensuremath{\datasymbol_\dative^\male}}
\newcommand{\dataDF}{\ensuremath{\datasymbol_\dative^\female}}
\newcommand{\dataDN}{\ensuremath{\datasymbol_\dative^\neutral}}
\newcommand{\dataAM}{\ensuremath{\datasymbol_\accusative^\male}}
\newcommand{\dataAF}{\ensuremath{\datasymbol_\accusative^\female}}
\newcommand{\dataAN}{\ensuremath{\datasymbol_\accusative^\neutral}}
\newcommand{\dataGM}{\ensuremath{\datasymbol_\genitive^\male}}
\newcommand{\dataGF}{\ensuremath{\datasymbol_\genitive^\female}}
\newcommand{\dataGN}{\ensuremath{\datasymbol_\genitive^\neutral}}

\newcommand{\dataNEUT}{$D_\textsc{Neutral}$}

\newacronym{gradiend}{\gradiend}{GRADIent ENcoder Decoder}

\newacronym{sae}{SAE}{Sparse AutoEncoder}

\newacronym{mlm}{MLM}{Masked Language Modeling}
\newacronym{clm}{CLM}{Causal Language Modeling}
\newacronym{tpt}{TPT}{Token Prediction Task}
\newacronym{lms}{LMS}{Language Modeling Score}

\newacronym{ma}{MA}{Mean Absolute}
\newacronym{mae}{MAE}{Mean Absolute Error}

\newacronym{glue}{GLUE}{General Language Understanding Evaluation}
\newacronym{sglue}{SuperGLUE}{Stickier \acrshort{glue}}
\newacronym{supergleber}{SuperGLEBer}{German Language Understanding Evaluation Benchmark}

\newacronym{bert}{BERT}{Bidirectional Encoder Representations from Transformers}
\newacronym{llama}{\llama}{Large Language Model Meta AI}
\newacronym{ai}{AI}{Artificial Intelligence}
\newacronym{lm}{LM}{Language Model}

\usepackage{tikz}
\usetikzlibrary{shapes.geometric, arrows, positioning, calc, arrows.meta, matrix, tikzmark, fit, calc, backgrounds}
\pgfdeclarelayer{background}
\pgfsetlayers{background,main}
\usepackage{booktabs}  
\usepackage[table]{xcolor}    
\usepackage{siunitx}   
\usepackage[most]{tcolorbox} 
\usepackage[inline,shortlabels]{enumitem}
\usepackage{subcaption}
\usepackage{pifont}
\usepackage{placeins}
\definecolor{darkgreen}{RGB}{0,150,0}  

\newtcolorbox{aclbox}[1][]{%
  enhanced,
  boxrule=0.6pt,
  arc=3pt,
left=3pt,right=3pt,top=3pt,bottom=3pt,
  colback=black!4,
  colframe=black!35,
  fonttitle=\bfseries,
  title=#1
}

\usepackage{mathtools}
\usepackage{amssymb}
\usepackage[normalem]{ulem}  
\usepackage{marginnote}

\usepackage{colortbl}
\usepackage{tabularx}
\usepackage{bm}

\definecolor{artDas}{RGB}{230,240,255}
\definecolor{artDer}{RGB}{255,235,230}
\definecolor{artDie}{RGB}{235,255,235}
\definecolor{artDen}{RGB}{255,245,220}
\definecolor{artDem}{RGB}{240,230,255}
\definecolor{artDes}{RGB}{230,255,255}

\definecolor{edgeAlt}{RGB}{180,130,20}
\definecolor{edgeId}{RGB}{115,60,105}

\definecolor{z1}{RGB}{46,134,193}   
\definecolor{z2}{RGB}{39,174,96}    
\definecolor{idz}{RGB}{120,120,120} 
\definecolor{boxbg}{RGB}{248,248,248}

\newcommand{\die}{\cellcolor{artDie}die}
\newcommand{\der}{\cellcolor{artDer}der}
\newcommand{\das}{\cellcolor{artDas}das}
\newcommand{\den}{\cellcolor{artDen}den}
\newcommand{\dem}{\cellcolor{artDem}dem}
\newcommand{\des}{\cellcolor{artDes}des}

\definecolor{lightgray}{rgb}{0.9, 0.9, 0.9}

\definecolor{custompink}{HTML}{FDD7D6}
\definecolor{customgray}{HTML}{E7FFDD}

\title{Understanding or Memorizing? A Case Study of German Definite Articles in Language Models}

\author{Jonathan Drechsel  \and Erisa Bytyqi \and Steffen Herbold \\
    Faculty of Computer Science and Mathematics \\
    University of Passau \\
    \small{
    \textbf{Correspondence:} \href{mailto:jonathan.drechsel@uni-passau.de}{jonathan.drechsel@uni-passau.de}
  }
}

\begin{document}
\maketitle
\begin{abstract}
Language models perform well on grammatical agreement, but it is unclear whether this reflects rule-based generalization or memorization. We study this question for German definite singular articles, whose forms depend on gender and case. Using \gradiend, a gradient-based interpretability method, we learn parameter update directions for gender-case specific article transitions. 
We find that updates learned for a specific gender-case article transition frequently affect unrelated gender-case settings, with substantial overlap among the most affected neurons across settings.
These results argue against a strictly rule-based encoding of German definite articles, indicating that models at least partly rely on memorized associations rather than abstract grammatical rules.
\end{abstract}

\section{Introduction}

Modern Language Models (\glsunset{lm}\glspl{lm}; \citealt{attention-is-all-you-need}) achieve a near-perfect accuracy on many grammatical phenomena, yet it remains unclear \emph{how} this competence is realized internally~\cite{rogers-etal-2020-primer, belinkovGlass2019, lindsey2025biology}. Do \glspl{lm} 
encode abstract grammatical rules, or do they rely on surface-level memorization of frequent token-context associations? This question is particularly interesting for morphologically rich languages such as German, where grammatical gender, case, and number jointly determine surface forms~\cite{seeker-kuhn-2013-morphological}. Crucially, \textbf{German definite singular articles} are syncretic: 
the same article can appear across multiple genders and cases (e.g., \emph{der} appears as nominative masculine and as dative/genitive feminine;  see Table~\ref{tab:case-gender-table}).
This ambiguity lets us test whether article behavior reflects rule-based generalization or context-specific memorization, framed through two 
hypotheses.

\begin{enumerate}[label=\textbf{H\arabic*}]
    \item \textbf{Memorization hypothesis:} \glspl{lm} memorize surface-level grammatical associations without utilizing the underlying rules. \label{hyp:memorization}
    \item \textbf{Rule-encoding hypothesis:} \glspl{lm} generate text based on internally represented abstract grammatical rules. \label{hyp:ruleencoding}
    \vspace{-0.2em}
\end{enumerate}
To investigate these hypotheses, we apply \gradiend\ \cite{gradiend}, a \rev{simple encoder-decoder} gradient-based interpretability method for feature learning based on parameter update directions for controlled substitutions (here, \rev{gender-case specific} article swaps like $die\to der$ \rev{or $der\to die$}). 
\rev{\gradiend\ learns a one-dimensional latent feature: its encoder maps gradients from opposite swap directions to different scalar values $(\pm 1)$, while neutral inputs are encoded near $0$. This scalar is then used to reconstruct gradients, which can be applied to rewrite the base model along the learned feature direction.}
By learning such features for different gender-case \rev{pairs}, we analyze how grammatical information for article prediction is encoded internally and whether these transition-specific  updates generalize across grammatical settings.
\rev{See Figure~\ref{fig:overview} for an overview.}


\begin{table}
\centering
\small
\begin{tabular}{lcccc}
\toprule
 & \textbf{Nom.} & \textbf{Acc.} & \textbf{Dat.} & \textbf{Gen.} \\
\midrule
\textbf{Male}     & \der & \den & \dem & \des \\
\textbf{Neutral} & \das & \das & \dem & \des \\
\textbf{Female}  & \die & \die & \der & \der \\
\bottomrule
\end{tabular}
\vspace{-0.3em}
\caption{German definite singular articles.}
\vspace{-0.6em}
\label{tab:case-gender-table}
\end{table}

Our analysis examines 
\begin{enumerate*}[label=(\roman*)]
    \item how applying a learned gradient direction affects article probabilities beyond the specific trained gender–case transition, and \item the overlap among the most affected model parameters across  gender–case settings.
\end{enumerate*}
We find statistically significant generalization across gender and case, as well as substantial neuron overlap between different transformations.
Overall, our results argue against a strictly rule-based encoding of German definite articles (\ref{hyp:ruleencoding}), indicating that in some contexts and nouns, article prediction is learned via memorized associations (\ref{hyp:memorization}) rather than  abstract grammatical rules.

For brevity, we will use \emph{article} to refer exclusively to \emph{German definite singular articles}.

\section{Related Work}

\subsection{Morphosyntactic Information in Model Representations}
A large body of work asks whether transformers encode linguistic information internally.
Probing studies suggest that syntactic structure is recoverable from representations, including hierarchical relations captured by structural probes \cite{hewitt-manning-2019-structural} and a layer-wise organization resembling a classical NLP pipeline \cite{tenney-etal-2019-bert}.
However, probe accuracy is not mechanistic evidence: the presence of a feature in model representations does not entail that it causally drives the model’s predictions \cite{belinkov-2022-probing}.

Beyond English, multilingual probing shows that morphosyntactic features such as case, gender, and number are often accessible in model representations, with substantial variation across languages and phenomena \cite{acs2024morphosyntactic}.
Recoverability depends on how directly and unambiguously a feature is realized in surface form.
German case is explicitly identified as difficult because nouns are not case-inflected and case is marked on articles that jointly encode case and gender under high syncretism.
In gender-marking languages, noun representations also exhibit distributional traces of grammatical gender  (e.g., nouns sharing gender are closer in embedding space) \cite{gonen-etal-2019-grammatical-gender}, 
but such effects doesn’t imply rule-based use.

\subsection{Behavioral and Mechanistic Analyses of Grammar}

Controlled minimal pairs provide fine-grained behavioral tests of grammatical sensitivity in \glspl{lm}.
Early studies show that models can prefer grammatical continuations over minimally perturbed alternatives, yet show systematic failures as constructions become more complex \cite{linzen-etal-2016-assessing, marvin-linzen-2018-targeted}.
Large-scale benchmarks such as BLiMP 
reveal wide variation across phenomena \cite{warstadt-etal-2020-blimp-benchmark}, 
leaving open whether correct behavior reflects abstract rules or surface heuristics and memorized patterns.

To move beyond behavior, causal and mechanistic work intervenes on internal representations.
\citet{finlayson-etal-2021-causal}
show that modifying internal representations yields systematic changes in subject–verb agreement predictions, indicating that grammatical behavior depends on specific internal states.
Relatedly, \citet{ferrando-costa-jussa-2024-similarity} find highly similar circuit structures for subject–verb agreement across languages despite surface-level topological differences.

Recent sparse autoencoder (SAE) approaches \cite{bricken2023monosemanticity} further decompose activations into sparse features: \citet{brinkmann-etal-2025-large} identify multilingual features corresponding to morphosyntactic concepts such as number, gender, and tense, and \citet{jing-etal-2025-lingualens-sae} introduce LinguaLens, combining SAE features with counterfactual datasets and interventions to identify and manipulate mechanisms across linguistic phenomena.
These results suggest that grammatical concepts can align with reusable internal feature directions. We complement this line by testing how article-transition interventions distribute across the German gender-case paradigm.

\subsection{Memorization vs.\ Generalization}
A separate line of work documents that neural \glspl{lm} can \emph{memorize} training sequences in ways that enable verbatim extraction, and that memorization increases with scale and with training-data duplication \cite{carlini2022quantifying}.
For grammar, \emph{morphological productivity} offers a controlled test of rule-like generalization beyond frequent lexical items, e.g., 
Wug-style evaluations show uneven morphological generalization even for strong LMs \cite{weissweiler-etal-2023-counting}.
Complementarily, \citet{anh2024morphology} find that generalization to nonce words varies systematically across languages and is predicted by morphological complexity.
Together, these findings motivate our case study: high surface-level agreement can coexist with non-uniform generalization, and our gradient-based interventions probe whether German article behavior reflects reusable grammatical variables or surface-level associations.

\begin{figure*}[!t]
\centering
\def\predheight{0.5em}
\def\preddepth{0.1em}

\hspace{-2mm}

\begin{subfigure}[t]{0.55\linewidth}
\resizebox{1.015\linewidth}{!}{
\begin{tikzpicture}[
    font=\tiny,
    >=Latex,
    remember picture,
    meta/.style={
        rectangle,
        rounded corners=6pt,
        line width=0.9pt,
        inner sep=2pt,
        inner xsep=2pt,
        align=center
    },
  pred/.style={
    rectangle,
    rounded corners=2pt,
    line width=0.9pt,
    inner sep=2pt,
    align=center,
    text height=\predheight,
    text depth=\preddepth
  },
    arr/.style={->, very thick},
    procarr/.style={
    -{Triangle[width=6pt,length=8pt]},
    line width=1.6pt,
    draw=black!50
},
  stack/.style={
    draw=#1!40!black,
    fill=#1!10,
    rounded corners=6pt,
    line width=0.6pt,
    inner sep=2pt,
    text width=.31\columnwidth,
    text height=.5em,
    text depth=.7em,
    anchor=north
  },
    altcell/.style={
    line width=1.0pt
  },
  idcell/.style={
    line width=0.8pt
  }
]

\node (tbl) {
\begin{tabular}{lcccc}
\toprule
 & \textbf{Nom.} & \textbf{Acc.} & \textbf{Dat.} & \textbf{Gen.} \\
\midrule
\textbf{Male}    & \tikzmarknode{cell-other1}{\der} & \tikzmarknode{cell-other2}{\den} & \tikzmarknode{cell-other3}{\dem} &\tikzmarknode{cell-des}{\des} \\
\textbf{Neutral} & \tikzmarknode{cell-das}{\das} & \tikzmarknode{cell-other4}{\das} & \tikzmarknode{cell-dem}{\dem} & \tikzmarknode{cell-other5}{\des} \\
\textbf{Female}  & \tikzmarknode{cell-other6}{\die} & \tikzmarknode{cell-other7}{\die} & \tikzmarknode{cell-other8}{\der} & \tikzmarknode{cell-other9}{\der} \\
\bottomrule
\end{tabular}
};


      fit=(cell-other1), inner sep=1.6pt, opacity=0.4] (box-other1) {};

      fit=(cell-other2), inner sep=1.6pt, opacity=0.3] (box-other2) {};
      
      fit=(cell-other3), inner sep=1.6pt, opacity=0.2] (box-other3) {};
      
      fit=(cell-other4), inner sep=1.6pt, opacity=0.1] (box-other4) {};
      
      fit=(cell-other5), inner sep=1.6pt, opacity=0.08] (box-other5) {};
      
      fit=(cell-other6), inner sep=1.6pt, opacity=0.06] (box-other6) {};
      
      fit=(cell-other7), inner sep=1.6pt, opacity=0.04] (box-other7) {};

      fit=(cell-other8), inner sep=1.6pt, opacity=0.02] (box-other8) {};
      fit=(cell-other9), inner sep=1.6pt, opacity=0.01] (box-other9) {};

\node[draw=edgeAlt, altcell, line width=1.1pt, rounded corners=1pt,
      fit=(cell-das), inner sep=1.6pt] (box-das) {};
\node[draw=edgeAlt, line width=1.1pt, rounded corners=1pt,
      fit=(cell-dem), inner sep=1.6pt] (box-dem) {};
\node[draw=edgeId, line width=1.1pt, rounded corners=1pt,
      fit=(cell-des), inner sep=1.6pt] (box-des) {};

\coordinate (col-das) at (box-das.south);
\coordinate (col-dem) at (box-dem.south);
\coordinate (col-des) at (box-des.south);
\coordinate (left-align)  at (tbl.west);
\coordinate (right-align) at (tbl.east);

\coordinate (hyp-center) at ($(tbl.south) + (0,-0.4cm)$);

\def\xsep{0.35\columnwidth}

\node[meta,
      draw=edgeAlt, fill=artDas,
      text width=.31\columnwidth,
      text height=.5em,
    text depth=0.7em,
      anchor=north]
      (hyp-nn)
      at ($(hyp-center) + (-\xsep,0)$)
{[MASK] Ergebnis\\bestätigt die Hypothese.};

\node[meta,
      draw=edgeAlt, fill=artDem,
      text width=.31\columnwidth,
      text height=.5em,
       text depth=0.7em,
      anchor=north]
      (hyp-dn)
      at (hyp-center)
{Beispiele befinden\\sich in [MASK] Datensatz.};

\node[stack=artDes, opacity=0.25, fill=artDer, draw=edgeId]
  at ($(hyp-center)+(\xsep+0.2em,0.2em)$) {};

\node[stack=artDes, opacity=0.4, fill=artDie, draw=edgeId]
  at ($(hyp-center)+(\xsep+0.1em,0.1em)$) {};

\node[meta,
      draw=artDes!70!black, fill=artDes,
      text width=.31\columnwidth,
      text height=.5em,
      text depth=0.7em,
      draw=edgeId,
      anchor=north]
      (hyp-ds)
      at ($(hyp-center)+(\xsep,0)$)
{Die Aktivierung [MASK] Neurons ist hoch.};

\draw[arr, draw=black!50]
  (box-das) --
  node[right,  pos=.7]  {\fontsize{5}{6}\selectfont \dataNN}
  (hyp-nn.north);

\draw[arr, draw=black!50]
  (box-dem) -- 
  node[right, pos=.7] {\fontsize{5}{6}\selectfont\dataDN}
  (hyp-dn.north);

\draw[arr, draw=black!50]
  (box-des) -- 
  node[right, pos=.7] {\fontsize{5}{6}\selectfont\dataGM} 
  (hyp-ds.north);

\node[pred,
      draw=edgeAlt, fill=artDas,
      text width=.14\columnwidth,
      anchor=north east]
      (nn-ok)
      at ($(hyp-nn.south) + (-0.1em,-0.5cm)$)
{\tiny$y^F{=}das$};

\node[pred,
      draw=edgeAlt, fill=artDem,
      text width=.15\columnwidth,
      anchor=north west]
      (nn-bad)
      at ($(hyp-nn.south) + (0.05em,-0.5cm)$)
{\tiny$y^A{=}dem$};

\node[pred,
      draw=edgeAlt, fill=artDem,
      text width=.15\columnwidth,
      anchor=north east]
      (dn-bad)
      at ($(hyp-dn.south) + (-0.05em,-0.5cm)$)
{\tiny$y^F{=}dem$};

\node[pred,
      draw=edgeAlt, fill=artDas,
      text width=.145\columnwidth,
      anchor=north west]
      (dn-ok)
      at ($(hyp-dn.south) + (0.1em,-0.5cm)$)
{\tiny$y^A{=}das$};

\node[pred,
      draw=edgeId, fill=artDer, opacity=0.30,
      text width=.145\columnwidth, anchor=north east]
      at ($(hyp-ds.south)+(-0.1em+0.1em,-0.5cm+0.1em)$) {};

\node[pred,
      draw=edgeId, fill=artDie, opacity=0.18,
      text width=.145\columnwidth, anchor=north east]
      at ($(hyp-ds.south)+(-0.1em+0.2em,-0.5cm+0.2em)$) {};

\node[pred,
      draw=edgeId, fill=artDer, opacity=0.30,
      text width=.14\columnwidth, anchor=north west]
      at ($(hyp-ds.south)+(0.1em+0.1em,-0.5cm+0.1em)$) {};

\node[pred,
      draw=edgeId, fill=artDie, opacity=0.18,
      text width=.14\columnwidth, anchor=north west]
      at ($(hyp-ds.south)+(0.1em+0.2em,-0.5cm+0.2em)$) {};

\node[pred,
      fill=artDes, draw=edgeId,
      text width=.145\columnwidth, anchor=north east]
      (ds-bad)
      at ($(hyp-ds.south)+(-0.1em,-0.5cm)$)
{\tiny $y^F{=}des$};

\node[pred,
      fill=artDes, draw=edgeId,
      text width=.14\columnwidth, anchor=north west]
      (ds-ok)
      at ($(hyp-ds.south)+(0.1em,-0.5cm)$)
{\tiny $y^A{=}des$};

\draw[arr, draw=black!50] (hyp-nn.south) -- (nn-ok.north);
\draw[arr, draw=black!50] (hyp-nn.south) -- (nn-bad.north);

\draw[arr, draw=black!50] (hyp-dn.south) -- (dn-bad.north);
\draw[arr, draw=black!50] (hyp-dn.south) -- (dn-ok.north);

\draw[arr, draw=black!50] (hyp-ds.south) -- (ds-bad.north);
\draw[arr, draw=black!50] (hyp-ds.south) -- (ds-ok.north);

\end{tikzpicture}}
\caption{Illustration of factual ($y^F$) and alternative ($y^A$) targets for the gender-case transition $(\neutral,\nominative)\rightleftarrows(\neutral,\dative)$.
Non-target cells form identity pairs (only one shown). Dataset labels (e.g., \dataNN) denote the corresponding gender-case datasets (Section~\ref{sec:data}).}
\label{fig:overview-factual-counterfactual}
\end{subfigure}
\hspace{10pt}
\begin{subfigure}[t]{0.4\linewidth}
    \begin{tikzpicture}[
      netnode/.style={circle, draw, minimum size=2.5mm, inner sep=0pt, fill=white},
  netedge/.style={-Latex, line width=0.55pt},
  altcell/.style={line width=1.1pt},
    ]

\coordinate (Xin)  at (0,0);
\coordinate (Xh)   at (0.8cm,0);   
\coordinate (Xout) at (1.6cm,0);   

\foreach \i in {1,...,7}{
  \node[netnode] (in\i) at ([yshift=-(\i-1)*3mm]Xin) {};
}

\node[netnode] (h) at ([yshift=-9mm]Xh) {};

\foreach \j in {1,...,7}{
  \node[netnode] (out\j) at ([yshift=-(\j-1)*3mm]Xout) {};
}

\foreach \i in {1,...,7}{
  \draw[netedge] (in\i) -- (h);
}
\foreach \j in {1,...,7}{
  \draw[netedge] (h) -- (out\j);
}

\node[font=\scriptsize, anchor=north] at ([yshift=-0mm]in7.south) {$\nabla^{A}W_m$};
\node[font=\scriptsize, anchor=north] at ([yshift=-10mm]h.south) {$h$};
\node[font=\scriptsize, anchor=north] at ([yshift=-0mm]out7.south) {$\nabla^{\Delta}W_m$};


\node[font=\scriptsize, anchor=north] (gradiendmodel)
  at ([yshift=16.6mm]h.north) {\textbf{\gradiend\ Model}};

\node[font=\tiny, anchor=north west]
  at ([yshift=1.3mm]gradiendmodel.south west) {Encoder};

\node[font=\tiny, anchor=north east]
  at ([yshift=1.3mm]gradiendmodel.south east) {Decoder};

\def\PlotW{1.7cm}        
\def\PlotGap{15mm}        
\def\ArrowStartGap{2.5mm}
\def\ArrowEndGap{1.5mm}  

\node[draw=none, fit=(in1)(in7)(h)(out1)(out5), inner sep=2mm] (Gproc) {};

\coordinate (yEnc)  at ($(Gproc.north)!0.1!(Gproc.south)$);
\coordinate (yProb) at ($(Gproc.north)!0.50!(Gproc.south)$);
\coordinate (yVenn) at ($(Gproc.north)!0.9!(Gproc.south)$);

\coordinate (plotsX) at ([xshift=\PlotGap]Gproc.east);

\node[anchor=west] (r1img) at (plotsX |- yEnc)
  {\includegraphics[width=\PlotW]{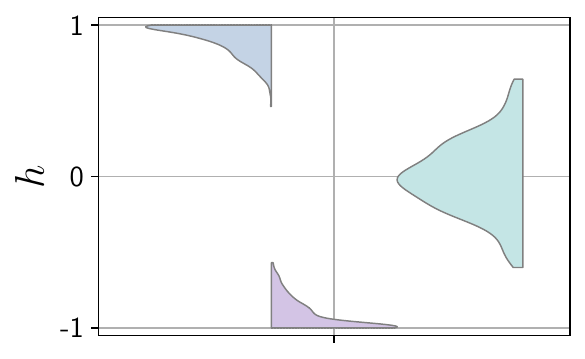}};

\node[anchor=west] (r2img) at (plotsX |- yProb)
  {\includegraphics[width=\PlotW]{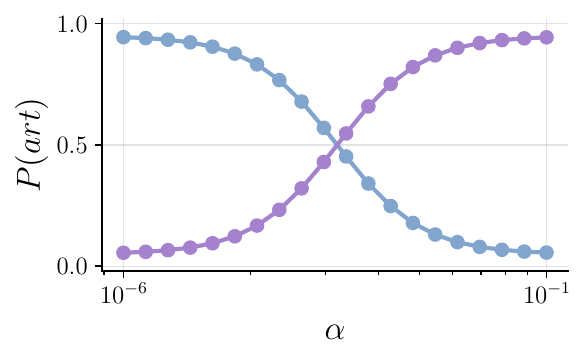}};

\node[anchor=west] (r3img) at (plotsX |- yVenn)
  {\includegraphics[width=\PlotW]{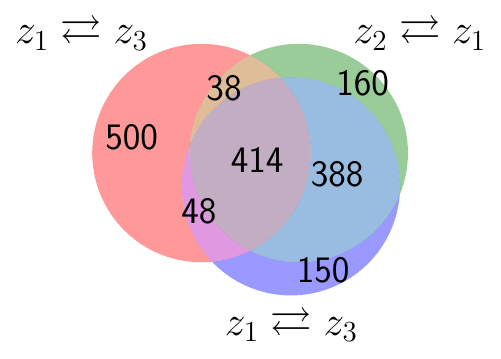}};

\tikzset{procarrow/.style={-{Triangle[length=5.2mm,width=3.4mm]}, line width=3.0pt}}

\coordinate (procX) at ([xshift=\ArrowStartGap]Gproc.east);

\draw[procarrow] (procX |- yEnc)  -- ([xshift=-\ArrowEndGap]r1img.west);
\draw[procarrow] (procX |- yProb) -- ([xshift=-\ArrowEndGap]r2img.west);
\draw[procarrow] (procX |- yVenn) -- ([xshift=-\ArrowEndGap]r3img.west);

    \end{tikzpicture}
    \caption{\rev{
    \gradiend\ is trained for a given transition using alternative gradients ($\nabla^A$) and the difference to factual gradients ($\nabla^\Delta$). We evaluate the model via \begin{enumerate*}[label=(\roman*)]
\item encoder analysis,
\item decoder-based probability shifts, and
\item weight overlap across transitions.
\end{enumerate*}
    }
    }\label{fig:gradiend}

\end{subfigure}

\caption{\rev{
Overview: Using gender-case-specific data, we learn \gradiend\ models from MLM/CLM gradients of factual vs.\ alternative targets for two gender-case cells. \gradiend\ learns to map input gradients to a scalar label $h$ ($\pm 1$ for the two target classes; $0$ for neutral inputs) and enables analyses to test our hypotheses.
}}\label{fig:overview}

\end{figure*}

\section{Methodology}\label{sec:methodology}

\rev{
Our goal is to probe how a \gls{lm} chooses German definite articles (see Table~\ref{tab:case-gender-table}).
We do this by asking a counterfactual question: \emph{if the article in a given context were different, which parameters would need to change to make the model prefer the alternative article?} 
Concretely, we use a factual target article and a controlled counterfactual (an article swap between two gender-case cells). 
The resulting gradient difference isolates an update direction associated with this specific grammatical transition.
The \acrlong{gradiend} (\acrshort{gradiend}\glsunset{gradiend}; \citealt{gradiend}) operationalizes this idea by compressing gradients into a single scalar feature and decoding it back into an update direction.
This enables three analyses (see Figure~\ref{fig:overview}):
\begin{enumerate*}[label=(\roman*)]
    \item \textbf{Encoder analysis:} whether and how the learned feature separates gradients from opposite swap directions and neutral inputs, revealing similarities across inputs;
    \item \textbf{Probability shifts:} applying decoded update directions to test which article probabilities change (and which remain stable), linking interventions to our hypotheses; and
    \item \textbf{Weight overlap:} comparing learned directions across transitions to quantify parameter reuse between conceptually different updates.
\end{enumerate*}
}

\subsection{German Definite Articles as a Controlled Morphosyntactic System}
German articles form a small closed-class paradigm whose surface form is determined by grammatical gender and case.
Male, neutral, and female gender labels are represented by  $\mathcal{G}\,{\coloneqq}\,\{\textsc{Masc},\textsc{Neut},\textsc{Fem}\}$, and the German cases nominative, accusative, dative, and genitive are represented by $\mathcal{C}\,{\coloneqq}\,\{\nominative, \accusative, \dative, \genitive\}$. We represent each gender-case combination as a \emph{cell} $z{=}(g,c)\,{\in}\,\mathcal{G}\,{\times}\,\mathcal{C}$
 and denote its article by $a(g,c)\,{\in}\,\mathcal{A}$ (defined by Table~\ref{tab:case-gender-table}) with
$\mathcal{A}\,{=}\,\{\textit{der, die, das, den, dem, des}\}$.
Due to \emph{syncretism}, multiple $(g,c)$ pairs share the same surface article (e.g., $a(\male,\nominative)\,{=}\,\textit{der}\,{=}\,a(\female,\dative)$).
This lets us test whether models condition article choice on abstract $(g,c)$ variables or on surface-level token-context associations.

\subsection{Article Prediction Task}
We study \glspl{lm} in a \acrlong{mlm} (\acrshort{mlm}\glsunset{mlm}; \citealt{bert}) setting using an article as masked target.
Given a sentence, we construct an input by masking every article occurrence corresponding to a targeted gender-case cell $z\,{=}\,(g, c)$, while leaving the remaining context unchanged.

For each masked instance, we define two targets:
\begin{enumerate*}[label=(\roman*)]
    \item \textbf{Factual target} $y^{F}$, the grammatically licensed article for $z\,{=}\,(g, c)$ specified by the sentence context, i.e., $y^{F}\,{=}\,a(g,c)$.
    \item \textbf{Alternative target} $y^{A}$, an article specified by a predefined transition between two cells (defined in the next subsection).
\end{enumerate*}

These induce corresponding factual ($\nabla^{F} W_m$) and alternative ($\nabla^{A} W_m$) gradients with respect to the selected model parameters $W_m$.
We define their difference as $\nabla^{\Delta} W_m \,{\coloneqq}\, \nabla^{F} W_m - \nabla^{A} W_m$.

\subsection{\gradiend\ for German Gender}\label{sec:gradiend-for-german-gender}
We train one \gradiend\ model (Figure~\ref{fig:gradiend}) per targeted transition
$T\,{=}\,(z_1\,{\rightleftarrows}\,z_2)$ between gender-case cells $z_i\,{=}\,(g_i,c_i)$ differing in exactly one dimension: gender at fixed case ($g_1\,{\neq}\, g_2$, $c_1\,{=}\,c_2$)  or case at fixed gender ($g_1\,{=}\,g_2$, $c_1\,{\neq}\, c_2$).
For instance, the nominative gender transition 
$z_1 \!,{=}\,(\textsc{Masc},\textsc{Nom})$ and $z_2\,{=}\,(\textsc{Fem},\textsc{Nom})$ corresponds to the article transition $\textit{der}\,{\leftrightarrow}\, \textit{die}$.
For masking tasks for $z_1$ and $z_2$, we construct \emph{swapped target pairs}: 
for $z_1$ we set $(y^{F},y^{A})\,{=}\,(a(z_1),a(z_2))$, and for $z_2$ we set $(y^{F},y^{A})\,{=}\,(a(z_2),a(z_1))$.
These swaps induce non-zero gradient differences $\nabla^{\Delta} W_m$ that encode the transition direction $T$.

To keep the learned update specific to $T$, we additionally include masking tasks from all other cells $z\,{\notin}\,\{z_1,z_2\}$ as \emph{identity pairs}. 
Under this construction, the factual and alternative gradients are identical by definition, yielding $\nabla^{\Delta} W_m = 0$.
This explicitly enforces a \emph{do-not-change} constraint: \gradiend\ is trained to produce no update for non-targeted gender-case settings.

Figure~\ref{fig:overview-factual-counterfactual} illustrates the construction.
We denote \gradiend\ models targeting gender transitions at fixed case $c$ as $G^{g_1,g_2}_c$ and case transitions at fixed gender $g$ as $G^{g}_{c_1,c_2}$, e.g., \gradcNgMF\ and \gradcNDgN.

\subsection{\gradiend\ Architecture and Training}
\gradiend\ learns a bottleneck encoder-decoder $f\,{=}\,\dec\circ\enc$ that maps gradient information to a single scalar feature and decodes it into a parameter-space update direction \rev{(see Figure~\ref{fig:gradiend})}.
We use a one-dimensional bottleneck $h\in[-1,1]$:
\begin{align*}
h = \enc(\nabla_{\text{in}}W_m)
  &= \tanh\!\big(W_e^\top \nabla_{\text{in}}W_m + b_e\big),\\
\nabla_{\text{out}}W_m \approx \dec(h)
  &= h\cdot W_d + b_d ,
\end{align*}
where $W_m\,{\in}\,\mathbb{R}^n$ are the selected model parameters and
$W_e,W_d,b_d\,{\in}\,\mathbb{R}^n$, $b_e\in\mathbb{R}$ are learned.

Departing from \citet{gradiend}, who use factual gradients as input, we set $\nabla_{\text{in}}W_m\,{\coloneqq}\,\nabla^{A}W_m$ and $\nabla_{\text{out}}W_m\,{\coloneqq}\,\nabla^{\Delta}W_m$.
Article prediction is often highly confident, making factual gradients near-zero and \gradiend\ training unstable. Alternative targets of the $(g,c)$, by construction, typically receive substantially lower probability, producing more informative gradients.

We train \gradiend\ with the 
reconstruction loss
\[
\mathcal{L}_\gradiend
=
\left\|
\dec(\enc(\nabla_{\text{in}}W_m))
-
\nabla_{\text{out}}W_m
\right\|_2^2,
\]
encouraging $h$ to encode the targeted transition while remaining neutral for identity pairs.

After training, \gradiend\ yields an update direction for any $h^\star\,{\in}\,\mathbb{R}$ via $\dec(h^\star)$.
We intervene on the base model with
$\widetilde{W}_m \,{=}\, W_m \,{+}\, \alpha \,{\cdot}\, \dec(h^\star)$ with \emph{learning rate} $\alpha$. 

\section{Data}\label{sec:data}


\begin{table*}[!t]
    \centering
    \small
    \begin{tabular}{lcc|cc}
    \toprule
   \textbf{Article Pair} & \multicolumn{2}{c}{\textbf{Datasets}} & \multicolumn{2}{c}{\textbf{\gradiend\ Variants}} \\ \cmidrule(lr){1-1}\cmidrule(lr){2-3}\cmidrule(lr){4-5}
     & \textbf{Left Article} & \textbf{Right Article} & \textbf{Gender Transition} & \textbf{Case Transition} \\\midrule
     \multicolumn{5}{c}{\rev{\textbf{Two-dimensional transitions (gender \emph{and} case vary)}}} \\ \midrule
    $der\!\leftrightarrow\!die$  & \dataNM, \dataDF, \dataGF & \dataNF, \dataAF & \gradcNgMF & \gradcNDgF, \gradcNGgF, \gradcDAgF, \gradcGAgF \\
       
        $der\!\leftrightarrow\! dem$ & \dataNM, \dataDF & \dataDM, \dataDN &  \gradcDgMF, \gradcDgFN &  \gradcNDgM \\
        $der\!\leftrightarrow\!des$ & \dataNM, \dataGF & \dataGM, \dataGN  &  \gradcGgFM, \gradcGgFN & \gradcNGgM \\
        
        \midrule

\multicolumn{5}{c}{\rev{\textbf{One-dimensional transitions (only gender \emph{or} only case varies)}}} \\\midrule
        
        $das\!\leftrightarrow\!die$ & \dataNF, \dataAF & \dataNN, \dataAN & \gradcNgFN, \gradcAgFN & \\
       $das\!\leftrightarrow\!dem$ & \dataNN, \dataAN & \dataDN & & \gradcNDgN, \gradcADgN \\
       $das\!\leftrightarrow\!des$ & \dataNN, \dataAN & \dataGN & & \gradcNGgN, \gradcAGgN \\
        \rev{$dem\!\leftrightarrow\!des$} & \rev{\dataDM, \dataDN} & \rev{\dataGM}, \rev{\dataGN} & & \rev{\gradcGDgM, \gradcGDgN} \\

        \bottomrule
    \end{tabular}
    \caption{Targeted bidirectional article transitions and \gradiend\ variants.
Listed are all trained transitions grouped by their structural diversity (two- vs. one-dimensional), together with the corresponding datasets and model variants.}
    \label{tab:article-transitions}
\vspace{-0.3em}
\end{table*}

\begin{figure}[!t]
\centering

\begin{tikzpicture}[
    font=\footnotesize,
    >=Latex,
    remember picture,
    data/.style={
        rectangle,
        rounded corners=2pt,
        draw=black!45,
        line width=0.6pt,
        fill=black!3,
        inner sep=3pt,
        align=center
    },
    meta/.style={
        rectangle,
        rounded corners=6pt,
        draw=black!60,
        line width=0.9pt,
        fill=black!10,
        inner sep=3pt,
        align=center
    },
    article-masc/.style={
        rectangle,
        draw=red!70,
        fill=red!10,
        line width=0.8pt,
        inner sep=2pt
    },
    article-fem/.style={
        rectangle,
        draw=blue!70,
        fill=blue!10,
        line width=0.8pt,
        inner sep=2pt
    },
    article-unused/.style={
        rectangle,
        draw=black!30,
        fill=black!5,
        line width=0.6pt,
        inner sep=2pt
    },
    arr/.style={->, very thick}
]

\node[data] (input) {%
\tikzmarknode[article-fem]{ArtNOMM}{\textbf{\textcolor{blue}{Der}}} Algorithmus analysiert
\tikzmarknode[article-fem]{ArtACCF}{\textbf{\textcolor{blue}{die}}} Struktur
\tikzmarknode[article-unused]{ArtPL}{\textbf{\textcolor{gray}{der}}} Daten.
};
\coordinate (col-masc) at (ArtNOMM.south);
\coordinate (col-fem)  at (ArtACCF.south);
\coordinate (left-align)  at (input.west);
\coordinate (right-align) at ($(input.east) + (-1cm,0)$);

\node[font=\scriptsize, text=black!50, anchor=north]
  at (ArtPL.south)
  {plural \textcolor{gray}{\ding{55}}};

\node[meta,
      anchor=north west]
      (meta-masc)
      at ($(left-align |- col-masc) + (0,-0.6cm)$)
{$z=(\textsc{Masc},\textsc{Nom})$};

\node[data,
      anchor=north west]
      (masked-masc)
      at ($(left-align |- meta-masc.south) + (0,-0.7cm)$)
{[MASK] Algorithmus \\ analysiert die\\ Struktur der Daten.};

\node[meta,
      anchor=north east]
      (meta-fem)
      at ($(right-align |- col-fem) + (0,-0.7cm)$)
{$z=(\textsc{Fem},\textsc{Acc})$};

\node[data,
      anchor=north east]
      (masked-fem)
      at ($(right-align |- meta-fem.south) + (0,-0.7cm)$)
{Der Algorithmus \\analysiert [MASK]\\ Struktur der Daten.};

\draw[arr, overlay, draw=black!50]
  (ArtNOMM.south) -- node[right] {spaCy} (col-masc |- meta-masc.north);

\draw[arr, overlay, draw=black!50]
  (ArtACCF.south) -- node[right] {spaCy} (col-fem |- meta-fem.north);

\draw[arr, draw=black!50]
  (col-masc |- meta-masc.south) -- node[right] {\dataNM} (col-masc |- masked-masc.north);

\draw[arr, draw=black!50]
  (col-fem |- meta-fem.south) -- node[right] {\dataAF} (col-fem |- masked-fem.north);

\end{tikzpicture}
\vspace{-0.4em}
\caption{Data generation: spaCy determines gender and case of articles to determine the target dataset.}
\label{fig:data-overview}
\end{figure}
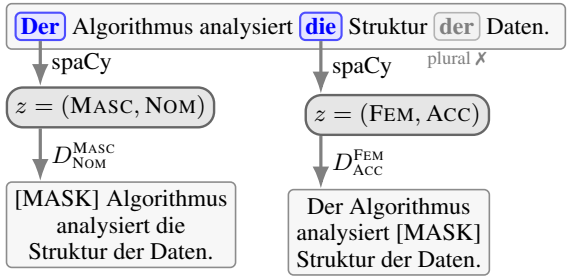

To extract gender-case–specific article transition gradients, we construct one dataset for each cell $z\,{=}\,(g,c)\,{\in}\,\mathcal{G}\,{\times}\,\mathcal{C}$.
Each dataset contains sentences in which only articles corresponding to  $z$ are masked.
We filter German Wikipedia sentences \cite{wikipedia}, retaining only those where spaCy \cite{Honnibal_spaCy_Industrial-strength_Natural_2020} identifies a definite singular article with the desired gender and case (Figure~\ref{fig:data-overview}).
We denote the resulting datasets by $\datasymbol^g_c$ (e.g., \dataNM\ for masculine-nominative).
Sizes range from 19K to 61K  (Table~\ref{tab:data-size}), and each dataset is split into train (80\%), validation (10\%), and test (10\%) subsets.

To probe behavior without gender/case cues, we construct \rev{\dataNEUT, a dataset with minimized gender and case cues.}
It is derived from the Wortschatz Leipzig German news corpus \cite{goldhahn2012lcc, wortschatz_deu_news_2024_300K} and filtered to exclude sentences containing determiners, definite or indefinite articles, or third-person pronouns.
The resulting dataset serves as a gender–case–independent reference in our analyses.

Full generation details are in Appendix~\ref{app:data}.

\section{Experiments}
We evaluate whether German definite articles in \glspl{lm} are memorized from context (\ref{hyp:memorization}) or  determined via abstract grammatical representations (\ref{hyp:ruleencoding}).
We analyze \gradiend\ models from three complementary angles:
\begin{enumerate*}[label=(\roman*)]
    \item how encoded values $h$ distribute, 
    \item how applying the decoded update affects article probabilities across gender-case cells, and
    \item how similar the learned update directions are in parameter space via Top-$k$ weight overlap.
\end{enumerate*}

\subsection{Experimental Setup}\label{sec:experiental-setup}

\textbf{Models.}
We study four German models (\bert, \gbert, \modernbert, \gpttwo) and two multilingual models (\eurobert, \llama), 
covering encoder-only and decoder-only transformers. 
We include \modernbert\ (1B parameters) as an intermediate-size model between the smaller German models (109M--336M) and \llama\ (3.2B).
Table~\ref{tab:models} summarizes  architectures and sizes.

\textbf{Targeted transitions.}
Across the German article paradigm, transitions cluster 
into:
\begin{enumerate*}[label=(\roman*)]
\item \emph{two-dimensional} groups (gender- \emph{and} case-based transitions within the same article pair),
\item \emph{one-dimensional} groups (multiple transitions along a single dimension),
and 
\item \emph{singleton} groups (a single transition).
\end{enumerate*}
We focus on two- and one-dimensional groups, which enable within-group comparisons, and list all these transitions in Table~\ref{tab:article-transitions}. 
For each of these transitions $T\,{=}\,(z_1\,{\rightleftarrows}\,z_2)$, we train one \gradiend\ model, inducing two \emph{directed} transitions $z_1\,{\to}\,z_2$ and $z_2\,{\to}\,z_1$. 

\textbf{Training.} We train \gradiend\ models as described in Section~\ref{sec:methodology}, using swapped targets for the two cells defining $T$ and identity pairs for all remaining cells, with the gender-case datasets from Section~\ref{sec:data}. 
For consistent visualization across base models, we normalize the sign of the encoded value so the same targeted directional article transition has a consistent polarity (positive vs.\ negative $h$).
Training details are provided in Appendix~\ref{app:training}.

\textbf{Evaluation.}
Unless stated otherwise, we evaluate on the test splits of the corresponding datasets.

\textbf{Decoder-only models.}
German articles depend on the noun’s gender, so left-to-right context is insufficient. Hence, we add a \gls{mlm}-style article classifier for bidirectional conditioning (Appendix~\ref{app:decoder-only}).

\subsection{Feature Encoding Analysis}

\begin{table*}[!tb]
    \centering
    \setlength{\tabcolsep}{3pt}   
   \fontsize{8}{9}\selectfont
\begin{tabular}{l *{20}{r}}
\toprule
& \multicolumn{5}{c}{$der\!\leftrightarrow\!die$} & \multicolumn{3}{c}{$der\!\leftrightarrow\!dem$} & \multicolumn{3}{c}{$der\! \leftrightarrow\!des$} & \multicolumn{2}{c}{$das\!\leftrightarrow\!die$} & \multicolumn{2}{c}{$das\! \leftrightarrow\!dem$} & \multicolumn{2}{c}{$das\! \leftrightarrow\!des$} & \multicolumn{2}{c}{\rev{$dem\!\leftrightarrow\!des$}} \\
\cmidrule(lr){2-6}\cmidrule(lr){7-9}\cmidrule(lr){10-12}\cmidrule(lr){13-14}\cmidrule(lr){15-16}\cmidrule(lr){17-18}\cmidrule(lr){19-20}
 & \rotatebox{90}{\gradcNgMF}
 & \rotatebox{90}{\gradcNDgF}
 & \rotatebox{90}{\gradcNGgF}
 & \rotatebox{90}{\gradcADgF} 
 & \rotatebox{90}{\gradcGAgF}\,\,
 & \rotatebox{90}{\gradcDgMF}
 & \rotatebox{90}{\gradcDgFN}
 & \rotatebox{90}{\gradcNDgM}\,\,
 & \rotatebox{90}{\gradcGgFM}
 & \rotatebox{90}{\gradcGgFN}
 & \rotatebox{90}{\gradcNGgM}\,\,
 & \rotatebox{90}{\gradcAgFN}
 & \rotatebox{90}{\gradcNgFN}\,\,
 & \rotatebox{90}{\gradcNDgN}
 & \rotatebox{90}{\gradcADgN}\,\,
 & \rotatebox{90}{\gradcNGgN}
 & \rotatebox{90}{\gradcAGgN}\,\,
  & \rotatebox{90}{\rev{\gradcGDgM}}
 & \rotatebox{90}{\rev{\gradcGDgN}}\,\,
\\
\midrule

\bert	& 96.3	& 95.7	& 97.1	& 95.3	& 91.9	& 96.4	& 95.8	& 95.7	& 96.5	& 95.7	& 96.6	& 93.5	& 97.8	& 96.5	& 90.7	& 96.6	& 94.1	& \rev{98.4}	& \rev{97.0}\\
\gbert	& 98.4	& 97.8	& 96.3	& 96.1	& 96.2	& 98.6	& 98.4	& 97.8	& 97.9	& 97.7	& 95.6	& 96.3	& 98.6	& 96.6	& 95.3	& 97.1	& 97.3	& \rev{98.2}	& \rev{97.4} \\
\modernbert	& 94.1	& 93.9	& 93.3	& 84.3	& 82.7	& 90.2	& 88.7	& 93.9	& 87.2	& 85.5	& 92.2	& 88.7	& 95.0	& 93.5	& 84.8	& 91.2	& 81.3	& \rev{94.6}	& \rev{90.9}\\
\eurobert	& 61.9	& 61.3	& 59.4	& 52.2	& 50.6	& 61.2	& 59.8	& 61.3	& 50.7	& 50.6	& 64.7	& 64.7	& 72.9	& 61.7	& 52.3	& 65.7	& 55.1	& \rev{66.2}	& \rev{53.3}\\
\gpttwo	& 56.6	& 64.3	& 58.8	& 57.8	& 51.0	& 69.1	& 59.5	& 64.3	& 61.7	& 58.0	& 67.3	& 58.2	& 71.1	& 63.1	& 57.1	& 62.5	& 67.7	& \rev{65.6}	& \rev{57.1} \\
\llama	& 59.2	& 62.1	& 58.5	& 55.8	& 52.5	& 61.3	& 55.2	& 62.1	& 58.9	& 51.1	& 57.6	& 52.9	& 67.0	& 57.7	& 50.5	& 62.9	& 54.6	& \rev{63.9}	& \rev{55.8} \\

\bottomrule
\end{tabular}

    \caption{Pearson correlation of encoded values, scaled by $100$.}
    \label{tab:encoded-values-correlations}
    \vspace{-0.7em}
\end{table*}

\begin{figure}[!t]
    \centering

  \includegraphics[width=\linewidth]{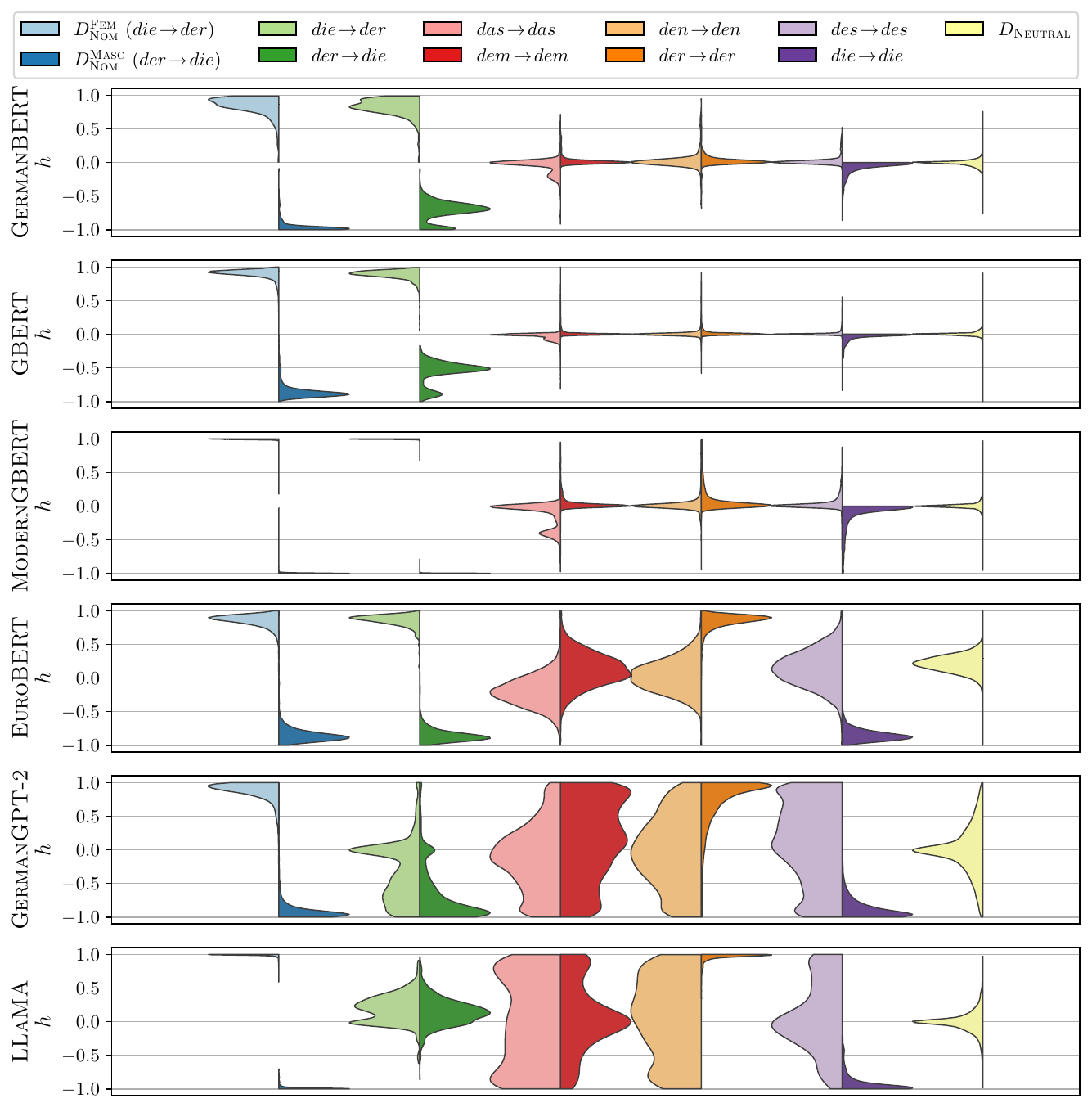}
    
    \caption{Encoded value distribution of \gradcNgMF\ 
    (other \glspl{gradiend} in Figures~\ref{fig:encoded-all-models-ND_F}-\ref{fig:encoded-all-models-GD_N}).}
    \label{fig:encoded-all-models-N_MF}
    \vspace{-0.5em}
\end{figure}

We analyze how \gradiend\ maps gradient inputs to the scalar bottleneck value $h$.
Figure~\ref{fig:encoded-all-models-N_MF} shows the encoded-value distributions for a representative \gradiend\ variant, \gradcNgMF, across all base models (other variants in Appendix~\ref{app:encoder}).
Table~\ref{tab:encoded-values-correlations} complements this view by reporting correlations between $h$ and our expected labels, assigning $\pm1$ to the two directed transition tasks and $0$ to identity pairs (neutral updates).
All models reach correlations of at least $50\%$ across all \glspl{gradiend}, while German encoder-only models often exceed $90\%$.

\textbf{Stable orientation on the targeted transition.}
Across models, the two targeted directed transitions (blue in Figure~\ref{fig:encoded-all-models-N_MF})  map to opposite signs, consistently separating the 
directional article transitions.

\textbf{Other realizations of the same article transition.}
Beyond the trained pair $(z_1,z_2)$, we evaluate the encoder on other gender-case pairs $(\tilde z_1,\tilde z_2)$ that realize the same article transitions $a(\tilde z_1)\,{\to}\,a(\tilde z_2)$ and vice versa (green).
Encoder-only models often assign these non-target transitions the same signed encoding as the trained transition, suggesting that gradient directions for different realizations of an article transition are closely aligned.
Decoder-only models show less stable behavior, with encodings frequently clustering around zero rather than the extremes $\pm 1$, probably due to the custom MLM-style prediction head used during \gradiend\ training.

\begin{figure*}[!t]
\centering
\begin{subfigure}[t]{0.3\linewidth}
\centering

\begin{tikzpicture}[
    remember picture,
    font=\scriptsize,
    >=Latex,
    arr/.style={->, thick, draw=black!60},
    boxsrc/.style={draw=blue!70!black, line width=1.1pt, rounded corners=1pt},
    boxtgt/.style={draw=red!70!black,  line width=1.1pt, rounded corners=1pt}
]
\node {
\setlength{\tabcolsep}{2pt}
\begin{tabular}{lcccc}
\toprule
 & \textbf{Nom.} & \textbf{Acc.} & \textbf{Dat.} & \textbf{Gen.} \\
\midrule
\textbf{Male}
 & {\tikzmarknode{a-mnom}{\der}} & \den & \dem & \des \\
\textbf{Neutral}
 & \das & \das & \dem & \des \\
\textbf{Female}
 & {\tikzmarknode{a-fnom}{\die}} & \die & \der & \der \\
\bottomrule \\  \\
\end{tabular}
};
\node[boxsrc, fit=(a-mnom), inner sep=1.5pt] {};
\node[boxtgt, fit=(a-fnom), inner sep=1.5pt] {};
\draw[arr] (a-mnom) -- (a-fnom);
\end{tikzpicture}
\caption{Local Rule (LR)}
\end{subfigure}
\hfill
\begin{subfigure}[t]{0.3\linewidth}
\centering
\begin{tikzpicture}[
    remember picture,
    font=\scriptsize,
    >=Latex,
    arr/.style={->, thick, draw=black!60},
    boxsrc/.style={draw=blue!70!black, line width=1.1pt, rounded corners=1pt},
    boxtgt/.style={draw=red!70!black,  line width=1.1pt, rounded corners=1pt}
]
\node[inner sep=0.pt] {
\setlength{\tabcolsep}{2pt}
\begin{tabular}{lcccc}
\toprule
 & \textbf{Nom.} & \textbf{Acc.} & \textbf{Dat.} & \textbf{Gen.} \\
\midrule
\textbf{Male}
 & {\tikzmarknode{b-mnom}{\der}}
 & {\tikzmarknode{b-macc}{\den}}
 & {\tikzmarknode{b-mdat}{\dem}}
 & {\tikzmarknode{b-mgen}{\des}} \\
\textbf{Neutral}
 & \das & \das & \dem & \des \\
\textbf{Female}
 & {\tikzmarknode{b-fnom}{\die}}
 & {\tikzmarknode{b-facc}{\die}}
 & {\tikzmarknode{b-fdat}{\der}}
 & {\tikzmarknode{b-fgen}{\der}} \\
\bottomrule \\ \\
\end{tabular}
};

\foreach \m/\f in {
    b-mnom/b-fnom,
    b-macc/b-facc,
    b-mdat/b-fdat,
    b-mgen/b-fgen%
}{
    \node[boxsrc, fit=(\m), inner sep=1.5pt] {};
    \node[boxtgt, fit=(\f), inner sep=1.5pt] {};
    \draw[arr] (\m) -- (\f);
}
\end{tikzpicture}
\caption{Generalized Rule (GR)}
\end{subfigure}
\hfill
\begin{subfigure}[t]{0.3\linewidth}
\centering
\begin{tikzpicture}[
    remember picture,
    font=\scriptsize,
    >=Latex,
    arr/.style={->, thick, draw=black!60},
    boxsrc/.style={draw=blue!70!black, line width=1.1pt, rounded corners=1pt},
    boxtgt/.style={draw=red!70!black,  line width=1.1pt, rounded corners=1pt},
    boxspill/.style={draw=red!60!black, dashed, line width=1.1pt, rounded corners=1pt}
]
\node[inner sep=0.pt] {
\setlength{\tabcolsep}{2pt}
\begin{tabular}{lcccc}
\toprule
 & \textbf{Nom.} & \textbf{Acc.} & \textbf{Dat.} & \textbf{Gen.} \\
\midrule
\textbf{Male}
 & {\tikzmarknode{c-mnom}{\der}} & \den & \dem & \des \\
\textbf{Neutral}
 & \das & \das & \dem & \des \\
\textbf{Female}
 & {\tikzmarknode{c-fnom}{\die}} &  {\tikzmarknode{c-facc}{\die}}
 & {\tikzmarknode{c-fdat}{\der}}
 & {\tikzmarknode{c-fgen}{\der}} \\
\bottomrule \\
& & 
& {\tikzmarknode{c-fdatdie}{\die}} & {\tikzmarknode{c-fgendie}{\die}}
\end{tabular}
};

\node[boxsrc, fit=(c-mnom), inner sep=1.5pt] {};
\node[boxtgt, fit=(c-fnom), inner sep=1.5pt] {};
\draw[arr] (c-mnom) -- (c-fnom);

\node[boxsrc, fit=(c-fdat), inner sep=1.5pt] {};
\node[boxsrc, fit=(c-fgen), inner sep=1.5pt] {};

\node[boxtgt, fit=(c-fdatdie), inner sep=1.5pt] {};
\node[boxtgt, fit=(c-fgendie), inner sep=1.5pt] {};

\draw[arr] (c-fdat) -- (c-fdatdie);
\draw[arr] (c-fgen) -- (c-fgendie);
\end{tikzpicture}
\caption{Spillover (SO).}
\end{subfigure}
\vspace{-0.3em}
\caption{Patterns of generalizations, exemplified using \gradcNgMF\ ($der\,{\to}\,die$).}
\label{fig:running-example-cases}
\vspace{-0.5em}
\end{figure*}

\textbf{Identity pairs.}
By construction, identity-pair gradients (red/orange/purple) map to $h \approx 0$, which is clearly observed for German encoder-only models.
\eurobert\ exhibits larger deviations: while most identity pairs remain centered near zero, gradients involving articles from the trained \gradiend\ variant shift toward the same signed encoding as the targeted transition, aligning with the sign of the target article.
A plausible explanation is that multilingual representations encode German article distinctions less sharply, so gradients for factual predictions may still share task-relevant information in the article prediction task.
Decoder-only models show a similar pattern and additionally spread identity pairs involving other articles across much of $[-1,1]$, likely reflecting variance introduced by our lightweight article classification head.

\textbf{Neutral control.}
Finally, gradients on \dataNEUT\  (where no articles are masked by construction) map consistently close to zero across models and configurations (yellow), providing a sanity check.

\begin{aclbox}
\gradiend\ learns a meaningful scalar representation that robustly separates transition directions and often generalizes across gender-case pairs of the same surface article transition.
\end{aclbox}

\subsection{Intervention Effects on Articles}\label{sec:intervention-article-probabilities}

Next, we evaluate how \gradiend\ updates affect article probabilities and how these effects distribute across gender-case cells.
Examining where probability changes occur is central to our analysis, since different internal mechanisms imply different patterns of generalization.
A grammar-tracking mechanism should yield either
\begin{enumerate*}[label=(\roman*)]
    \item \emph{local rule-based (LR)} effects restricted to the trained cell, or
    \item \emph{generalized rule-based (GR)} effects that  propagate systematically to grammatically related cells (e.g., along gender while preserving case).
\end{enumerate*}
In contrast, surface-level behavior can yield \emph{spillover (SO)}, where the same surface transition (e.g., $der\,{\to}\,die$) appears in grammatically unrelated cells that share the same source article.
Figure~\ref{fig:running-example-cases} illustrates these patterns.

\textbf{Selected article transitions.}
To focus on the most diagnostic settings, 
we restrict this analysis to  article groups containing gender and case transitions, enabling evaluation of a trained \gradiend\  along the other dimension (Table~\ref{tab:article-transitions}).
Such evaluations are possible within each two-dimensional transition group. 
For example, we assess the impact of \gradcNgMF\ $der\,{\to}\,die$ on \dataDF\ and \dataGF, since these datasets share the source article \textit{der}.

\begin{figure}[!t]
    \centering
    \includegraphics[width=\linewidth]{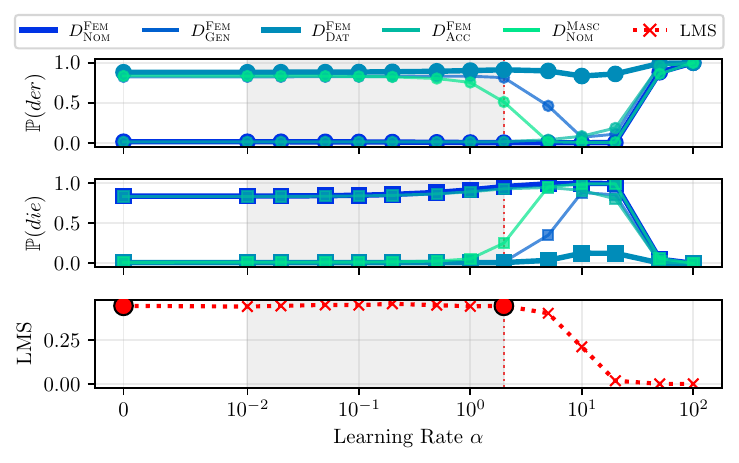}
     \vspace{-2em}
    \caption{\gradcNDgF\ applied to \bert\ for the $der\,{\to}\,die$ transition: mean article probabilities and LMS across learning rates $\alpha$. The candidate range ($\alpha>0$ and before the $99\%$ LMS drop) is shaded gray. Highlighted LMS points mark the base model (left) and $\alpha^\star$ (maximizing $\mathbb{P}(der)$ on \dataDF\ in gray area).}
    \label{fig:model-modification}
\end{figure}

\begin{table*}[!t]
    \centering
    \fontsize{8}{9}\selectfont
    \setlength{\tabcolsep}{4pt}
    \renewcommand{\arraystretch}{1.05}
    \begin{tabular}{llrrrrrrrrrrrrrrrrr}\toprule
    & & & \multicolumn{3}{c}{\textbf{$\dataNM(der)$}} & \multicolumn{3}{c}{\textbf{$\dataGF(der)$}} & \multicolumn{3}{c}{\textbf{$\dataDF(der)$}} & 
    \multicolumn{3}{c}{\textbf{\dataNEUT}} &
    \\ \cmidrule(lr){4-6} \cmidrule(lr){7-9} \cmidrule(lr){10-12} \cmidrule(lr){13-15}
  \textbf{Model} & \textbf{Art. Trans.} & $\alpha$ &  
   $\Delta \mathbb{P}$ & $d$ & \textbf{Sig.}  &  
   $\Delta \mathbb{P}$ & $d$ & \textbf{Sig.}   &  
   $\Delta \mathbb{P}$ & $d$ & \textbf{Sig.}   &  
   $\Delta \mathbb{P}$ & $d$ & \textbf{Sig.}   & \textbf{\acrshort{supergleber}} \\ \midrule
   
        \bert & -- & 0.0 &  -- & -- & -- & -- & -- & -- & -- & -- & -- & -- & -- & -- &  $70.7 \pm 0.4$  \\
\, + \gradcNgMF & $der \to die$ & 0.01 & \textbf{0.04} & \textbf{0.30} & \textbf{***} & 0.00 & 0.12 & *** & 0.01 & 0.16 & *** & 0.05 & 0.01 & n.s. & $70.1 \pm 0.4$  \\
\, + \gradcADgF & $der \to die$ & 0.1 & 0.05 & 0.25 & *** & 0.04 & 0.16 & *** & \textbf{0.33} & \textbf{0.24} & \textbf{***} & 0.09 & 0.02 & n.s. & $70.2 \pm 0.4$  \\
\, + \gradcGAgF & $der \to die$ & 0.5 & 0.66 & 0.32 & *** & \textbf{1.85} & \textbf{0.47} & \textbf{***} & 1.40 & 0.37 & *** & 0.18 & 0.03 & ** & $70.2 \pm 0.4$  \\
\, + \gradcNDgM & $der \to dem$ & 0.05 & \textbf{0.04} & \textbf{0.16} & \textbf{***} & 0.00 & 0.08 & *** & 0.02 & 0.13 & *** & 0.02 & 0.01 & n.s. & $70.1 \pm 0.4$  \\
\, + \gradcDgFN & $der \to dem$ & 0.01 & 0.00 & 0.11 & *** & 0.00 & 0.08 & *** & \textbf{0.01} & \textbf{0.17} & \textbf{***} & 0.03 & 0.02 & n.s. & $70.1 \pm 0.4$  \\
\, + \gradcNGgM & $der \to des$ & 0.2 & \textbf{0.14} & \textbf{0.15} & \textbf{***} & 0.28 & 0.35 & *** & 0.02 & 0.04 & *** & 0.01 & 0.01 & n.s. & $70.2 \pm 0.4$  \\
\, + \gradcGgFN & $der \to des$ & 0.5 & 0.25 & 0.12 & *** & \textbf{3.70} & \textbf{0.55} & \textbf{***} & 0.08 & 0.07 & *** & 0.07 & 0.05 & *** &  $70.2 \pm 0.4$  \\
         \bottomrule
    \end{tabular}
    \caption{\gradiend-modified  \bert\ models (others in Table~\ref{tab:prob-changes-other-models}): $\Delta\mathbb{P}$ of target article (scaled by 100), Cohen's $d$, and significance as *** ($p<.001$, ** $p<.01$, * $p<.05$; n.s. otherwise). 
    Bold marks corresponding \gradiend\ datasets. \acrshort{supergleber} score (scaled by 100) use bootstrapped $95\%$ confidence intervals ($n\,{=}\,1000$).
    }
    \label{tab:prob-changes-bert}
\end{table*}

\begin{figure*}[!t]
    \centering
    \includegraphics[width=0.925\linewidth,trim=0.1cm 0.2cm 0.1cm 0.1cm, 
        clip]{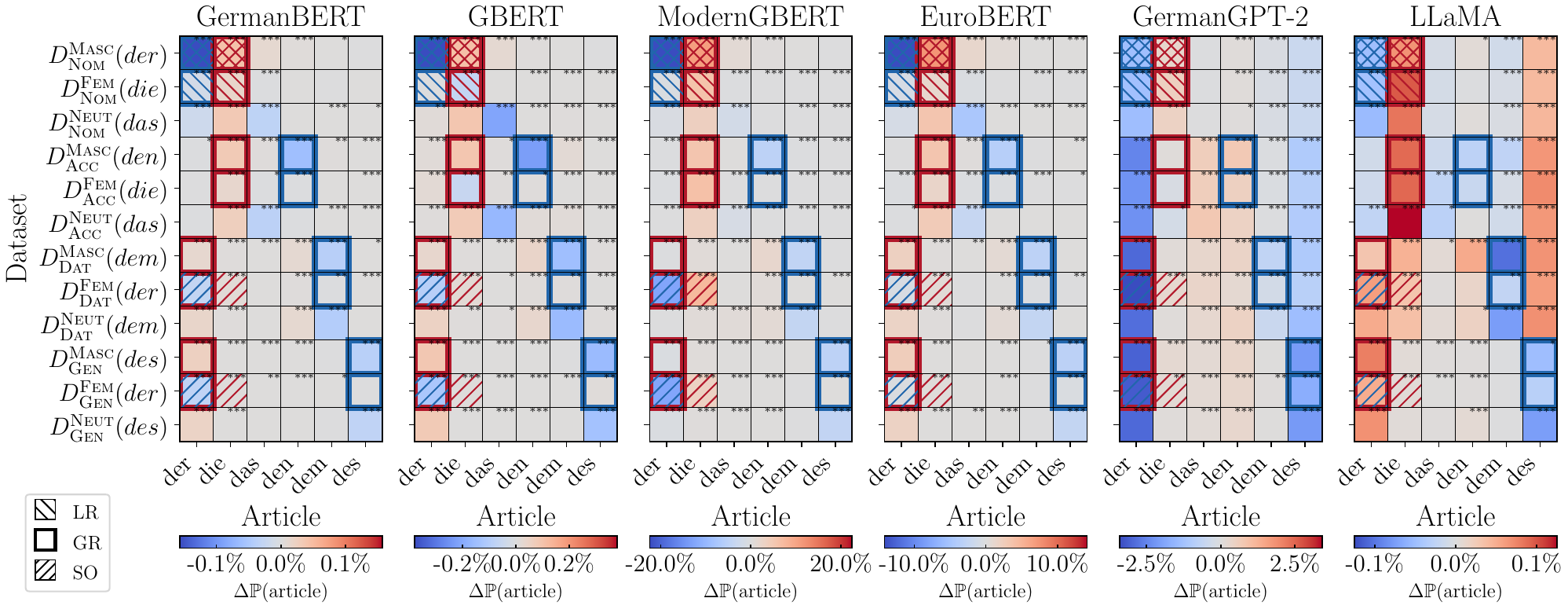}
        \vspace{-0.1em}
    \caption{Mean probability change of articles between \gradiend-modified and base model for \gradcNgMF\ $der\,{\to}\,die$ (others in Figures~\ref{fig:heatmap-AD_F}-\ref{fig:heatmap-G_FN}). 
   Stars mark statistical significance after Benjamini-Hochberg FDR correction \cite{benjamin-hochberg-correction} applied per model. 
    Marked cells are expectations for LR, GR, and SO (Figure~\ref{fig:running-example-cases}). 
    }
    \label{fig:bert-probability-heatmap-N-MF}
\end{figure*}

\textbf{Intervention strength and $\alpha$ selection.}
Large updates can change predictions by degrading language modeling~\cite{gradiend}.
Figure~\ref{fig:model-modification} shows this trade-off for \bert\ under \gradcNDgF\ $der\,{\to}\,die$: as $\alpha$ increases, $\mathbb{P}(\textit{die})$ rises (until a certain point) while a \gls{lms} measured on \dataNEUT\ drops.
We therefore analyze probability shifts only under an explicit \gls{lms}-preservation constraint.
Concretely, we apply scaled decoder updates
$\widetilde{W}_m = W_m \,{+}\, \alpha \cdot \dec(h^\star)$, where $h^\star=\pm1$ selects the transition direction, and evaluate a grid of $\alpha>0$ values.
We retain only candidates that preserve at least $99\%$ of the base-model \gls{lms} on \dataNEUT\ (masked-token accuracy for encoder-only models, perplexity for decoder-only models), and choose $\alpha^\star$ as the candidate that maximizes the mean probability of the target article on the corresponding target-article dataset (e.g., \dataDF\ for \gradcNDgF\ $der\,{\to}\,die$).
The candidate range and $\alpha^\star$ are highlighted in Figure~\ref{fig:model-modification}. Details are in Appendix~\ref{app:alpha-selection}.

\textbf{Probability evaluation.}
For each gender-case dataset, we compute the mean article-probability change $\Delta \mathbb{P}(art)$ of the \gradiend-modified model relative to the base (positive indicates an increase).
We report Cohen's $d$ with pooled variance \cite{cohen1988power} and test significance with a permutation test \cite{good2005permutation}.
Table~\ref{tab:prob-changes-bert} shows representative results for \bert\ (other models in Table~\ref{tab:prob-changes-other-models}).

\textbf{Effects occur before broad degradation.}
Across transitions, \gradiend-updates induce significant shifts on article datasets while changes on \dataNEUT\ are mostly non-significant.
Because $\alpha^\star$ is chosen conservatively, 
$\Delta \mathbb{P}$ is typically below $1\%$, but effect sizes and significance indicate consistent directional shifts.
Effects are usually strongest on the trained cell, yet remain substantially larger on other article datasets sharing the same source article than on \dataNEUT, suggesting the changes are not due to broad degradation.
This is further supported by mostly unchanged \acrshort{supergleber} scores, a German NLP benchmark consisting of 29 tasks  \cite{supergleberr}. 

\textbf{Effects on all cells.}
Figure~\ref{fig:bert-probability-heatmap-N-MF} visualizes $\Delta \mathbb{P}$ over the full gender-case grid for the  \gradcNgMF\ ($der\,{\to}\,die$) across all base models as heatmap.
We additionally overlay the three patterns of generalization from Figure~\ref{fig:running-example-cases}.
The heatmap partially matches GR,  but with deviations: some  GR-predicted cells are neutral or opposite-signed (e.g., \bert\ \dataDF\ for $der$), and several effects appear in cells that are not predicted by a clean grammar-preserving rule.
Notably, the two clearest GR contradictions in Figure~\ref{fig:bert-probability-heatmap-N-MF}, $\mathbb{P}(der)$ on \dataDF\ and \dataGF, align with SO, which explains most of its predicted cells in terms of probability-change direction.
Decoder-only models show more  deviations, probably due to the custom small MLM head. 
\llama\ is the only model not showing the $\mathbb{P}(der)$ increase on \dataDF/\dataGF, 
possibly indicating a trend of less memorization in larger models.
Across models, cells sharing a surface article with the same gender \emph{or} case (e.g., \dataDM/\dataDN\ and \dataGM/\dataGN) often behave similarly, indicating transitive spillover. 
For example, \gradcNgMF\ $die\,{\to}\,der$  increases $\mathbb{P}(des)$ on \dataGM\ (GR consistent), and concurrently on \dataGN. 



\begin{aclbox}
Probability-shift patterns are neither fully rule-based nor consistent with unrestricted spillover, but instead suggest a mixture of structured generalization and surface-article-linked effects.
\end{aclbox}

\begin{figure*}[!t]
    \centering
    \vspace{-0.1em}
    \includegraphics[width=0.18\linewidth]{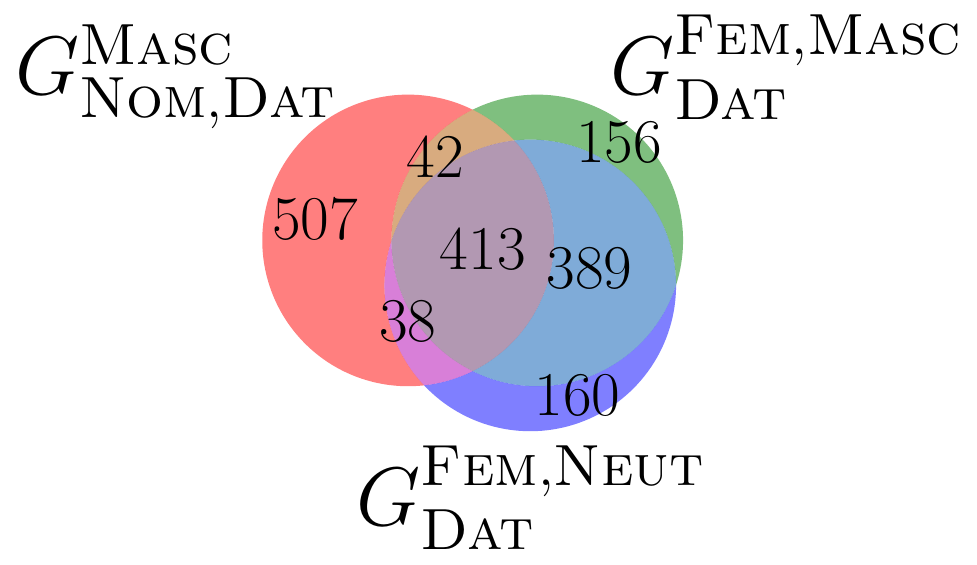}
    \includegraphics[width=0.18\linewidth]{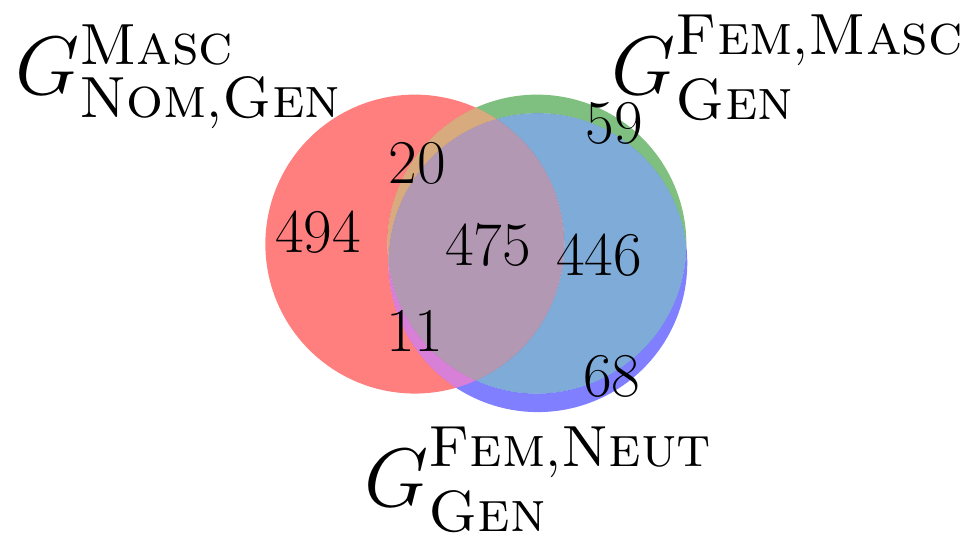}
    \includegraphics[width=0.15\linewidth]{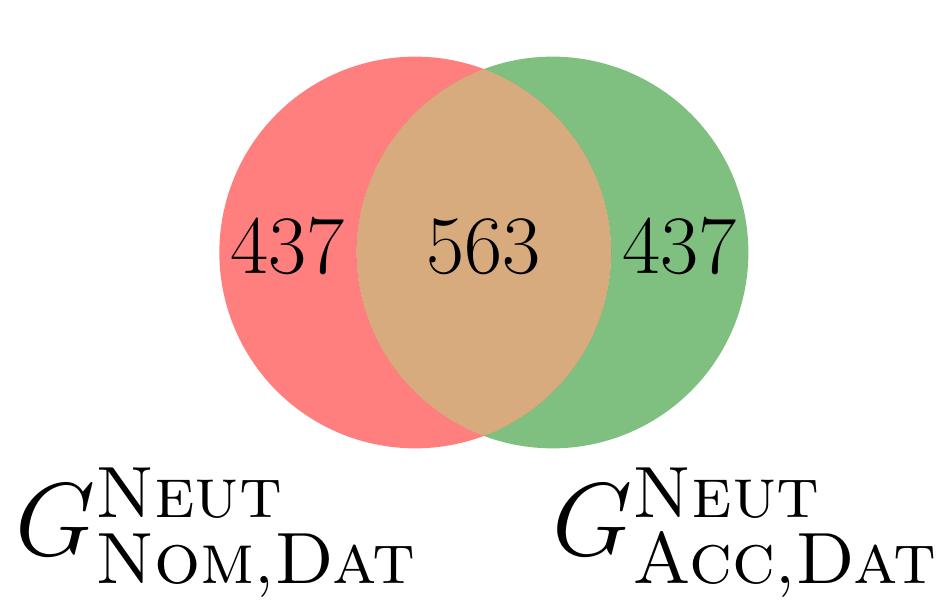}
    \includegraphics[width=0.16\linewidth]{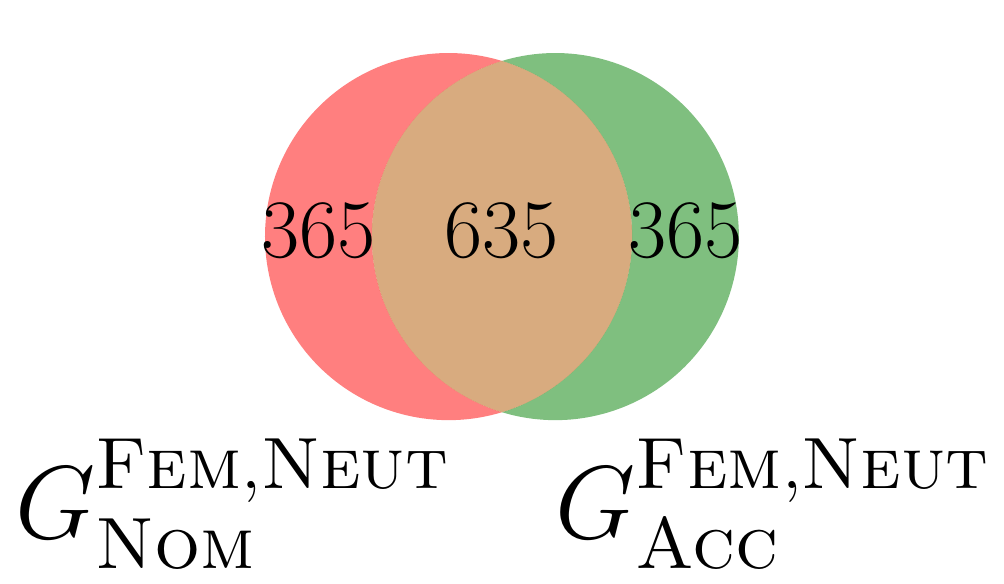}
    \includegraphics[width=0.13\linewidth]{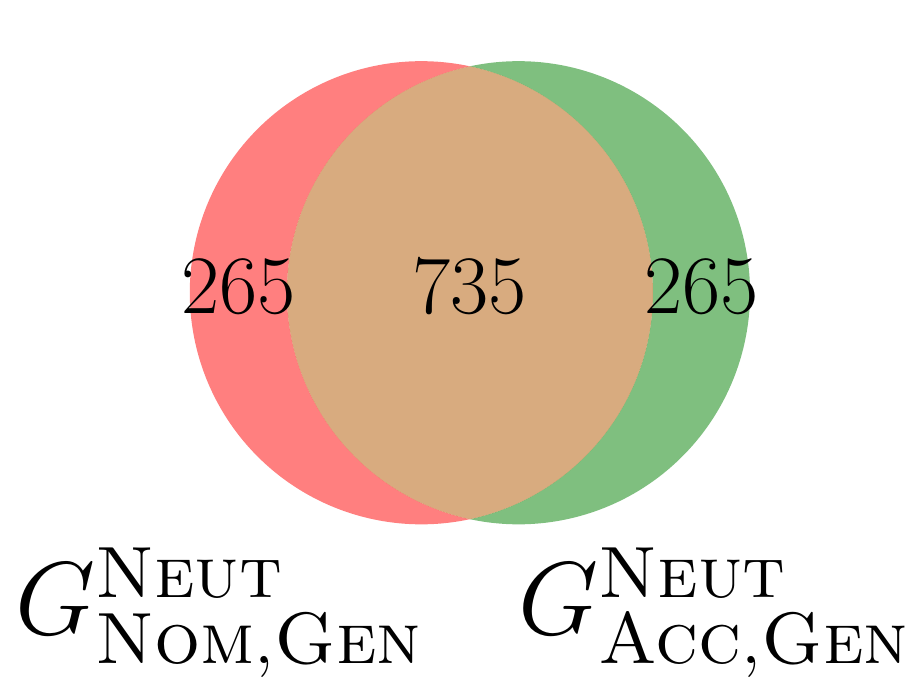}
\revbg{
  \includegraphics[width=0.12\linewidth]{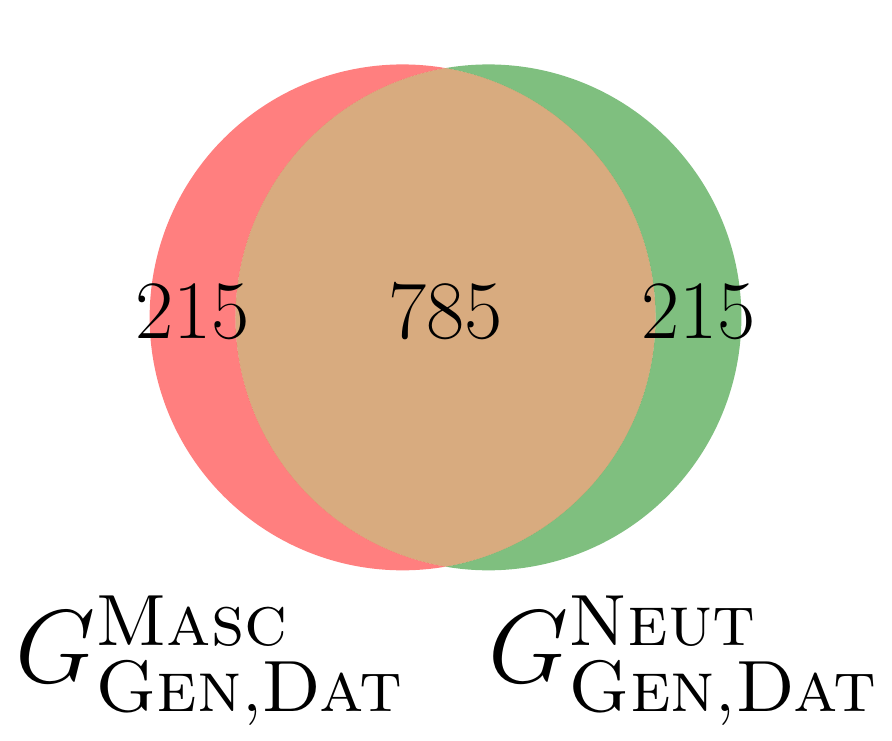}
}
    \caption{Top-$1,000$ weight overlaps across different \glspl{gradiend} for \bert\ (other models in Figure~\ref{fig:venn-other-models-weight}).}  
    \label{fig:venn-bert-weights}
\end{figure*}

\begin{figure*}[!t]
\centering

\begin{minipage}[t]{0.48\textwidth}
    \centering
    \includegraphics[width=\textwidth,
        trim=0cm 1.0cm 0cm 1.0cm,
        clip]{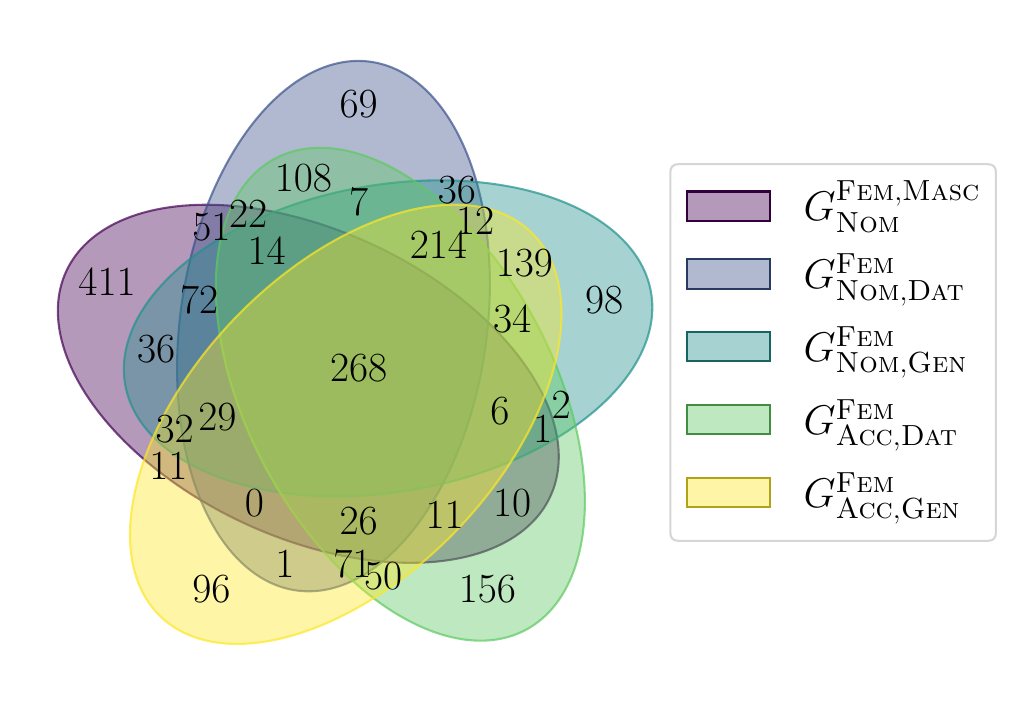}
    \caption{Top-$1,000$ weight overlaps for \bert\ $der\,{\leftrightarrow}\,die$ article group (other models in Figure~\ref{fig:venn-der-die-other-models-weight}).}
    \label{fig:venn-der-die-bert-weight}
\end{minipage}
\hfill
\begin{minipage}[t]{0.48\textwidth}
    \centering
    \includegraphics[width=\textwidth,
        trim=0cm 1.85cm 0cm 2.9cm,
        clip]{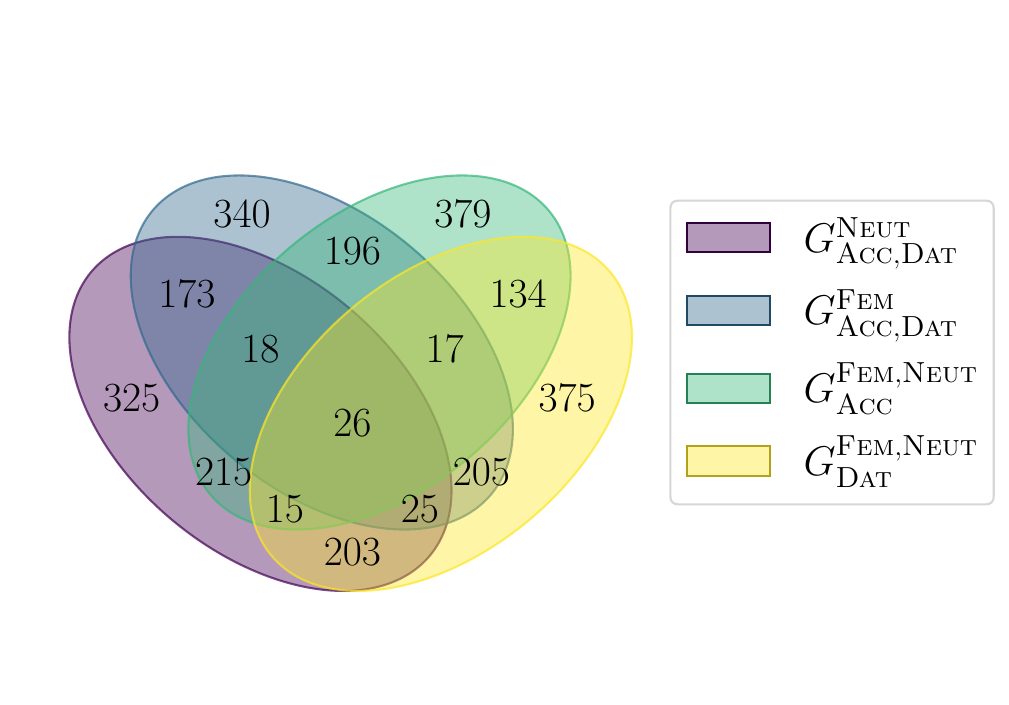}
    \caption{Top-$1,000$ weight overlaps for the \bert\ control group (other models in Figure~\ref{fig:venn-control-other-models-weight}).}
    \label{fig:venn-control-group-bert-weight}
\end{minipage}
\end{figure*}

\subsection{Overlap of Most Affected Parameters}\label{sec:overlap-topk}

Finally, we compare \gradiend\ models directly in the parameter space.
We define parameter importance by the absolute value of decoder weights $W_d$ and extract the Top-$k$ weights for $k=1000$.

\textbf{Overlap within article groups.}
Figures~\ref{fig:venn-bert-weights} and~\ref{fig:venn-der-die-bert-weight} show article group Venn diagrams for \bert\ (additional models are reported in Appendix~\ref{app:topk}).
Across base models, Top-$k$ weights overlap substantially across \glspl{gradiend} within an article group.
For instance, overlap in the $der\!\leftrightarrow\!die$ group remains high even between variants trained on different case/gender axes (e.g., \gradcNgMF\ vs.\ \gradcNDgF\ and \gradcNGgF), suggesting a shared subset of weights despite differing grammatical contexts.

\textbf{Control group.}
To test whether overlap is expected in general, we analyze a control group of four \gradiend\ variants spanning \accusative/\dative\ and \female/\neutral\ whose cells realize disjoint surface articles.
Figure~\ref{fig:venn-control-group-bert-weight} shows the Top-$k$ overlap for \bert, which is smaller than in the article groups, indicating that the high overlap in Figures~\ref{fig:venn-bert-weights} and \ref{fig:venn-der-die-bert-weight} is not a generic artifact.

\textbf{Quantifying overlap.}
Table~\ref{tab:overlap} quantifies these observations.
For each article group (including the control), we report the maximum pairwise Top-$k$ overlap, 
$
\max_{A,B} \frac{|A{\cap} B|}{k},
$
where $A$ and $B$ are the Top-$k$  sets of two variants.
Groups with at least two variants realizing the same surface article pair show consistently high overlap (${>}\,75\%$ on average), whereas the control group is much lower (mean $38.9\%$). 
This suggests that gender-case transitions of the same articles rely on a shared subset of parameters rather than disjoint, transition-specific mechanisms.

\begin{aclbox}
Top-$k$ weights overlap strongly within article groups and far less in the disjoint-article control groups, suggesting shared parameters for the same article transitions.
\end{aclbox}

\begin{table}[!t]
    \centering
   \fontsize{8}{9}\selectfont
\setlength{\tabcolsep}{3pt}
    \begin{tabular}{lrrrrrrrr}\toprule
       \textbf{Model} &
       \multicolumn{1}{c}{\rotatebox{90}{\textbf{$der\!\leftrightarrow\!die$}}} &
      \multicolumn{1}{c}{ \rotatebox{90}{$der\!\leftrightarrow\!dem$}} &
       \multicolumn{1}{c}{\rotatebox{90}{$der\!\leftrightarrow\!des$}} &
       \multicolumn{1}{c}{\rotatebox{90}{$das\!\leftrightarrow\!die$}} &
       \multicolumn{1}{c}{\rotatebox{90}{$das\!\leftrightarrow\!das$}} &
       \multicolumn{1}{c}{\rotatebox{90}{$das\!\leftrightarrow\!des$}} &
     \multicolumn{1}{c}{\rotatebox{90}{\rev{$dem\!\leftrightarrow\!des$}}} &
      \multicolumn{1}{c}{ \rotatebox{90}{Control}}
       \\ \midrule
        \bert & 73.4 & 92.1 & 80.2 & 63.5 & 56.3 & 73.5 & \rev{78.5} & 27.4 \\
        \gbert & 77.4 & 86.4 & 76.8 & 64.2 & 81.1 & 61.3 & \rev{69.5} & 27.5 \\
        \modernbert & 87.2 & 89.7 & 70.9 & 80.7 & 70.0 & 68.5 & \rev{66.2} & 27.5 \\
        \eurobert & 83.7 & 95.3 & 88.1 & 71.6 & 78.3 & 78.0 & \rev{85.4} & 51.2 \\
        \gpttwo & 90.8 & 93.9 & 86.5 & 85.4 & 86.8 & 83.7 & \rev{95.7} & 51.7 \\
        \llama & 91.7 & 82.1 & 91.5 & 88.1 & 83.6 & 85.3 & \rev{87.2} & 48.2 \\ \midrule
       Mean & 84.0 & 89.9 & 82.3 & 75.6 & 76.0 & 75.1 & \rev{80.4} & 38.9 \\
         \bottomrule
    \end{tabular}
   \caption{Maximum pairwise \rev{Top-1,000} weight overlap (scaled by 100) for article groups including the control group.}
    \label{tab:overlap}
\end{table}

\subsection{Discussion}

Taken together, our analyses show that German definite article behavior is not fully explained by a uniformly rule-based mechanism, providing evidence against \ref{hyp:ruleencoding}.
First, the encoding analysis shows that the learned bottleneck value $h$ reliably separates the two swapped training cells, yet often assigns similar encodings to gradients from other gender-case transitions with the same articles, suggesting limited cell-specific disentanglement.
Second, interventions shift article probabilities beyond the trained cell 
and only partially follow clean gender-/case-preserving generalization, while  
exhibiting spillover.
We also observe a tentative size trend: larger models (\llama) show less spillover (as also suggested by the encoded value distributions, e.g., Figure~\ref{fig:encoded-all-models-N_MF}).
Third, the Top-$k$ overlap analysis reveals substantial intersections of the most affected weights across variants within an article group, with smaller overlaps in a control group with disjoint surface articles, indicating that multiple transitions rely on a shared parameter subspace.

\begin{aclbox}
Our results provide evidence that a purely rule-based account (\ref{hyp:ruleencoding}) is unlikely to be sufficient, and suggest that the memorization hypothesis (\ref{hyp:memorization}) holds at least to some extent.
\end{aclbox}

\section{Conclusion}

We studied whether German definite singular articles in \glspl{lm} reflect abstract rule encoding (\ref{hyp:ruleencoding}) or surface-level memorization (\ref{hyp:memorization}).
Using \gradiend\ across multiple gender-case cells, 
we find that article transition updates shift article probabilities  significantly beyond the trained cell under a \gls{lms} constraint and only partially follow rule-based generalization.
The results are not consistent with a purely rule-based encoding of the grammar and suggest that memorization-like mechanisms are important. 
This suggests that while \glspl{lm} can be used reliably to assess if text is grammatical and to produce grammatical text, they should be used with care 
\rev{when using them to analyze grammatical systems, since rules might not be encoded as expected.}

We release our code and datasets: \iflink\url{https://github.com/aieng-lab/gradiend-german-articles}.\else anonymous.\fi

\section{Limitations}

Our study has several limitations that constrain the scope of the conclusions.

First, we focus exclusively on \textbf{German definite singular articles}, a small and highly regular closed-class system. While this makes the analysis controlled and interpretable, the findings may not transfer to other morphosyntactic phenomena (e.g., adjective agreement, verb inflection, or freer word order) or to other languages, where grammatical cues are distributed differently. However, the lack of a strict, rule-based encoding of such a regular, closed-class system indicates that more complex systems are also not learned through rules, but rather memorized. 

Second, our conclusions rely on \textbf{gradient-based interventions} using \gradiend.
Although the applied update is dense and affects (in principle) all parameters, it is restricted to a single update direction scaled by a scalar $\alpha$.
Thus, our interventions primarily reveal mechanisms that can be expressed as a coherent, scalable update direction. More distributed or highly context-specific rule implementations may not be captured by this probe.

Third, the measured \textbf{intervention effects are small by design}. Because we select $\alpha^\star$ under a strict \gls{lms}-preservation criterion, mean probability shifts are typically below $1\%$. While effect sizes and significance indicate consistent directional changes, small magnitudes make it harder to judge downstream behavioral impact and may understate the extent of generalization that would appear under less conservative constraints.

Fourth, we evaluate only a \textbf{limited model scale range} (up to $\sim$3B parameters). Larger models or models trained with substantially different data mixtures or objectives may encode grammatical information differently, and scaling trends cannot be confidently concluded from our setup. Nevertheless, we highlight that all models we analyzed consistently yielded results consistent with our memorization hypothesis, indicating that this is a general pattern.

Fifth, we rely on \textbf{spaCy-based annotation} to construct gender-case datasets. While effective at scale, automatic annotation can introduce noise, especially in ambiguous or syntactically complex sentences, which may affect gradient estimates and significance patterns. \rev{To estimate labeling quality, we manually annotated 20 randomly sampled instances per gender-case dataset and obtain an overall accuracy of 82\% (Appendix~\ref{app:spacy-quality}). Many mis-classifications are not for other definite articles and are rather other errors that should only introduce random noise, e.g., mislabelling the relative pronoun \textit{der} as an article. Errors that mix up definite singular articles are only commonly occurring for \textit{des}, but there is no visible different trend when restricting to transitions and dataset intersections with near-perfect labeling, suggesting that spaCy noise is unlikely to impact our findings. This noise-tolerance of \gradiend\ is further supported by earlier results, where noisy features for race and religion could also be identified by \gradiend~\cite{gradiend}.}

Sixth, decoder-only models are evaluated using a \textbf{custom MLM-style head} to enable bidirectional conditioning. This departs from their native training objective and may influence gradient structure and intervention behavior, limiting direct comparability with encoder-only models. However, the general consistence of the results with the encoder-only models indicates a limited impact on our study of this, though we believe that this is -- at least partially -- responsible for different pattern for neutral results that the MLM heads are not specifically trained for. 

Finally, our evaluation focuses on \textbf{controlled probability shifts} and parameter overlap rather than downstream generation behavior.


\bibliography{custom}

@inproceedings{ferrando-costa-jussa-2024-similarity,
    title = "On the Similarity of Circuits across Languages: a Case Study on the Subject-verb Agreement Task",
    author = "Ferrando, Javier  and
      Costa-juss{\`a}, Marta R.",
    editor = "Al-Onaizan, Yaser  and
      Bansal, Mohit  and
      Chen, Yun-Nung",
    booktitle = "Findings of the Association for Computational Linguistics: EMNLP 2024",
    month = nov,
    year = "2024",
    address = "Miami, Florida, USA",
    publisher = "Association for Computational Linguistics",
    url = "https://aclanthology.org/2024.findings-emnlp.591/",
    doi = "10.18653/v1/2024.findings-emnlp.591",
    pages = "10115--10125"
}

@article{attention-is-all-you-need,
  title={Attention is all you need},
  author={Vaswani, Ashish and Shazeer, Noam and Parmar, Niki and Uszkoreit, Jakob and Jones, Llion and Gomez, Aidan N and Kaiser, {\L}ukasz and Polosukhin, Illia},
  journal={Advances in neural information processing systems},
  volume={30},
  year={2017},
url={https://arxiv.org/abs/1706.03762}
}

@article{rogers-etal-2020-primer,
    title = "A Primer in {BERT}ology: What We Know About How {BERT} Works",
    author = "Rogers, Anna  and
      Kovaleva, Olga  and
      Rumshisky, Anna",
    editor = "Johnson, Mark  and
      Roark, Brian  and
      Nenkova, Ani",
    journal = "Transactions of the Association for Computational Linguistics",
    volume = "8",
    year = "2020",
    address = "Cambridge, MA",
    publisher = "MIT Press",
    url = "https://aclanthology.org/2020.tacl-1.54/",
    doi = "10.1162/tacl_a_00349",
    pages = "842--866"
}

@article{benjamin-hochberg-correction,
author = {Benjamini, Yoav and Hochberg, Yosef},
title = {Controlling the False Discovery Rate: A Practical and Powerful Approach to Multiple Testing},
journal = {Journal of the Royal Statistical Society: Series B (Methodological)},
volume = {57},
number = {1},
pages = {289-300},
doi = {https://doi.org/10.1111/j.2517-6161.1995.tb02031.x},
url = {https://rss.onlinelibrary.wiley.com/doi/abs/10.1111/j.2517-6161.1995.tb02031.x},
eprint = {https://rss.onlinelibrary.wiley.com/doi/pdf/10.1111/j.2517-6161.1995.tb02031.x},
year = {1995}
}

@online{wikipedia,
  author = {{Wikimedia Foundation}},
  title = {Wikimedia Wikipedia Dataset},
  howpublished = {\url{https://huggingface.co/datasets/openskyml/wikipedia}},
  note = {Version: “20220301.de”},  
  year = {2022},
  url = {"https://dumps.wikimedia.org"}
}

@misc{wortschatz_deu_news_2024_300K,
  author       = {{Leipzig Corpora Collection}},
  title        = {German News Corpus 2024 (300K subset)},
  year         = {2024},
  howpublished = {\url{https://downloads.wortschatz-leipzig.de/corpora/deu_news_2024_300K.tar.gz}},
  note         = {Leipzig Corpora Collection. Dataset. Accessed: 2025-12-18}
}

@inproceedings{goldhahn2012lcc,
  author    = {Goldhahn, Dirk and Eckart, Thomas and Quasthoff, Uwe},
  title     = {Building Large Monolingual Dictionaries at the Leipzig Corpora Collection: From 100 to 200 Languages},
  booktitle = {Proceedings of the Eighth International Conference on Language Resources and Evaluation (LREC 2012)},
  year      = {2012},
  address   = {Istanbul, Turkey},
  publisher = {European Language Resources Association (ELRA)},
  url       = {https://wortschatz.uni-leipzig.de/en/publications}
}

@article{grattafiori2024llama,
  title={The llama 3 herd of models},
  author={Grattafiori, Aaron and Dubey, Abhimanyu and Jauhri, Abhinav and Pandey, Abhinav and Kadian, Abhishek and Al-Dahle, Ahmad and Letman, Aiesha and Mathur, Akhil and Schelten, Alan and Vaughan, Alex and others},
  journal={arXiv preprint arXiv:2407.21783},
  year={2024},
url={https://arxiv.org/pdf/2407.21783}
}

@article{bert,
  author       = {Jacob Devlin and
                  Ming{-}Wei Chang and
                  Kenton Lee and
                  Kristina Toutanova},
  title        = {{BERT:} Pre-training of Deep Bidirectional Transformers for Language
                  Understanding},
  journal      = {CoRR},
  volume       = {abs/1810.04805},
  year         = {2018},
  url          = {http://arxiv.org/abs/1810.04805},
  eprinttype    = {arXiv},
  eprint       = {1810.04805},
  bibsource    = {dblp computer science bibliography, https://dblp.org}
}

@article{gpt,
  title={Language models are unsupervised multitask learners},
  author={Radford, Alec and Wu, Jeffrey and Child, Rewon and Luan, David and Amodei, Dario and Sutskever, Ilya and others},
  journal={OpenAI blog},
  volume={1},
  number={8},
  pages={9},
  year={2019},
  url={https://cdn.openai.com/better-language-models/language_models_are_unsupervised_multitask_learners.pdf}
}

@inproceedings{supergleberr,
    title = "{S}uper{GLEB}er: {G}erman Language Understanding Evaluation Benchmark",
    author = "Pfister, Jan  and
      Hotho, Andreas",
    editor = "Duh, Kevin  and
      Gomez, Helena  and
      Bethard, Steven",
    booktitle = "Proceedings of the 2024 Conference of the North American Chapter of the Association for Computational Linguistics: Human Language Technologies (Volume 1: Long Papers)",
    month = jun,
    year = "2024",
    address = "Mexico City, Mexico",
    publisher = "Association for Computational Linguistics",
    url = "https://aclanthology.org/2024.naacl-long.438/",
    doi = "10.18653/v1/2024.naacl-long.438",
    pages = "7904--7923"
}

@inproceedings{
eurobert,
title={Euro{BERT}: Scaling Multilingual Encoders for European Languages},
author={Nicolas Boizard and Hippolyte Gisserot-Boukhlef and Duarte Miguel Alves and Andre Martins and Ayoub Hammal and Caio Corro and CELINE HUDELOT and Emmanuel Malherbe and Etienne Malaboeuf and Fanny Jourdan and Gabriel Hautreux and Jo{\~a}o Alves and Kevin El Haddad and Manuel Faysse and Maxime Peyrard and Nuno M Guerreiro and Patrick Fernandes and Ricardo Rei and Pierre Colombo},
booktitle={Second Conference on Language Modeling},
year={2025},
url={https://openreview.net/forum?id=jdOC24msVq}
}

@misc{moderngbert,
      title={New Encoders for German Trained from Scratch: Comparing ModernGBERT with Converted LLM2Vec Models}, 
      author={Julia Wunderle and Anton Ehrmanntraut and Jan Pfister and Fotis Jannidis and Andreas Hotho},
      year={2025},
      eprint={2505.13136},
      archivePrefix={arXiv},
      primaryClass={cs.CL},
      url={https://arxiv.org/abs/2505.13136}, 
}

@inproceedings{gbert,
    title = "{G}erman{'}s Next Language Model",
    author = {Chan, Branden  and
      Schweter, Stefan  and
      M{\"o}ller, Timo},
    editor = "Scott, Donia  and
      Bel, Nuria  and
      Zong, Chengqing",
    booktitle = "Proceedings of the 28th International Conference on Computational Linguistics",
    month = dec,
    year = "2020",
    address = "Barcelona, Spain (Online)",
    publisher = "International Committee on Computational Linguistics",
    url = "https://aclanthology.org/2020.coling-main.598/",
    doi = "10.18653/v1/2020.coling-main.598",
    pages = "6788--6796"
}

@article{belinkov-2022-probing,
    title = "Probing Classifiers: Promises, Shortcomings, and Advances",
    author = "Belinkov, Yonatan",
    journal = "Computational Linguistics",
    volume = "48",
    number = "1",
    month = mar,
    year = "2022",
    address = "Cambridge, MA",
    publisher = "MIT Press",
    url = "https://aclanthology.org/2022.cl-1.7/",
    doi = "10.1162/coli_a_00422",
    pages = "207--219"
}

@inproceedings{gonen-etal-2019-grammatical-gender,
    title = "How does Grammatical Gender Affect Noun Representations in Gender-Marking Languages?",
    author = "Gonen, Hila  and
      Kementchedjhieva, Yova  and
      Goldberg, Yoav",
    editor = "Axelrod, Amittai  and
      Yang, Diyi  and
      Cunha, Rossana  and
      Shaikh, Samira  and
      Waseem, Zeerak",
    booktitle = "Proceedings of the 2019 Workshop on Widening NLP",
    month = aug,
    year = "2019",
    address = "Florence, Italy",
    publisher = "Association for Computational Linguistics",
    url = "https://aclanthology.org/W19-3622/",
    pages = "64--67"
}

@inproceedings{carlini2022quantifying,
  title={Quantifying Memorization Across Neural Language Models},
  author={Carlini, Nicholas and Ippolito, Daphne and Jagielski, Matthew and Lee, Katherine and Tram{\`e}r, Florian and Zhang, Chiyuan},
  booktitle={Proceedings of the 11th International Conference on Learning Representations (ICLR 2023)},
  year={2023},
  url={https://arxiv.org/pdf/2202.07646}
}

@article{belinkovGlass2019,
    author = {Belinkov, Yonatan and Glass, James},
    title = {Analysis Methods in Neural Language Processing: A Survey},
    journal = {Transactions of the Association for Computational Linguistics},
    volume = {7},
    pages = {49-72},
    year = {2019},
    month = {04},
    issn = {2307-387X},
    doi = {10.1162/tacl_a_00254},
    url = {https://doi.org/10.1162/tacl_a_00254},
    eprint = {https://direct.mit.edu/tacl/article-pdf/doi/10.1162/tacl_a_00254/1923061/tacl_a_00254.pdf},
}

@book{cohen1988power,
  title     = {Statistical Power Analysis for the Behavioral Sciences},
  author    = {Cohen, Jacob},
  year      = {1988},
  edition   = {2},
  publisher = {Lawrence Erlbaum Associates},
  address   = {Hillsdale, NJ},
url={https://utstat.utoronto.ca/~brunner/oldclass/378f16/readings/CohenPower.pdf}
}

@book{good2005permutation,
  title     = {Permutation, Parametric, and Bootstrap Tests of Hypotheses},
  author    = {Good, Phillip I.},
  year      = {2005},
  edition   = {3},
  publisher = {Springer},
  address   = {New York},
url={https://tewaharoa.victoria.ac.nz/discovery/fulldisplay/alma99178047438702386/64VUW_INST:VUWNUI}
}

@article{seeker-kuhn-2013-morphological,
    title = "Morphological and Syntactic Case in Statistical Dependency Parsing",
    author = "Seeker, Wolfgang  and
      Kuhn, Jonas",
    journal = "Computational Linguistics",
    volume = "39",
    number = "1",
    month = mar,
    year = "2013",
    address = "Cambridge, MA",
    publisher = "MIT Press",
    url = "https://aclanthology.org/J13-1004/",
    doi = "10.1162/COLI_a_00134",
    pages = "23--55"
}

@article{lindsey2025biology,
  author={Lindsey, Jack and Gurnee, Wes and Ameisen, Emmanuel and Chen, Brian and Pearce, Adam and Turner, Nicholas L. and Citro, Craig and Abrahams, David and Carter, Shan and Hosmer, Basil and Marcus, Jonathan and Sklar, Michael and Templeton, Adly and Bricken, Trenton and McDougall, Callum and Cunningham, Hoagy and Henighan, Thomas and Jermyn, Adam and Jones, Andy and Persic, Andrew and Qi, Zhenyi and Thompson, T. Ben and Zimmerman, Sam and Rivoire, Kelley and Conerly, Thomas and Olah, Chris and Batson, Joshua},
  title={On the Biology of a Large Language Model},
  journal={Transformer Circuits Thread},
  year={2025},
  url={https://transformer-circuits.pub/2025/attribution-graphs/biology.html}
}

@article{linzen-etal-2016-assessing,
    title = "Assessing the Ability of {LSTM}s to Learn Syntax-Sensitive Dependencies",
    author = "Linzen, Tal  and
      Dupoux, Emmanuel  and
      Goldberg, Yoav",
    editor = "Lee, Lillian  and
      Johnson, Mark  and
      Toutanova, Kristina",
    journal = "Transactions of the Association for Computational Linguistics",
    volume = "4",
    year = "2016",
    address = "Cambridge, MA",
    publisher = "MIT Press",
    url = "https://aclanthology.org/Q16-1037/",
    doi = "10.1162/tacl_a_00115",
    pages = "521--535"
}

@inproceedings{marvin-linzen-2018-targeted,
    title = "Targeted Syntactic Evaluation of Language Models",
    author = "Marvin, Rebecca  and
      Linzen, Tal",
    editor = "Riloff, Ellen  and
      Chiang, David  and
      Hockenmaier, Julia  and
      Tsujii, Jun{'}ichi",
    booktitle = "Proceedings of the 2018 Conference on Empirical Methods in Natural Language Processing",
    month = oct # "-" # nov,
    year = "2018",
    address = "Brussels, Belgium",
    publisher = "Association for Computational Linguistics",
    url = "https://aclanthology.org/D18-1151/",
    doi = "10.18653/v1/D18-1151",
    pages = "1192--1202"
}

@article{warstadt-etal-2020-blimp-benchmark,
    title = "{BL}i{MP}: The Benchmark of Linguistic Minimal Pairs for {E}nglish",
    author = "Warstadt, Alex  and
      Parrish, Alicia  and
      Liu, Haokun  and
      Mohananey, Anhad  and
      Peng, Wei  and
      Wang, Sheng-Fu  and
      Bowman, Samuel R.",
    editor = "Johnson, Mark  and
      Roark, Brian  and
      Nenkova, Ani",
    journal = "Transactions of the Association for Computational Linguistics",
    volume = "8",
    year = "2020",
    address = "Cambridge, MA",
    publisher = "MIT Press",
    url = "https://aclanthology.org/2020.tacl-1.25/",
    doi = "10.1162/tacl_a_00321",
    pages = "377--392"
}

@inproceedings{anh2024morphology,
    title = "Morphology Matters: Probing the Cross-linguistic Morphological Generalization Abilities of Large Language Models through a Wug Test",
    author = "Anh, Dang  and
      Raviv, Limor  and
      Galke, Lukas",
    editor = "Kuribayashi, Tatsuki  and
      Rambelli, Giulia  and
      Takmaz, Ece  and
      Wicke, Philipp  and
      Oseki, Yohei",
    booktitle = "Proceedings of the Workshop on Cognitive Modeling and Computational Linguistics",
    month = aug,
    year = "2024",
    address = "Bangkok, Thailand",
    publisher = "Association for Computational Linguistics",
    url = "https://aclanthology.org/2024.cmcl-1.15/",
    doi = "10.18653/v1/2024.cmcl-1.15",
    pages = "177--188"
}

@inproceedings{finlayson-etal-2021-causal,
    title = "Causal Analysis of Syntactic Agreement Mechanisms in Neural Language Models",
    author = "Finlayson, Matthew  and
      Mueller, Aaron  and
      Gehrmann, Sebastian  and
      Shieber, Stuart  and
      Linzen, Tal  and
      Belinkov, Yonatan",
    editor = "Zong, Chengqing  and
      Xia, Fei  and
      Li, Wenjie  and
      Navigli, Roberto",
    booktitle = "Proceedings of the 59th Annual Meeting of the Association for Computational Linguistics and the 11th International Joint Conference on Natural Language Processing (Volume 1: Long Papers)",
    month = aug,
    year = "2021",
    address = "Online",
    publisher = "Association for Computational Linguistics",
    url = "https://aclanthology.org/2021.acl-long.144/",
    doi = "10.18653/v1/2021.acl-long.144",
    pages = "1828--1843"
}

@inproceedings{brinkmann-etal-2025-large,
    title = "Large Language Models Share Representations of Latent Grammatical Concepts Across Typologically Diverse Languages",
    author = "Brinkmann, Jannik  and
      Wendler, Chris  and
      Bartelt, Christian  and
      Mueller, Aaron",
    editor = "Chiruzzo, Luis  and
      Ritter, Alan  and
      Wang, Lu",
    booktitle = "Proceedings of the 2025 Conference of the Nations of the Americas Chapter of the Association for Computational Linguistics: Human Language Technologies (Volume 1: Long Papers)",
    month = apr,
    year = "2025",
    address = "Albuquerque, New Mexico",
    publisher = "Association for Computational Linguistics",
    url = "https://aclanthology.org/2025.naacl-long.312/",
    doi = "10.18653/v1/2025.naacl-long.312",
    pages = "6131--6150",
    ISBN = "979-8-89176-189-6"
}

@inproceedings{weissweiler-etal-2023-counting,
    title = "Counting the Bugs in {C}hat{GPT}{'}s Wugs: A Multilingual Investigation into the Morphological Capabilities of a Large Language Model",
    author = "Weissweiler, Leonie  and
      Hofmann, Valentin  and
      Kantharuban, Anjali  and
      Cai, Anna  and
      Dutt, Ritam  and
      Hengle, Amey  and
      Kabra, Anubha  and
      Kulkarni, Atharva  and
      Vijayakumar, Abhishek  and
      Yu, Haofei  and
      Schuetze, Hinrich  and
      Oflazer, Kemal  and
      Mortensen, David",
    editor = "Bouamor, Houda  and
      Pino, Juan  and
      Bali, Kalika",
    booktitle = "Proceedings of the 2023 Conference on Empirical Methods in Natural Language Processing",
    month = dec,
    year = "2023",
    address = "Singapore",
    publisher = "Association for Computational Linguistics",
    url = "https://aclanthology.org/2023.emnlp-main.401/",
    doi = "10.18653/v1/2023.emnlp-main.401",
    pages = "6508--6524"
}

@inproceedings{gradiend,
title={{GRADIEND}: Feature Learning within Neural Networks Exemplified through Biases},
author={Jonathan Drechsel and Steffen Herbold},
booktitle={The Fourteenth International Conference on Learning Representations},
year={2026},
url={https://openreview.net/forum?id=1vBNAnAgCD}
}

@article{bricken2023monosemanticity,
   title={Towards Monosemanticity: Decomposing Language Models With Dictionary Learning},
   author={Bricken, Trenton and Templeton, Adly and Batson, Joshua and Chen, Brian and Jermyn, Adam and Conerly, Tom and Turner, Nick and Anil, Cem and Denison, Carson and Askell, Amanda and Lasenby, Robert and Wu, Yifan and Kravec, Shauna and Schiefer, Nicholas and Maxwell, Tim and Joseph, Nicholas and Hatfield-Dodds, Zac and Tamkin, Alex and Nguyen, Karina and McLean, Brayden and Burke, Josiah E and Hume, Tristan and Carter, Shan and Henighan, Tom and Olah, Christopher},
   year={2023},
   journal={Transformer Circuits Thread},
   url={https://transformer-circuits.pub/2023/monosemantic-features/index.html}
}

@article{acs2024morphosyntactic,
  title={Morphosyntactic probing of multilingual {BERT} models},
  author={Acs, Judit and Hamerlik, Endre and Schwartz, Roy and Smith, Noah A and Kornai, Andras},
  journal={Natural Language Engineering},
  volume={30},
  year={2022},
  publisher={Cambridge University Press},
    url={https://arxiv.org/pdf/2306.06205}
}

@inproceedings{hewitt-manning-2019-structural,
    title = "{A} Structural Probe for Finding Syntax in Word Representations",
    author = "Hewitt, John  and
      Manning, Christopher D.",
    editor = "Burstein, Jill  and
      Doran, Christy  and
      Solorio, Thamar",
    booktitle = "Proceedings of the 2019 Conference of the North {A}merican Chapter of the Association for Computational Linguistics: Human Language Technologies, Volume 1 (Long and Short Papers)",
    month = jun,
    year = "2019",
    address = "Minneapolis, Minnesota",
    publisher = "Association for Computational Linguistics",
    url = "https://aclanthology.org/N19-1419/",
    doi = "10.18653/v1/N19-1419",
    pages = "4129--4138"
}

@inproceedings{jing-etal-2025-lingualens-sae,
    title = "{L}ingua{L}ens: Towards Interpreting Linguistic Mechanisms of Large Language Models via Sparse Auto-Encoder",
    author = "Jing, Yi  and
      Yao, Zijun  and
      Guo, Hongzhu  and
      Ran, Lingxu  and
      Wang, Xiaozhi  and
      Hou, Lei  and
      Li, Juanzi",
    editor = "Christodoulopoulos, Christos  and
      Chakraborty, Tanmoy  and
      Rose, Carolyn  and
      Peng, Violet",
    booktitle = "Proceedings of the 2025 Conference on Empirical Methods in Natural Language Processing",
    month = nov,
    year = "2025",
    address = "Suzhou, China",
    publisher = "Association for Computational Linguistics",
    url = "https://aclanthology.org/2025.emnlp-main.1433/",
    doi = "10.18653/v1/2025.emnlp-main.1433",
    pages = "28220--28239",
    ISBN = "979-8-89176-332-6"
}

@inproceedings{glue,
    title = "{GLUE}: A Multi-Task Benchmark and Analysis Platform for Natural Language Understanding",
    author = "Wang, Alex  and
      Singh, Amanpreet  and
      Michael, Julian  and
      Hill, Felix  and
      Levy, Omer  and
      Bowman, Samuel",
    editor = "Linzen, Tal  and
      Chrupa{\l}a, Grzegorz  and
      Alishahi, Afra",
    booktitle = "Proceedings of the 2018 {EMNLP} Workshop {B}lackbox{NLP}: Analyzing and Interpreting Neural Networks for {NLP}",
    month = nov,
    year = "2018",
    address = "Brussels, Belgium",
    publisher = "Association for Computational Linguistics",
    url = "https://aclanthology.org/W18-5446/",
    doi = "10.18653/v1/W18-5446",
    pages = "353--355"
}

@article{Honnibal_spaCy_Industrial-strength_Natural_2020,
author = {Honnibal, Matthew and Montani, Ines and Van Landeghem, Sofie and Boyd, Adriane},
doi = {10.5281/zenodo.1212303},
title = {{spaCy: Industrial-strength Natural Language Processing in Python}},
year = {2020}
}

@inbook{superglue,
author = {Wang, Alex and Pruksachatkun, Yada and Nangia, Nikita and Singh, Amanpreet and Michael, Julian and Hill, Felix and Levy, Omer and Bowman, Samuel R.},
title = {SuperGLUE: a stickier benchmark for general-purpose language understanding systems},
year = {2019},
publisher = {Curran Associates Inc.},
address = {Red Hook, NY, USA},
booktitle = {Proceedings of the 33rd International Conference on Neural Information Processing Systems},
articleno = {294},
numpages = {15},
url={https://arxiv.org/abs/1905.00537}
}

@book{davison1997bootstrap, 
place={Cambridge}, 
series={Cambridge Series in Statistical and Probabilistic Mathematics}, 
title={Bootstrap Methods and their Application}, 
publisher={Cambridge University Press}, 
author={Davison, A. C. and Hinkley, D. V.}, 
year={1997}, 
collection={Cambridge Series in Statistical and Probabilistic Mathematics},
doi={10.1017/CBO9780511802843}
}

@inproceedings{tenney-etal-2019-bert,
    title = "{BERT} Rediscovers the Classical {NLP} Pipeline",
    author = "Tenney, Ian  and
      Das, Dipanjan  and
      Pavlick, Ellie",
    editor = "Korhonen, Anna  and
      Traum, David  and
      M{\`a}rquez, Llu{\'i}s",
    booktitle = "Proceedings of the 57th Annual Meeting of the Association for Computational Linguistics",
    month = jul,
    year = "2019",
    address = "Florence, Italy",
    publisher = "Association for Computational Linguistics",
    url = "https://aclanthology.org/P19-1452/",
    doi = "10.18653/v1/P19-1452",
    pages = "4593--4601"
}

\appendix

\section{Data}\label{app:data}

Detailed data generation details for the article data and \dataNEUT, introduced in Section~\ref{sec:data}.

\subsection{Article Data}

This section provides details on generation for the \gradiend\ training datasets, like \dataNM. Table~\ref{tab:data-size} provides an overview of the generated datasets including sizes. Note that in this study, sometimes only a subset of the subsets is used for specific experiments, e.g., to balance datasets by min sampling.
We release the dataset on Hugging Face \iflink under \href{https://huggingface.co/datasets/aieng-lab/de-gender-case-articles}{\path{aieng-lab/de-gender-case-articles}}.\else anonymous.\fi

\subsubsection{Source Corpus and Sentence Segmentation}

We use the German Wikipedia dump (snapshot \texttt{20220301.de}) as the underlying text corpus \cite{wikipedia}.

Due to the size of the German Wikipedia dump and the fact that the \gradiend\ training does not require a very large number of data points, we do not process the corpus exhaustively.
Instead, articles are subsampled using a fixed stride.
Specifically, we extract short contiguous blocks of articles and skip large intervals between blocks.
This yields a lightweight subset with broad topical coverage while avoiding locality effects introduced by processing consecutive articles only.
The resulting subset is used solely as a source of naturally occurring sentences for morphosyntactic filtering and is not intended to represent a statistically uniform sample of Wikipedia.

\subsubsection{Morphosyntactic Annotation}

Each sentence is processed with spaCy to obtain token-level part-of-speech tags and morphological features.
We rely on spaCy’s morphological annotations to identify the grammatical \emph{case}, \emph{gender}, and \emph{number} of determiner tokens.
Only tokens tagged as determiners (\texttt{POS=DET}) are considered as candidates for definite singular articles.

\subsubsection{Sentence Filtering}

For each gender-case combination \( z\,{=}\,(g,c)\,{\in}\,\mathcal{G}\times\mathcal{C} \), we construct a separate dataset by retaining only sentences that satisfy all of the following constraints:

\begin{table}[!t]
    \centering
    \footnotesize
        \setlength{\tabcolsep}{5pt}
    \begin{tabular}{llrrrr}\toprule
          &&  \multicolumn{4}{c}{\textbf{Size}}   \\\cmidrule{3-6}
         \textbf{Dataset} & \textbf{ID} & \textbf{Total} & \textbf{Train} & \textbf{Val} & \textbf{Test} \\\midrule
\dataNM & \texttt{masc\_nom} & 34,350 & 27,829 & 3,084 & 3,437 \\
\dataAM & \texttt{masc\_acc} & 30,538 & 24,781 & 2,705 & 3,052 \\
\dataDM & \texttt{masc\_dat} & 23,437 & 18,918 & 2,176 & 2,343 \\
\dataGM & \texttt{masc\_gen} & 34,087 & 27,417 & 3,254 & 3,416 \\

\dataNF & \texttt{fem\_nom} & 61,328 & 49,399 & 5,796 & 6,133 \\
\dataAF & \texttt{fem\_acc} & 34,801 & 28,155 & 3,166 & 3,480 \\
\dataDF & \texttt{fem\_dat} & 46,601 & 37,458 & 4,482 & 4,661 \\
\dataGF & \texttt{fem\_gen} & 38,811 & 31,219 & 3,711 & 3,881 \\

\dataNN & \texttt{neut\_nom} & 33,350 & 26,680 & 3,335 & 3,335 \\
\dataAN & \texttt{neut\_acc} & 19,012 & 15,209 & 1,901 & 1,902 \\
\dataDN & \texttt{neut\_dat} & 16,075 & 13,020 & 1,447 & 1,608 \\
\dataGN & \texttt{neut\_gen} & 25,351 & 20,436 & 2,387 & 2,528 \\
          \bottomrule
    \end{tabular}
    \caption{Dataset overview with full dataset sizes.}
    \label{tab:data-size}
\end{table}

\begin{enumerate}
    \item \textbf{Article presence.}  
    The sentence contains at least one occurrence of the surface form corresponding to the target definite article.

    \item \textbf{Morphological agreement.}  
    All occurrences of the target article in the sentence are annotated with
    \(\textsc{Gender}=g\), \(\textsc{Case}=c\), and \(\textsc{Number}=\textsc{Sing}\).
    Sentences containing plural uses of the article are excluded.

    \item \textbf{Limited ambiguity.}  
    Sentences containing more than four occurrences of the target article are discarded to reduce structural ambiguity.

    \item \textbf{Length constraints.}  
    Only sentences with a character length between 50 and 500 are retained.

    \item \textbf{Named entity control.}  
    Sentences containing more than three named entities are excluded to reduce confounds introduced by entity-heavy contexts.

    \item \textbf{Duplicate removal.} Duplicate sentences are removed.
\end{enumerate}

\subsubsection{Data Quality}\label{app:spacy-quality}

\rev{
\textbf{Error types.}
To quantify noise introduced by spaCy-based labeling, we manually annotated a small sample of instances from each gender-case dataset and analyzed the resulting errors.
We distinguish two error categories:
\begin{itemize}[leftmargin=*]
    \item \textbf{In-paradigm cell errors (Cell):} spaCy assigns a token to the wrong gender-case cell \emph{within} the definite singular article paradigm (e.g., \dataNF\ mislabeled as \dataAF). These errors can be problematic because they effectively move examples between our datasets and may introduce misleading transition cues.
    \item \textbf{Out-of-paradigm errors (Other):} spaCy assigns a token to a category \emph{outside} singular definite articles, e.g., relative pronouns, plural determiners, or other non-target uses. These instances primarily add \emph{random} noise, since they do not systematically correspond to any other gender-case dataset used in our analyses.
\end{itemize}
}

\rev{
\textbf{Evaluation protocol.}
We manually labeled 20 randomly sampled text instances from each of the 12 gender-case datasets (Table~\ref{tab:case-gender-table}).
Labels correspond to the intended gender-case cell of the masked singular definite article (or \emph{Other} if the token is not a singular definite article in context).
}

\rev{
\textbf{Overall accuracy.}
Table~\ref{tab:data-quality} reports accuracies aggregated by gender and case.
spaCy achieves an overall accuracy of $0.83$ across all datasets, with the lowest performance in the genitive cells ($0.7$).
Accuracy is measured based on texts, i.e., texts containing multiple masks are only classified as True Positive if all its masks are correct.
}

\begin{table}[!t]
\centering
\small
\begin{tabular}{lrrrrr}
\toprule
 & \textbf{Nom.} & \textbf{Acc.} & \textbf{Dat.} & \textbf{Gen.} & \textbf{All} \\
\midrule
\textbf{Male} & 0.80 & 1.00  & 0.80 & 0.70 & 0.83 \\
\textbf{Neutral} & 0.90 & 0.85 & 0.95 & 0.70 & 0.85  \\
\textbf{Female} & 0.80  & 0.70 & 0.85 & 0.90 & 0.81 \\
\midrule
\textbf{All} & 0.83 & 0.85 & 0.87 & 0.77 & 0.83 \\
\bottomrule
\end{tabular}
\caption{\rev{Accuracy of spaCy-labeled gender-case cell data.}}
\label{tab:data-quality}
\end{table}

\begin{table}[!t]
    \centering
    \begin{tabular}{lrrrr}
    \toprule
         & \dataNM & \dataDF & \dataGF & Other  \\ \midrule
        \dataNM & 20 & 0 & 1 & 4 \\
        \dataDF & 1 & 20 & 1 & 1 \\
        \dataGN & 0 & 1 & 23 & 1\\
        \bottomrule
    \end{tabular}
   \caption{\rev{\emph{der}: confusion matrix (spaCy rows; manual columns).}}
    \label{tab:spacy-der}
\end{table}

\begin{table}[!t]
    \centering
    \begin{tabular}{lrrr}
    \toprule
         & \dataNF & \dataAF & Other  \\ \midrule
        \dataNF & 21 & 0 & 4 \\
        \dataAF & 4 & 19 & 2 \\
        \bottomrule
    \end{tabular}
   \caption{\rev{\emph{die}: confusion matrix (spaCy rows; manual columns).}}
    \label{tab:spacy-die}
\end{table}

\begin{table}[!t]
    \centering
    \begin{tabular}{lrrr}
    \toprule
         & \dataNN & \dataAN & Other  \\ \midrule
        \dataNN & 19 & 1 & 1 \\
        \dataNN & 3 & 17 & 0 \\
        \bottomrule
    \end{tabular}
   \caption{\rev{\emph{das}: confusion matrix (spaCy rows; manual columns).}}
    \label{tab:spacy-das}
\end{table}

\begin{table}[!t]
    \centering
    \begin{tabular}{lrrr}
    \toprule
         & \dataDM & \dataDN & Other  \\ \midrule
        \dataDM & 18 & 3 & 1 \\
        \dataDN & 1 & 19 & 0 \\
        \bottomrule
    \end{tabular}
   \caption{\rev{\emph{dem}: confusion matrix (spaCy rows; manual columns).}}
    \label{tab:spacy-dem}
\end{table}

\begin{table}[!t]
    \centering
    \begin{tabular}{lrrr}
    \toprule
         & \dataGM & \dataGN & Other  \\ \midrule
        \dataGM & 15 & 7 & 1 \\
        \dataGN & 3 & 14 & 3 \\
        \bottomrule
    \end{tabular}
   \caption{\rev{\emph{des}: confusion matrix (spaCy rows; manual columns).}}
    \label{tab:spacy-des}
\end{table}

\rev{
\textbf{Confusion matrices.}
To characterize \emph{Cell} vs.\ \emph{Other} errors in more detail, Tables~\ref{tab:spacy-der}--\ref{tab:spacy-des} report confusion matrices for each surface form (excluding the non-syncretic \emph{den} because of perfect accuracy).
Rows correspond to spaCy assignments and columns to manual labels.
Counts are \emph{raw mask counts} (i.e., each mask is counted individually).
}

\rev{
\textbf{Discussion and implications for our analyses.}
Across forms, \emph{Other} errors are present, but they primarily inject random noise and are unlikely to induce systematic trends in gradient-based analyses.
More critical are \emph{Cell} errors within the article paradigm, which are most pronounced for genitive forms (notably \emph{des}).
Nevertheless, we observe that analyses relying on clean dataset combinations, in particular those involving \emph{der} datasets and \dataNF\ (\emph{die}), match the conclusions obtained from noisier dataset combinations (see e.g., Figures~\ref{fig:bert-probability-heatmap-N-MF}, \ref{fig:venn-der-die-bert-weight}, and \ref{fig:venn-der-die-other-models-weight} with \emph{clean} results).
This consistency suggests that the overall trends reported in the paper are not driven solely by spaCy labeling noise.
}

\subsection{\rev{Gender-Case} Neutral Dataset (\dataNEUT)}
The \dataNEUT\ dataset is constructed from the Wortschatz Leipzig German news corpus\footnote{\url{https://downloads.wortschatz-leipzig.de/corpora/deu_news_2024_300K.tar.gz}} \cite{goldhahn2012lcc, wortschatz_deu_news_2024_300K} to provide sentence contexts without grammatical gender or case cues.
We apply a series of linguistic filters using spaCy to remove sentences that could implicitly encode such information.

Specifically, we exclude sentences that contain determiners or definite and indefinite articles (including \textit{der/die/das}, \textit{ein/kein} and their inflected forms), as well as sentences containing third-person pronouns.
To further reduce implicit gender signals, we remove sentences dominated by named entities, as proper names can carry gender information.
We additionally filter out very short sentences and sentences containing the token \textit{das} to avoid homonym-induced ambiguity.

The resulting dataset with 9,570 entries consists of well-formed sentences that are largely free of explicit and implicit morphosyntactic gender–case cues and is used as a grammar-neutral reference throughout our experiments.

We release the dataset on Hugging Face \iflink under \href{https://huggingface.co/datasets/aieng-lab/wortschatz-leipzig-de-grammar-neutral}{\path{aieng-lab/wortschatz-leipzig-de-grammar-neutral}}.\else anonymous.\fi

\begin{table*}[!tb]
    \centering
    \small
    \begin{tabular}{lrll}
         \toprule
         \textbf{Model} & \textbf{\# Parameters} & \textbf{Checkpoint}  & \textbf{Reference} \\ \midrule
         \bert  & 109.1M  & \href{https://huggingface.co/google-bert/bert-base-german-cased}{\texttt{google-bert/bert-base-german-cased}}& \citet{bert} \\
         \gpttwo & 124.4M & \href{https://huggingface.co/dbmdz/german-gpt2}{\texttt{dbmdz/german-gpt2}}  & \citet{gpt} \\
         \eurobert & 211.8M &  \href{https://huggingface.co/EuroBERT/EuroBERT-210m}{\texttt{EuroBERT/EuroBERT-210m}} & \citet{eurobert} \\
         \gbert & 335.7M & \href{https://huggingface.co/deepset/gbert-large}{\texttt{deepset/gbert-large}}  & \citet{gbert} \\
         \modernbert & 1.06B &  \href{https://huggingface.co/LSX-UniWue/ModernGBERT\_1B}{\texttt{LSX-UniWue/ModernGBERT\_1B}} & \citet{moderngbert}  \\
         \llama  & 3.21B & \href{https://huggingface.co/meta-llama/Llama-3.2-3B}{\texttt{meta-llama/Llama-3.2-3B}} & \citet{grattafiori2024llama} \\
         \bottomrule
    \end{tabular}
    \caption{Hugging Face model checkpoints used in this study.}
    \label{tab:models}
\end{table*}

\section{Decoder-Only Models for MLM}\label{app:decoder-only}

Due to the fact that the gender of the definite article is usually only determined by the noun which naturally occurs in the right context, only using the left context is not an option for the considered problem, as done by \cite{gradiend}. Instead, we convert the decoder-only model into a model that can also predict a token given a right context, similar to the MLM task. We use the following general approach. We add a \texttt{[MASK]} token to the decoder's tokenizer, and use the next $N>0$ final hidden states of the decoder after the \texttt{[MASK]} token as mean-pooled input for a simple classifier network. The classifier has six classes, one for each German definite articles. This custom head makes it possible to use the decoder-only model for a bidirectional prediction task similar to a MLM task, at least to predict one of the six articles, and, importantly, create meaningful gradients through the entire model.

\begin{figure}[!tb]
    \centering
    \includegraphics[width=\linewidth]{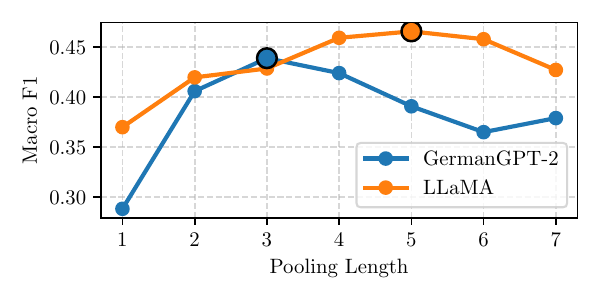}
    \vspace{-1.9em}
    \caption{Decoder-only article classifier performance across different pooling lengths.}
    \label{fig:pooling_length}
\end{figure}

For the training of the classifier, we froze the core model parameters. This avoids changing the model, which could invalidate the \gradiend\ models, as they may instead of analyzing the original model learn where the fine-tuning for this MLM-head updated the model. To choose an appropriate $N$, i.e., the hidden states after the \texttt{[MASK]} to consider as pooled classifier input, there is a natural trade-off. A low number might not include the encodings of the noun tokens (e.g., due to an adjective token(s) between the article and the noun). A too large number makes the relevant information from the noun less relevant, as it contributes less to the average pooling. We use $N\,{=}\,3$ for GPT2 and $N\,{=}\,5$ for \llama, as shown in Figure~\ref{fig:pooling_length}.

\begin{table}[!t]
    \centering
    \scriptsize
    \begin{tabular}{ll}
    \toprule
        \textbf{Hyperparameter} & \textbf{Value} \\\midrule
          Optimizer & Adam \\
          Learning Rate &  \num{1e-5} (encoder-only); \\&\num{1e-4} (decoder-only) \\ 
          Weight Decay & \num{1e-2} \\
          Batch Size Gradient Computation & 4 (\llama), 32 (others) \\
          Batch Size \gradiend & 1 \\
          Training Criterion & MSE \\
          Training Steps & 5,000 \\
          Evaluation Steps & 1,000 \\
          Evaluation Max Size & 500 \\
          Evaluation Criterion & \cor\ on validation split \\
          \bottomrule
    \end{tabular}
    \caption{\gradiend\ training hyperparameters.}\label{tab:hyperparameters}
\end{table}

Overall, the classification performance is not too great considering the low number of classes (six), but sufficient to train the \gradiend\ models.

\section{Training}\label{app:training}

Model details and training hyperparameters are reported in Tables~\ref{tab:models} and~\ref{tab:hyperparameters}, respectively.
Following \citet{gradiend}, we train each \gradiend\ variant with three random seeds and select the best run by the validation-set correlation as used in Table~\ref{tab:encoded-values-correlations} (see Appendix~\ref{app:encoder} for details).
For efficiency, we estimate this correlation on a 100-example subset per gender-case dataset from the validation split.
We train \llama\ and \modernbert\ in \texttt{torch.bfloat16} and all other models in \texttt{torch.float32}.

We use an oversampled single-label batch sampler that groups examples by gender–case dataset and constructs batches containing only one gender-case dataset label at a time. To ensure equal exposure across datasets, batches are oversampled to match the maximum number of batches per label and then interleaved in a round-robin fashion.

Experiments are run in Python 3.9.19.
\llama\ is trained on three NVIDIA A100 GPUs (80\,GB each), while all other models use a single A100.
Per-seed training time ranges from $\sim$1 hour (smaller models) to $\sim$3 hours (\llama) for a single variant.

\section{Encoded Values}\label{app:encoder}

\begin{figure}[!p]
    \centering
  \includegraphics[width=\linewidth]{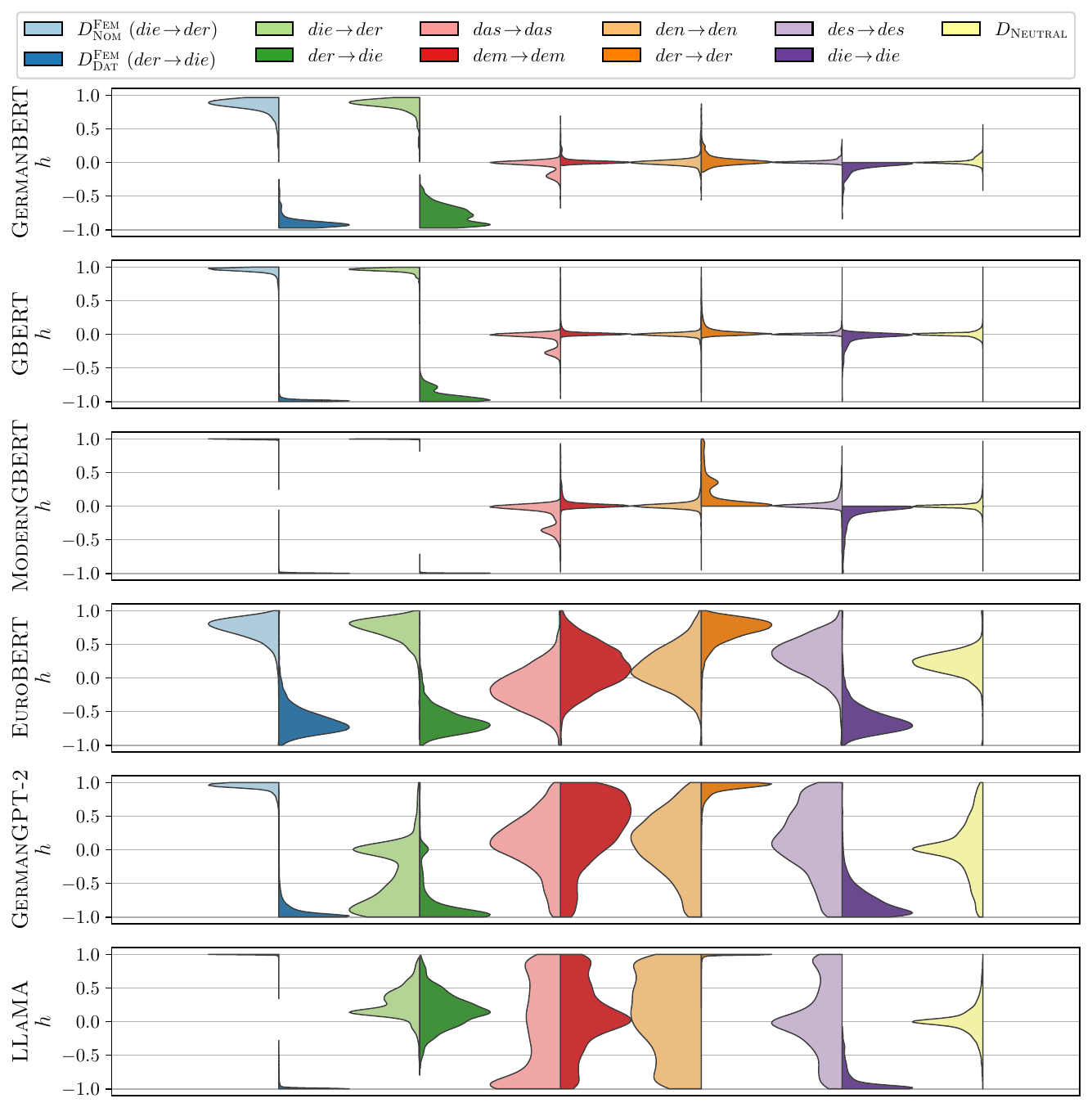}
    \caption{Encoded value distribution of \gradcNDgF\ for different input gradients.}
    \label{fig:encoded-all-models-ND_F}
\end{figure}

\begin{figure}[!p]
    \centering
  \includegraphics[width=\linewidth]{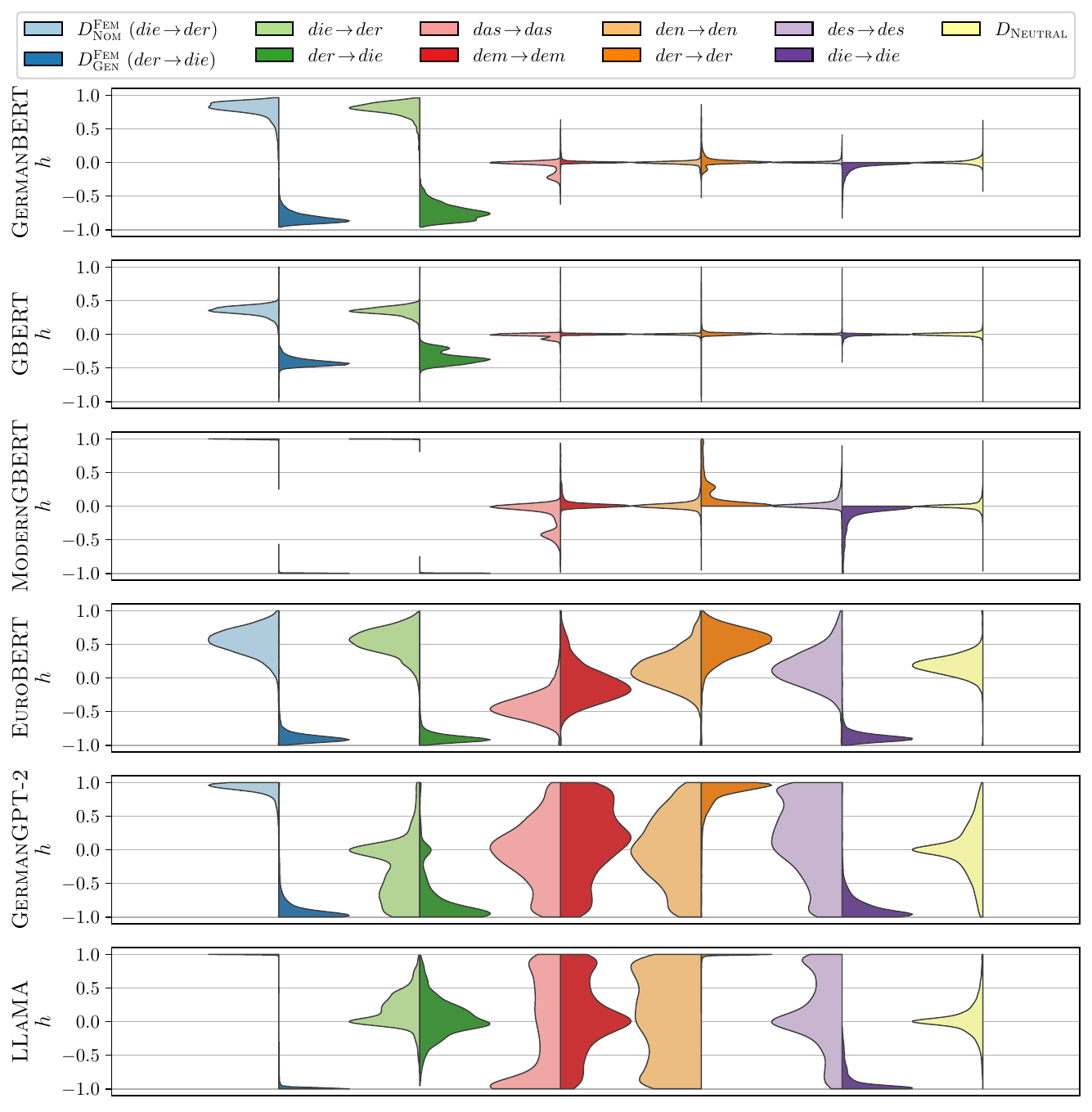}
    \caption{Encoded value distribution of \gradcNGgF\ for different input gradients.}
    \label{fig:encoded-all-models-NG_F}
\end{figure}

\begin{figure}[!p]
    \centering
  \includegraphics[width=\linewidth]{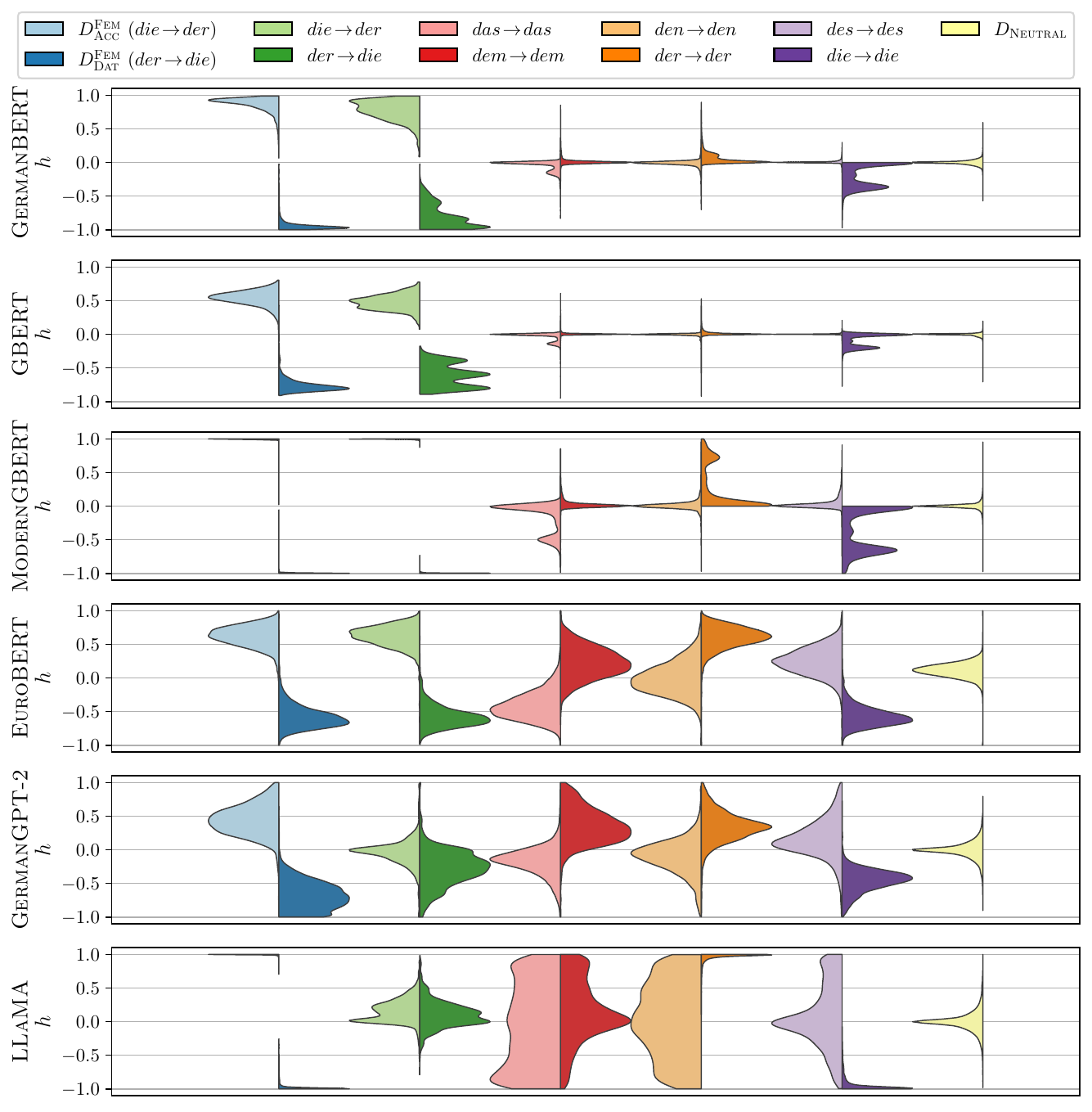}
    \caption{Encoded value distribution of \gradcDAgF\ for different input gradients.}
    \label{fig:encoded-all-models-DA_F}
\end{figure}

\begin{figure}[!p]
    \centering
  \includegraphics[width=\linewidth]{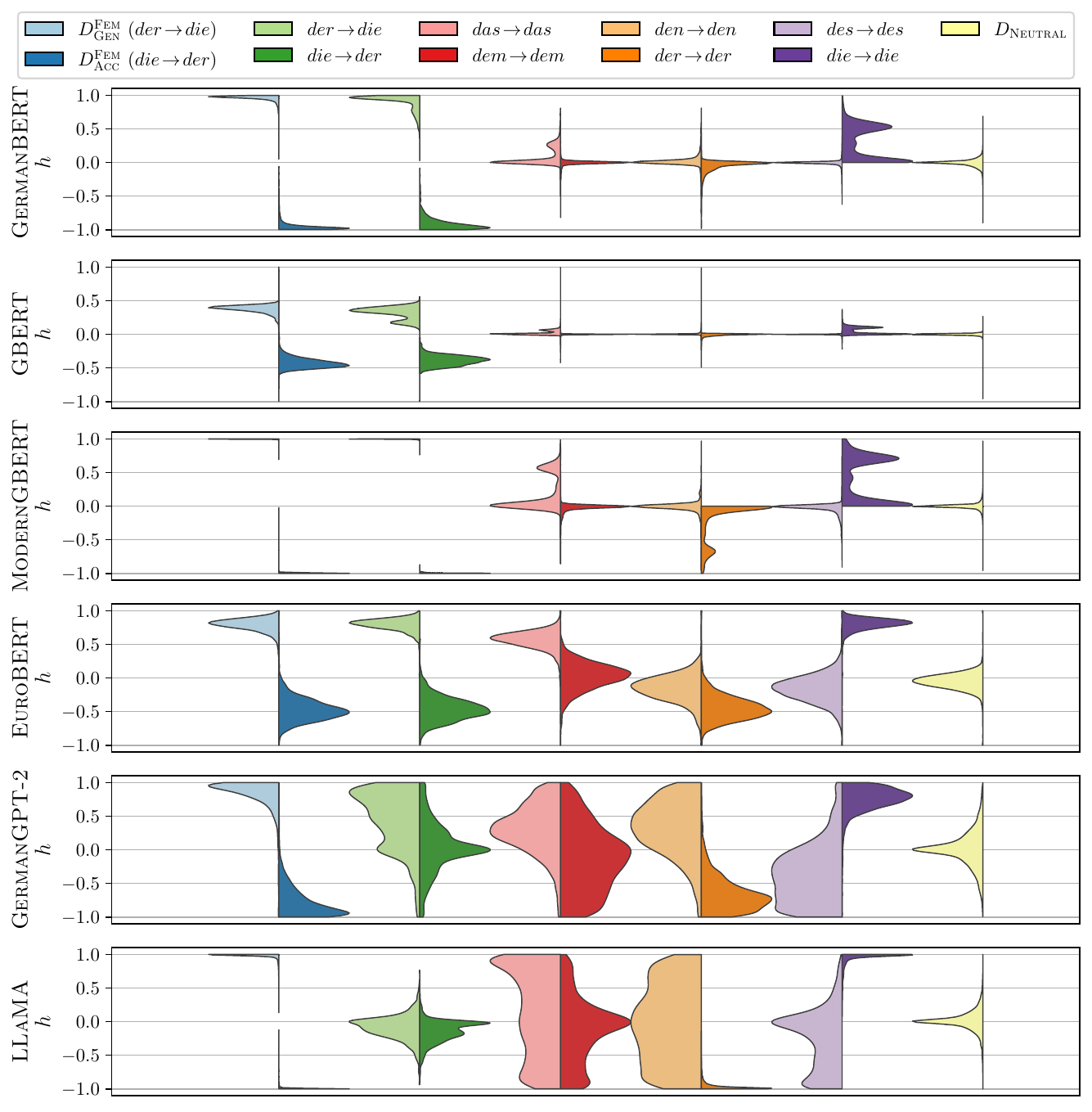}
    \caption{Encoded value distribution of \gradcGAgF\ for different input gradients.}
    \label{fig:encoded-all-models-GA_F}
\end{figure}

\begin{figure}[!p]
    \centering
  \includegraphics[width=\linewidth]{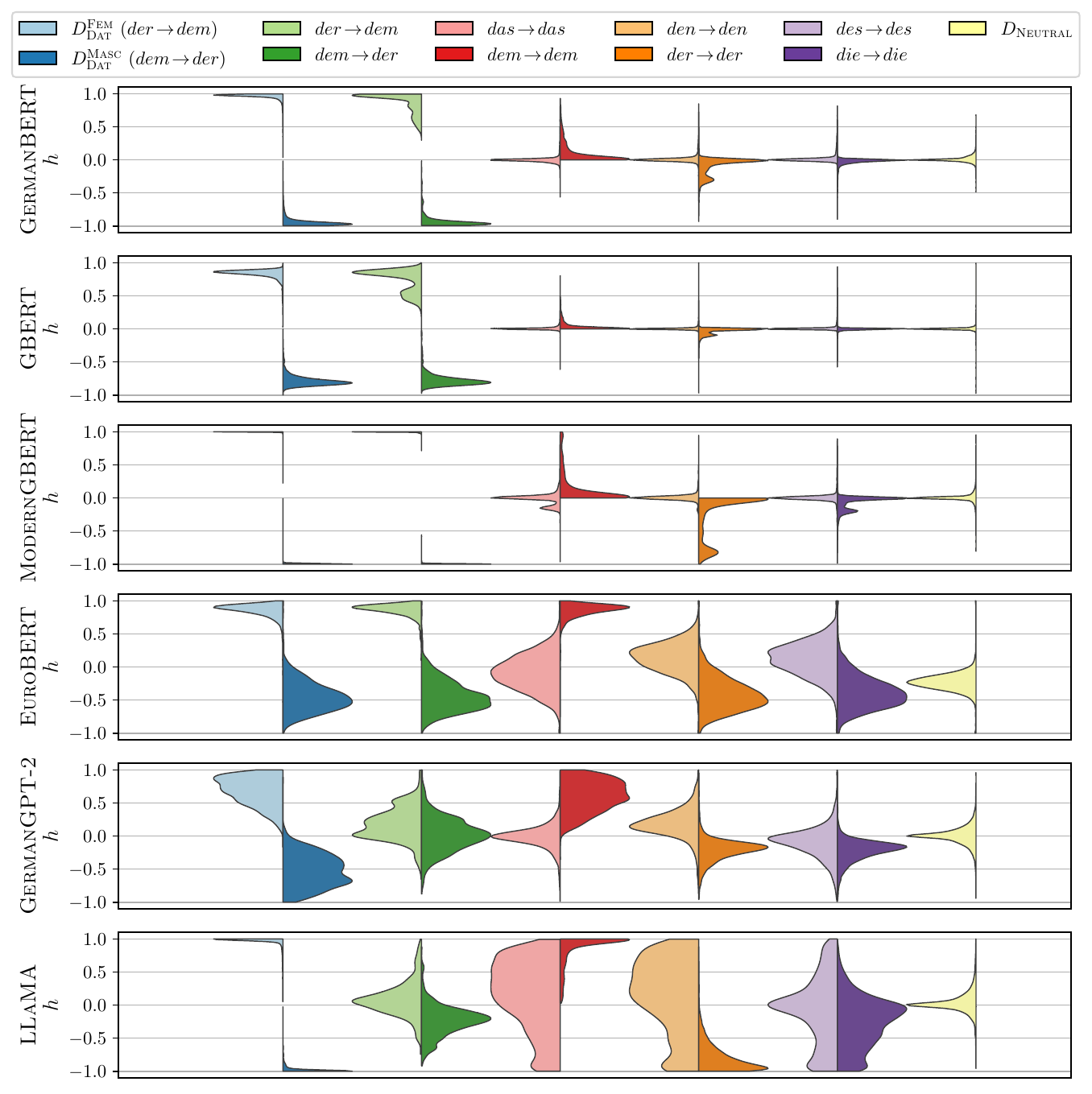}
    \caption{Encoded value distribution of \gradcDgMF\ for different input gradients.}
    \label{fig:encoded-all-models-D_MF}
\end{figure}

\begin{figure}[!p]
    \centering
  \includegraphics[width=\linewidth]{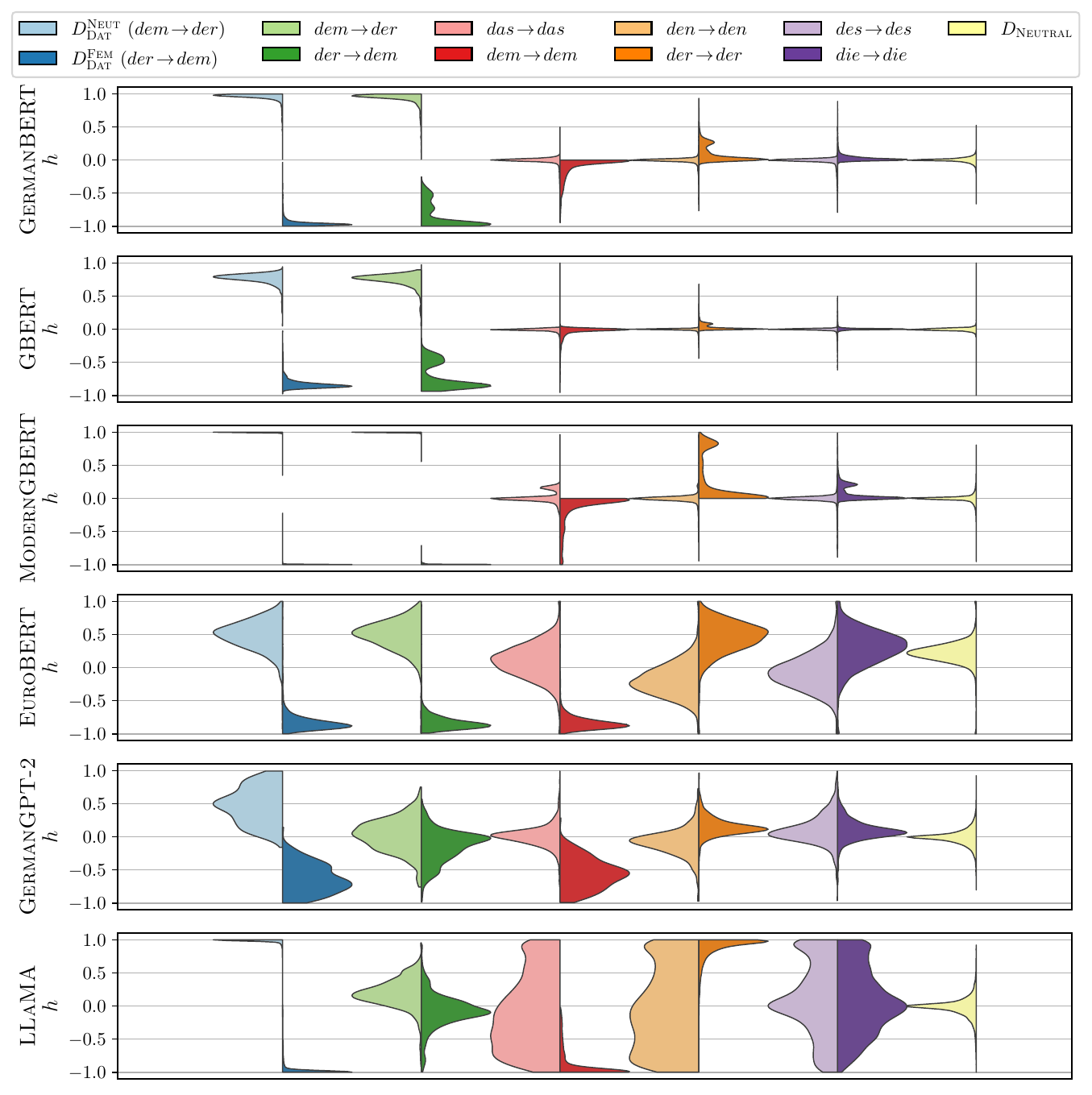}
    \caption{Encoded value distribution of \gradcDgFN\ for different input gradients.}
    \label{fig:encoded-all-models-D_FN}
\end{figure}

\begin{figure}[!p]
    \centering
  \includegraphics[width=\linewidth]{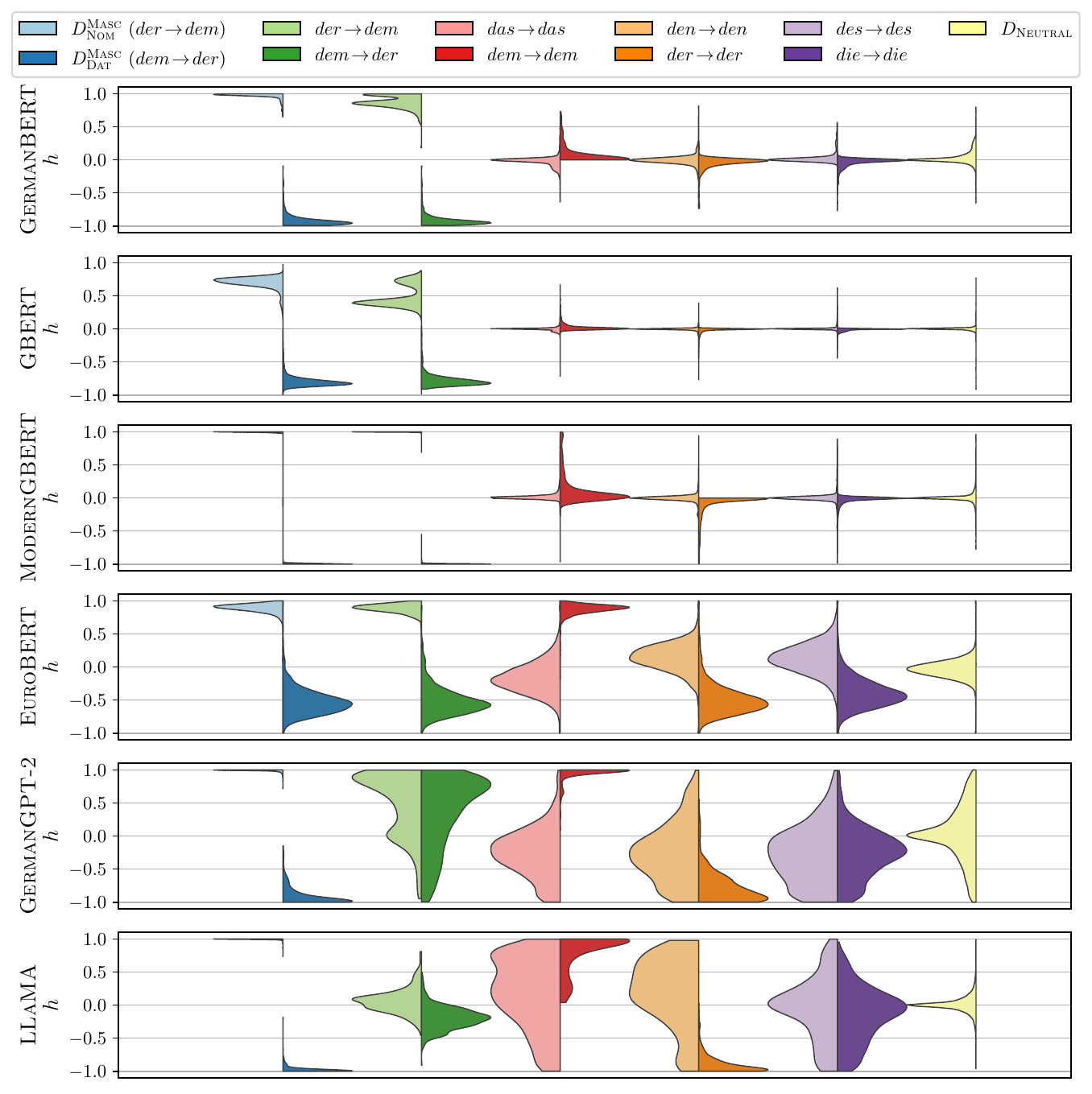}
    \caption{Encoded value distribution of \gradcNDgM\ for different input gradients.}
    \label{fig:encoded-all-models-ND_M}
\end{figure}

\begin{figure}[!p]
    \centering
  \includegraphics[width=\linewidth]{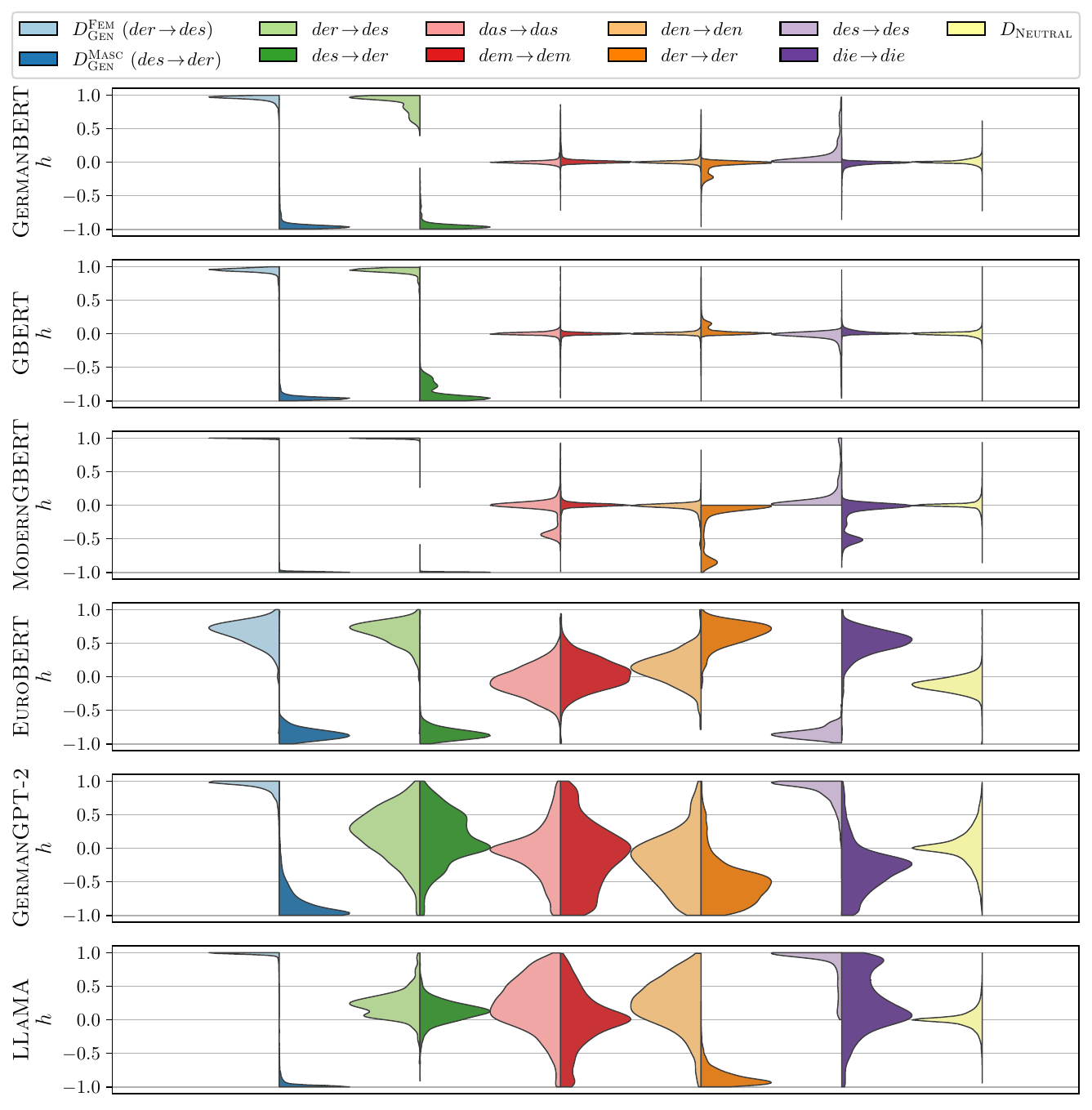}
    \caption{Encoded value distribution of \gradcGgMF\ for different input gradients.}
    \label{fig:encoded-all-models-G_MF}
\end{figure}

\begin{figure}[!p]
    \centering
  \includegraphics[width=\linewidth]{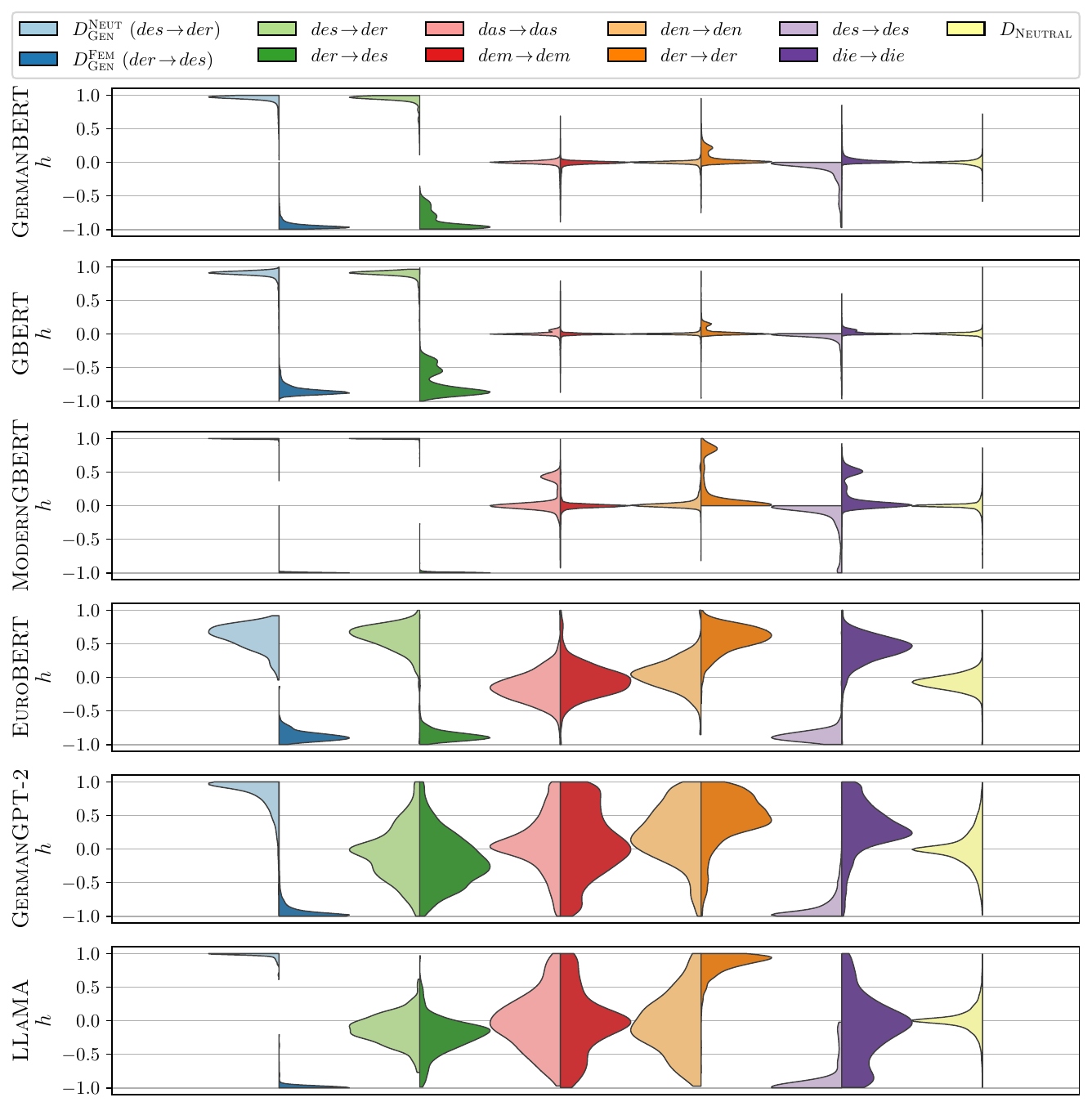}
    \caption{Encoded value distribution of \gradcGgFN\ for different input gradients.}
    \label{fig:encoded-all-models-G_FN}
\end{figure}

\begin{figure}[!p]
    \centering
  \includegraphics[width=\linewidth]{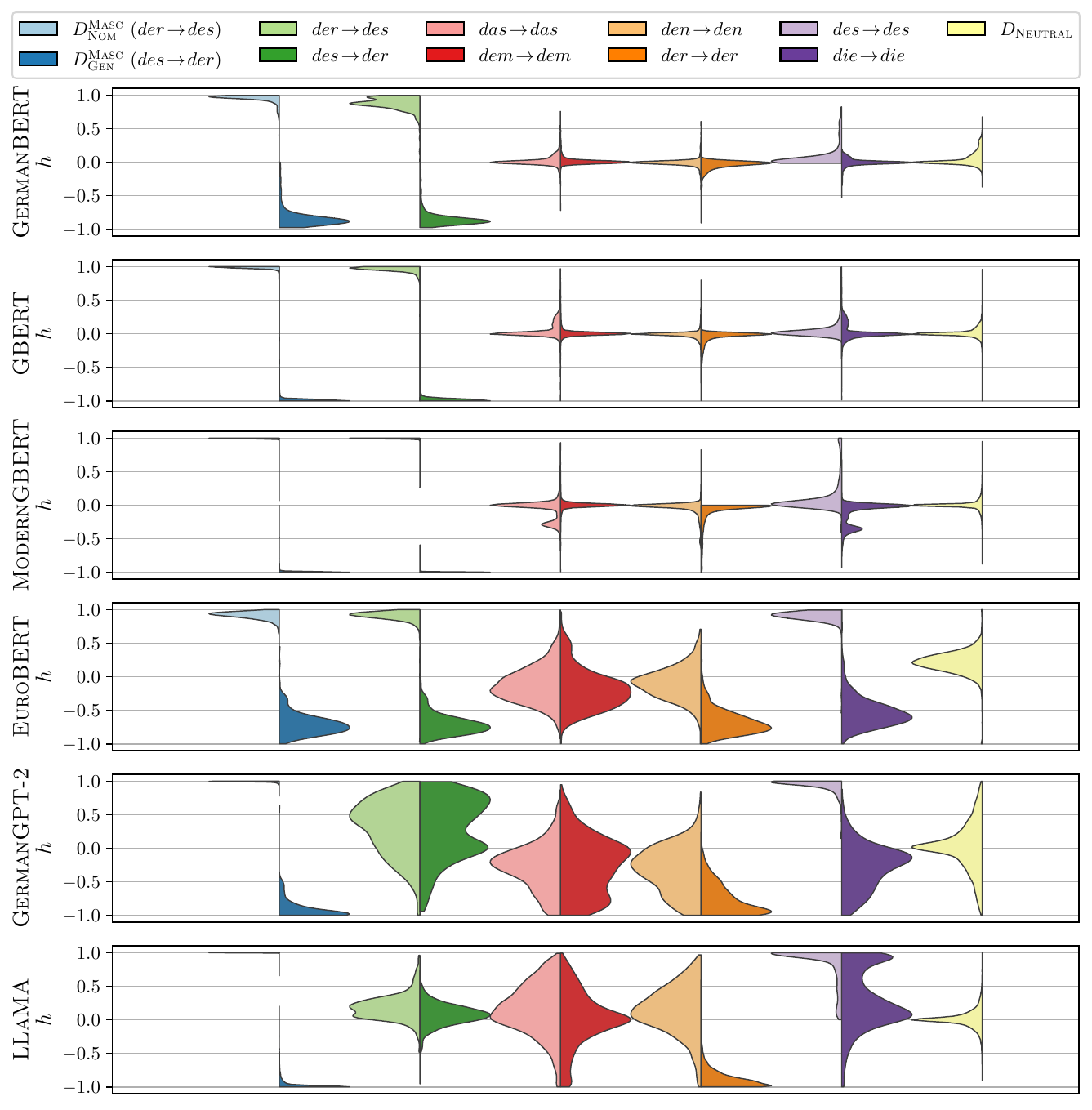}
    \caption{Encoded value distribution of \gradcNGgM\ for different input gradients.}
    \label{fig:encoded-all-models-NG_M}
\end{figure}

\begin{figure}[!p]
    \centering
  \includegraphics[width=\linewidth]{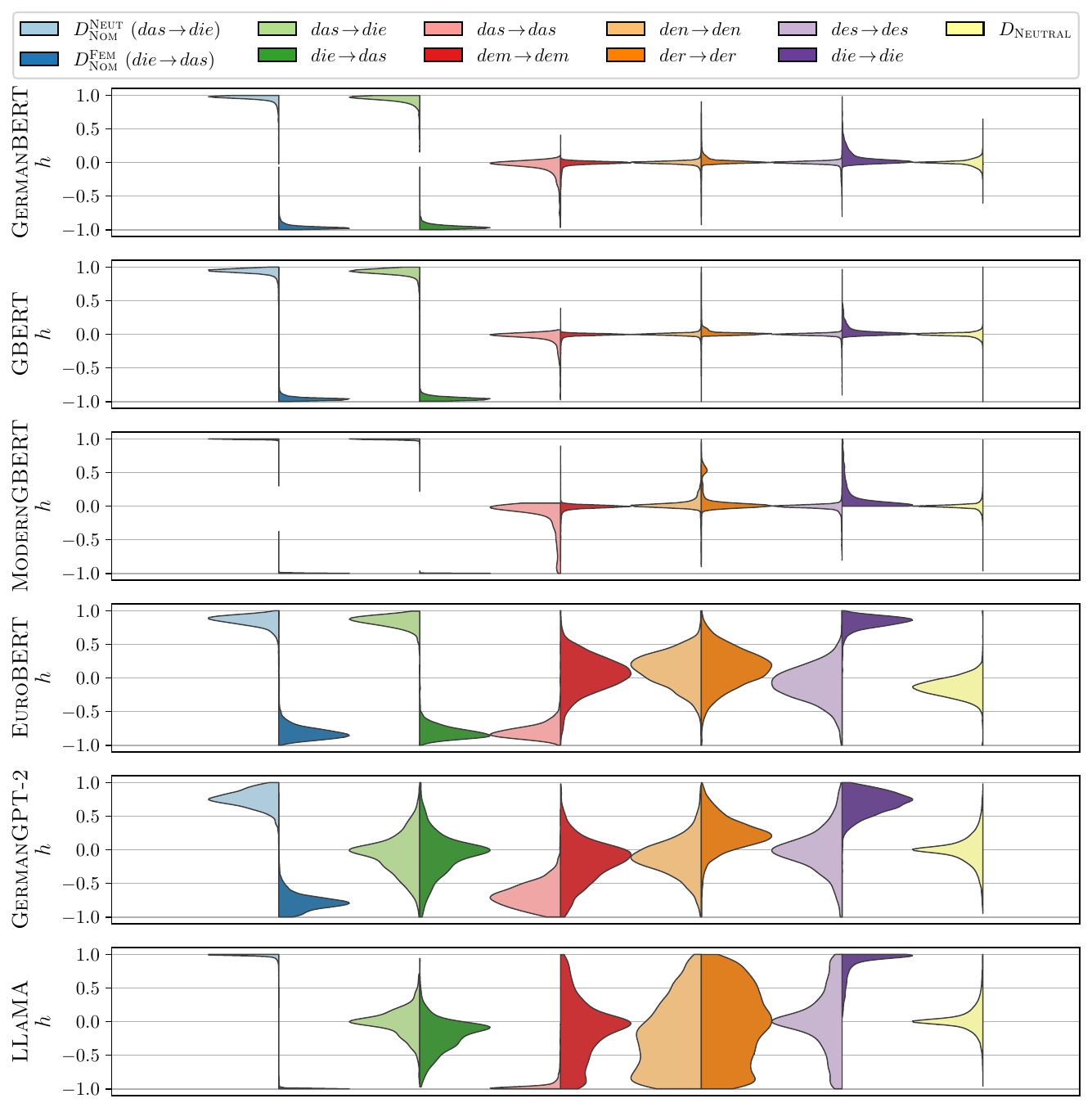}
    \caption{Encoded value distribution of \gradcNgFN\ for different input gradients.}
    \label{fig:encoded-all-models-N_FN}
\end{figure}

\begin{figure}[!p]
    \centering
  \includegraphics[width=\linewidth]{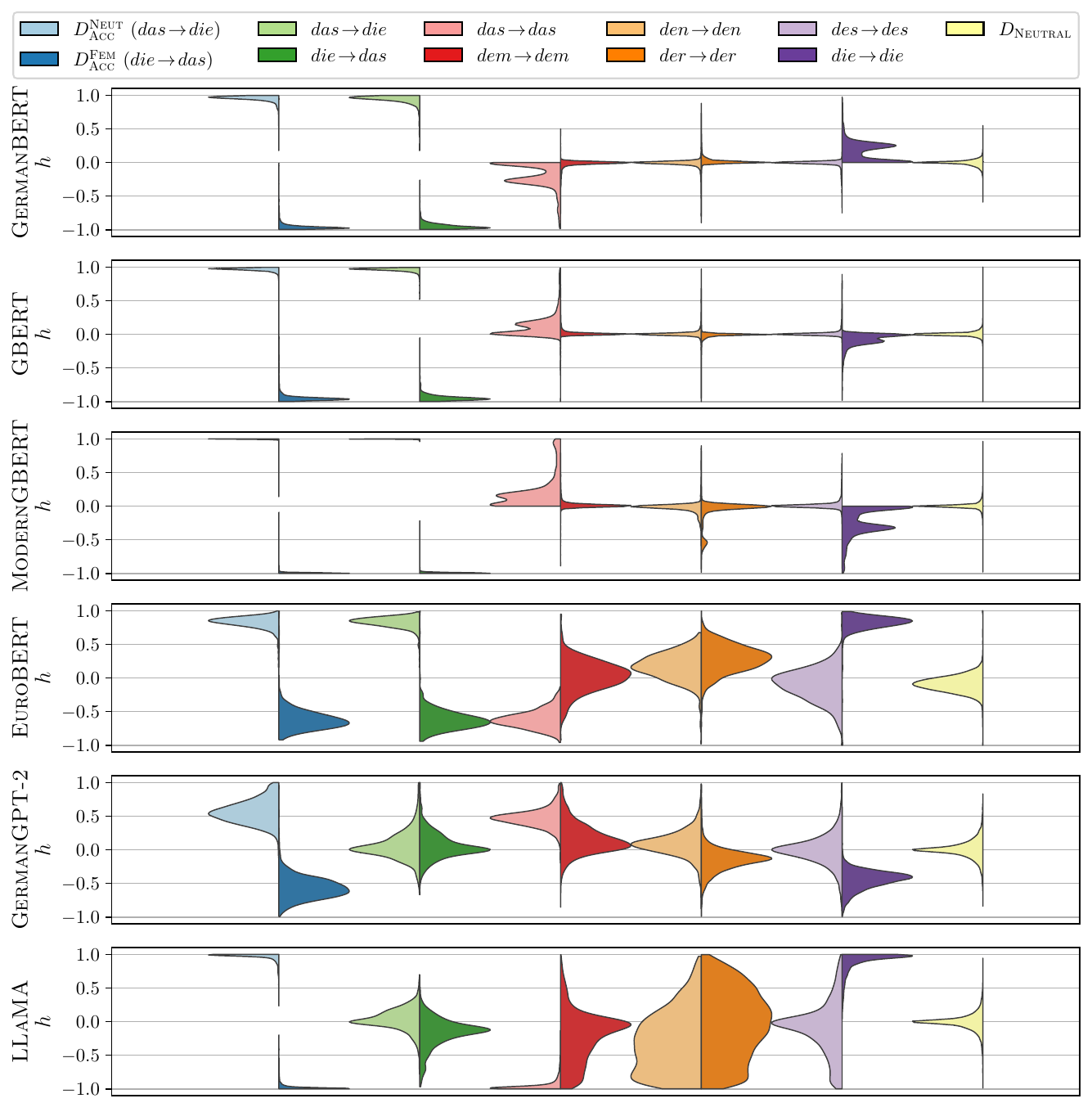}
    \caption{Encoded value distribution of \gradcAgFN\ for different input gradients.}
    \label{fig:encoded-all-models-A_FN}
\end{figure}

\begin{figure}[!p]
    \centering
  \includegraphics[width=\linewidth]{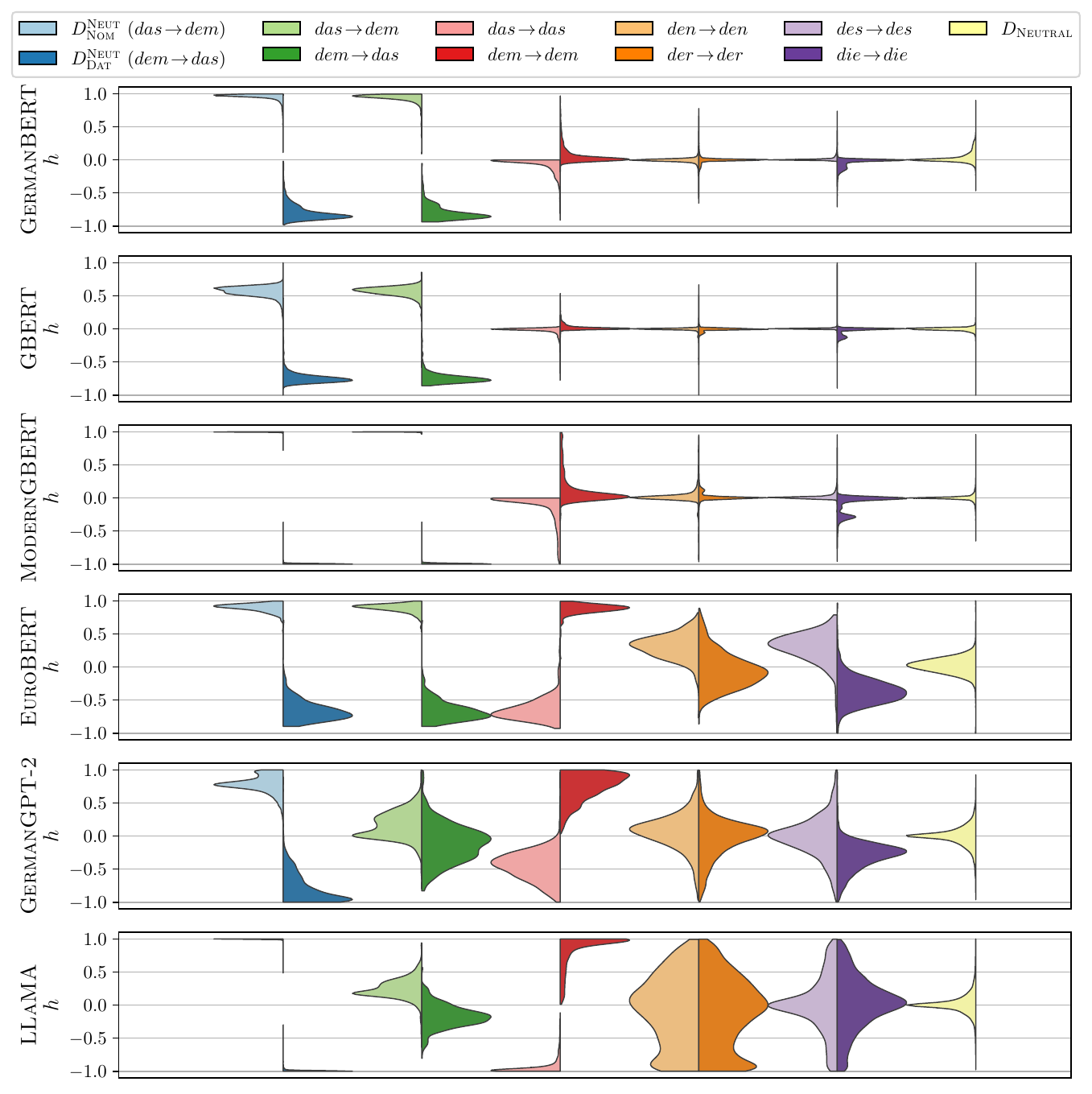}
    \caption{Encoded value distribution of \gradcNDgN\ for different input gradients.}
    \label{fig:encoded-all-models-ND_N}
\end{figure}

\begin{figure}[!p]
    \centering
  \includegraphics[width=\linewidth]{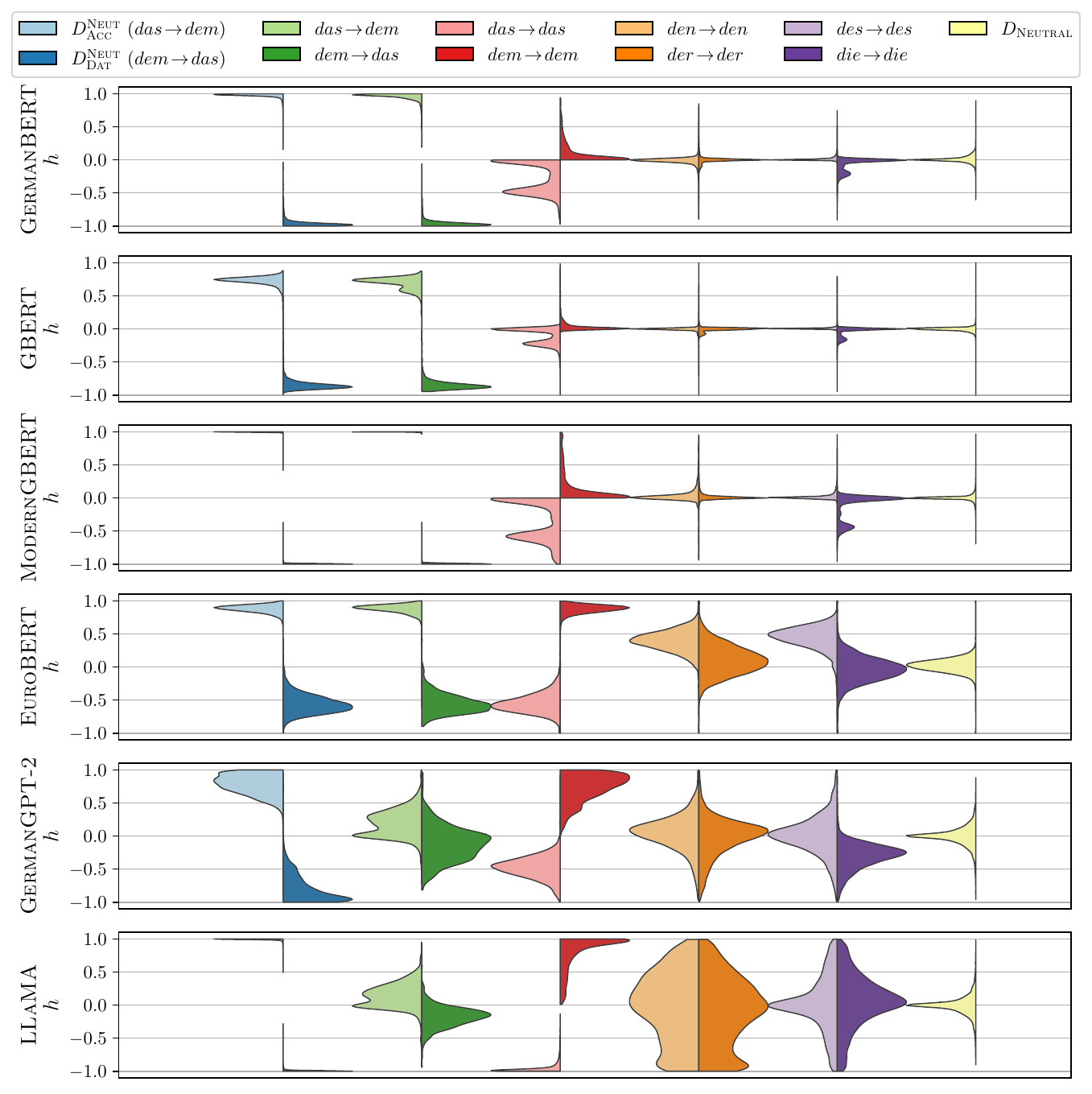}
    \caption{Encoded value distribution of \gradcADgN\ for different input gradients.}
    \label{fig:encoded-all-models-AD_N}
\end{figure}

\begin{figure}[!p]
    \centering
  \includegraphics[width=\linewidth]{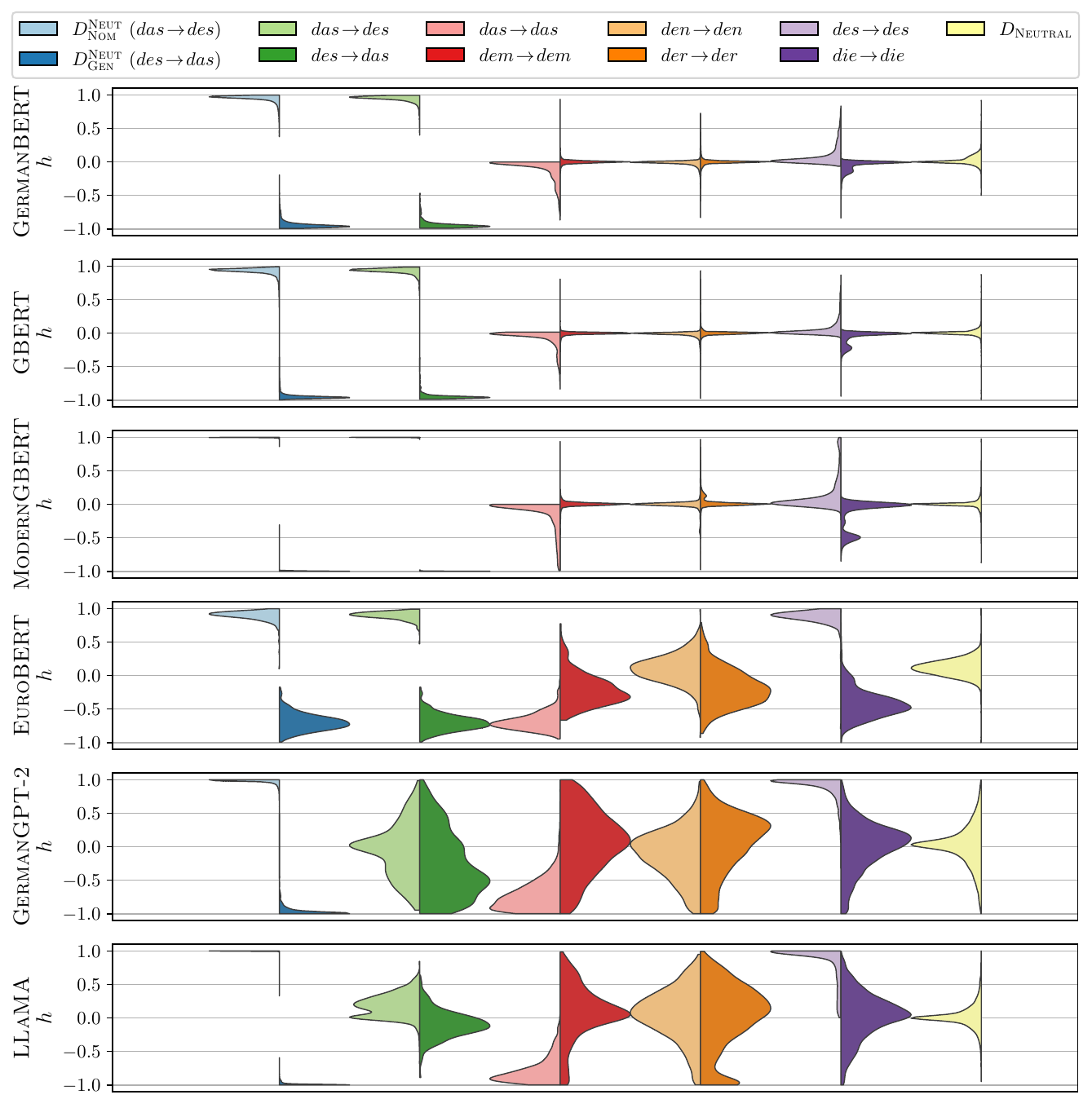}
    \caption{Encoded value distribution of \gradcNGgN\ for different input gradients.}
    \label{fig:encoded-all-models-NG_N}
\end{figure}

\begin{figure}[!p]
    \centering
  \includegraphics[width=\linewidth]{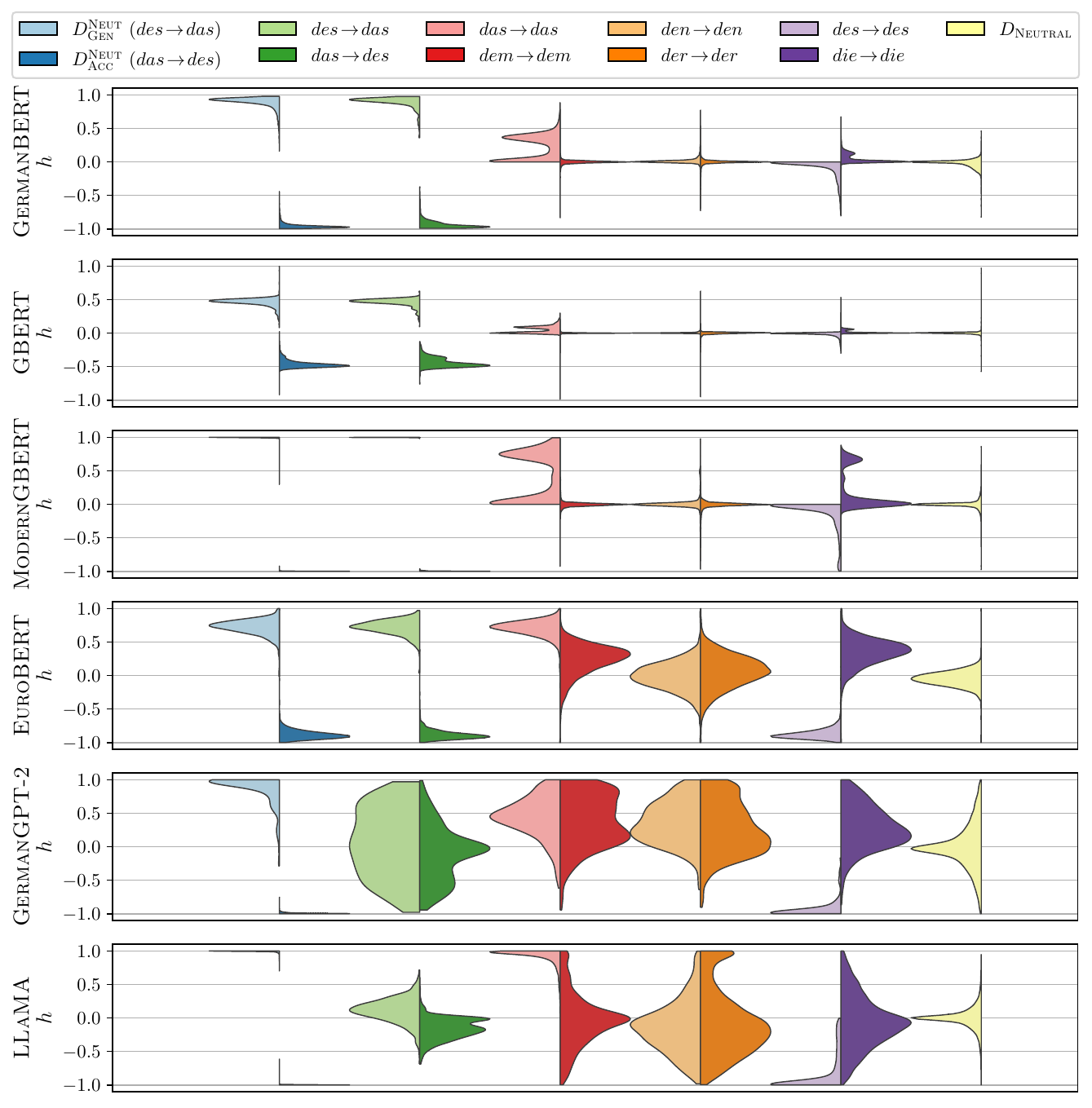}
    \caption{Encoded value distribution of \gradcGAgN\ for different input gradients.}
    \label{fig:encoded-all-models-GA_N}
\end{figure}

\begin{figure}[!p]
    \centering
  \includegraphics[width=\linewidth]{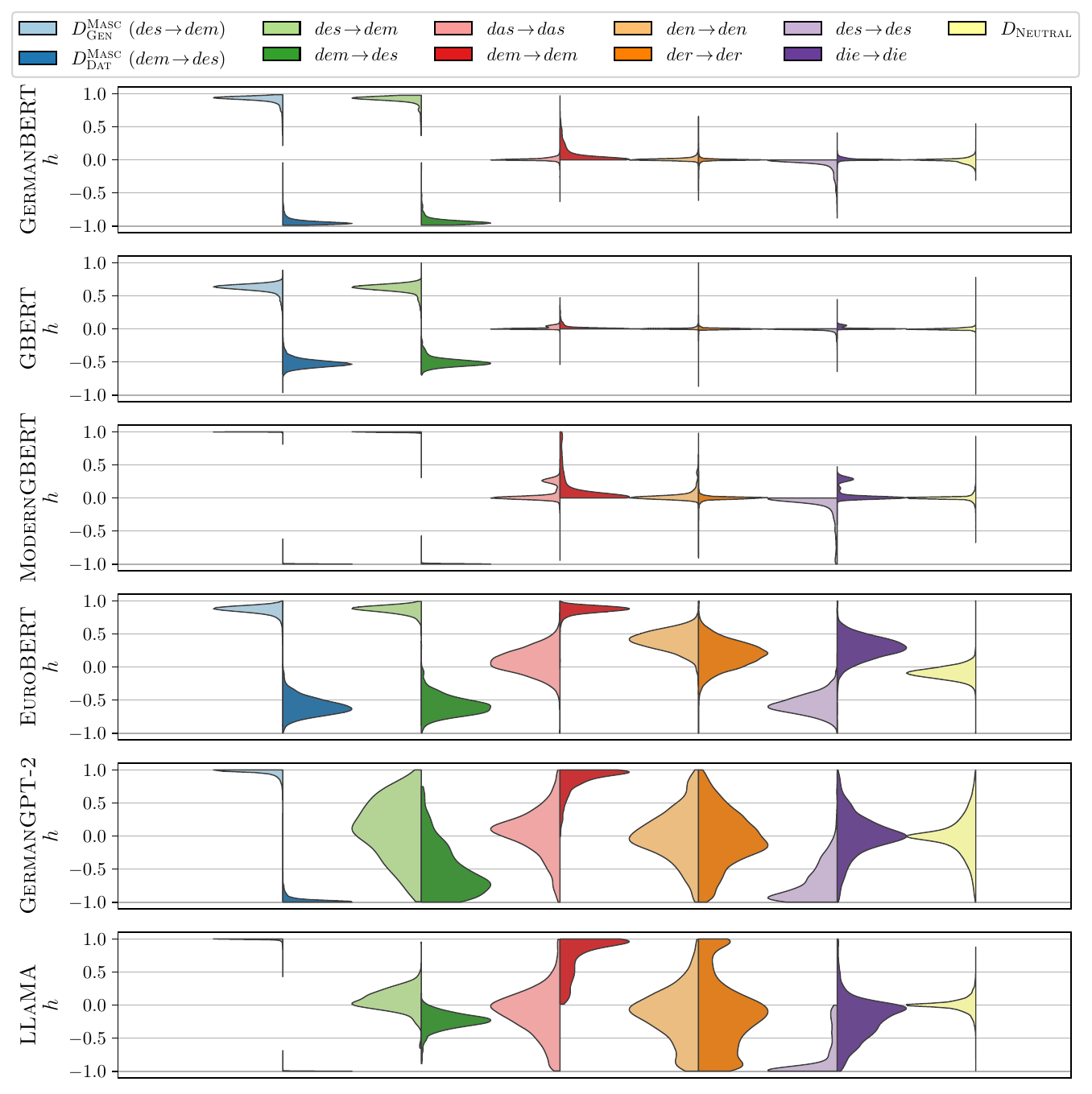}
    \caption{\rev{Encoded value distribution of \gradcGDgM\ for different input gradients.}}
    \label{fig:encoded-all-models-GD_M}
\end{figure}

\begin{figure}[!p]
    \centering
  \includegraphics[width=\linewidth]{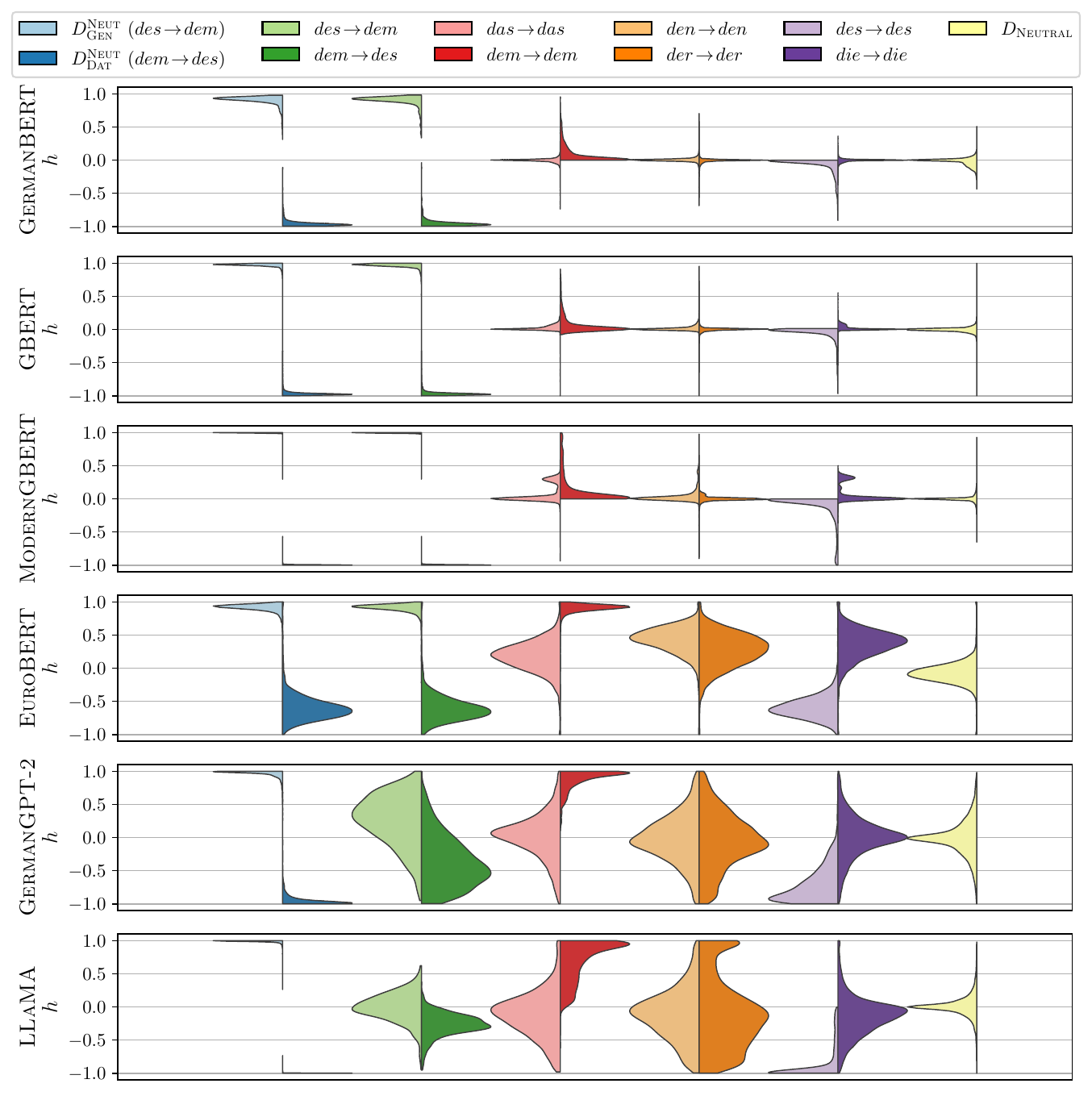}
    \caption{\rev{Encoded value distribution of \gradcGDgN\ for different input gradients.}}
    \label{fig:encoded-all-models-GD_N}
\end{figure}

Figures~\ref{fig:encoded-all-models-ND_F}--\ref{fig:encoded-all-models-GD_N} show encoded-value distributions for the remaining \gradiend\ variants.

Correlations in Table~\ref{tab:encoded-values-correlations} are computed from the same gradient types used during training.
We assign labels $+1$ and $-1$ to the two directed transitions $z_1\,{\to}\,z_2$ and $z_2\,{\to}\,z_1$, respectively, and label the ten identity tasks $\tilde z\,{\to}\,\tilde z$ for $\tilde z\notin\{z_1,z_2\}$ as $0$.
Normalizing the sign of $h$ during training makes correlations non-negative by construction (apart from extreme distribution shifts between the validation set used for normalization and the test set).

Since test splits differ in size across gender-case datasets, we compute correlations after randomly downsampling each dataset to the smallest test-set size.
Likewise, for violins that combine multiple data sources in Figures~\ref{fig:encoded-all-models-N_MF} and~\ref{fig:encoded-all-models-ND_F}--\ref{fig:encoded-all-models-GD_N}, we downsample each source to the smallest subset contributing to that violin.

For \dataNEUT\ on decoder-only models, we use \acrshort{clm} gradients, since the auxiliary \gls{mlm} head is restricted to predicting article tokens.

\section{Probability Analysis}\label{app:probability-analysis}

\begin{table*}[!tb]
    \centering
    \scriptsize
    \setlength{\tabcolsep}{4pt}
    \renewcommand{\arraystretch}{1.05}
    \begin{tabular}{llrrrrrrrrrrrrrrrrr}\toprule
    & & & \multicolumn{3}{c}{\textbf{$\dataNM(der)$}} & \multicolumn{3}{c}{\textbf{$\dataGF(der)$}} & \multicolumn{3}{c}{\textbf{$\dataDF(der)$}} & 
    \multicolumn{3}{c}{\textbf{\dataNEUT}} &
    \\ \cmidrule(lr){4-6} \cmidrule(lr){7-9} \cmidrule(lr){10-12} \cmidrule(lr){13-15}
  \textbf{Model} & \textbf{Art. Trans.} & $\alpha$ &  
   $\Delta \mathbb{P}$ & $d$ & \textbf{Sig.}  &  
   $\Delta \mathbb{P}$ & $d$ & \textbf{Sig.}   &  
   $\Delta \mathbb{P}$ & $d$ & \textbf{Sig.}   &  
   $\Delta \mathbb{P}$ & $d$ & \textbf{Sig.}   & \textbf{\acrshort{supergleber}} \\ \midrule

\gbert & -- & 0.0 & -- & -- & -- & -- & -- & -- & -- & -- & -- & -- & -- & -- & $76.7 \pm 0.4$ \\
\, + \gradcNgMF & $der \to die$ & 0.05 & \textbf{0.13} & \textbf{0.27} & \textbf{***} & 0.01 & 0.11 & *** & 0.02 & 0.14 & *** & -0.02 & -0.00 & n.s. & $76.6 \pm 0.4$  \\
\, + \gradcADgF & $der \to die$ & 1.0 & 0.70 & 0.36 & *** & 0.45 & 0.34 & *** & \textbf{4.20} & \textbf{0.71} & \textbf{***} & 0.12 & 0.03 & * &  $76.6 \pm 0.4$  \\
\, + \gradcGAgF & $der \to die$ & 0.05 & 0.01 & 0.25 & *** & \textbf{0.01} & \textbf{0.17} & \textbf{***} & 0.01 & 0.13 & *** & -0.06 & -0.01 & n.s. & $76.6 \pm 0.4$ \\
\, + \gradcNDgM & $der \to dem$ & 0.01 & \textbf{0.00} & \textbf{0.12} & \textbf{***} & 0.00 & 0.08 & *** & 0.00 & 0.04 & * & -0.00 & -0.00 & n.s. & $76.8 \pm 0.4$ \\
\, + \gradcDgFN & $der \to dem$ & 0.01 & 0.00 & 0.08 & *** & 0.00 & 0.04 & *** & \textbf{0.00} & \textbf{0.16} & \textbf{***} & 0.02 & 0.01 & n.s. & $76.7 \pm 0.4$ \\
\, + \gradcNGgM & $der \to des$ & 0.01 & \textbf{0.00} & \textbf{0.08} & \textbf{***} & 0.02 & 0.22 & *** & 0.00 & 0.07 & *** & -0.00 & -0.01 & n.s. &  $76.7 \pm 0.4$  \\
\, + \gradcGgFN & $der \to des$ & 0.2 & 0.03 & 0.07 & *** & \textbf{0.44} & \textbf{0.32} & \textbf{***} & 0.01 & 0.08 & *** & 0.02 & 0.03 & *** & $76.6 \pm 0.4$  \\

\midrule

\modernbert & -- & 0.0 & -- & -- & -- & -- & -- & -- & -- & -- & -- & -- & -- & -- & -- \\
\, + \gradcNgMF & $der\,{\to}\,die$ & 0.2 & \textbf{10.67} & \textbf{0.69} & \textbf{***} & 3.09 & 0.38 & *** & 7.28 & 0.52 & *** & 1.26 & 0.14 & *** & -- \\
\, + \gradcADgF & $der\,{\to}\,die$ & 0.2 & 8.09 & 0.59 & *** & 4.83 & 0.50 & *** & \textbf{13.73} & \textbf{0.76} & \textbf{***} & 1.26 & 0.14 & *** & -- \\
\, + \gradcGAgF & $der\,{\to}\,die$ & 0.001 & 0.01 & 0.19 & *** & \textbf{0.00} & \textbf{0.11} & \textbf{***} & 0.01 & 0.16 & *** & 0.02 & 0.00 & n.s. & -- \\
\, + \gradcNDgM & $der\,{\to}\,dem$ & 0.2 & \textbf{6.85} & \textbf{0.52} & \textbf{***} & 0.65 & 0.15 & *** & 7.22 & 0.51 & *** & 0.59 & 0.15 & *** & -- \\
\, + \gradcDgFN & $der\,{\to}\,dem$ & 0.2 & 7.15 & 0.47 & *** & 2.17 & 0.23 & *** & \textbf{27.06} & \textbf{1.02} & \textbf{***} & 0.83 & 0.17 & *** & -- \\
\, + \gradcNGgM & $der\,{\to}\,des$ & 0.1 & \textbf{0.57} & \textbf{0.16} & \textbf{***} & 2.62 & 0.33 & *** & 0.11 & 0.09 & *** & 0.13 & 0.06 & *** & -- \\
\, + \gradcGgFN & $der\,{\to}\,des$ & 0.2 & 3.21 & 0.34 & *** & \textbf{26.38} & \textbf{1.17} & \textbf{***} & 1.65 & 0.26 & *** & 0.81 & 0.14 & *** & -- \\

\midrule

\eurobert & -- & 0.0 & -- & -- & -- & -- & -- & -- & -- & -- & -- & -- & -- & -- &  $68.3 \pm 0.4$  \\
\, + \gradcNgMF & $der \to die$ & 0.01 & \textbf{7.36} & \textbf{0.60} & \textbf{***} & 0.25 & 0.11 & *** & 0.48 & 0.21 & *** & 0.01 & 0.01 & n.s. &  $68.4 \pm 0.4$ \\
\, + \gradcADgF & $der \to die$ & 0.01 & 0.17 & 0.34 & *** & 0.08 & 0.20 & *** & \textbf{0.45} & \textbf{0.30} & \textbf{***} & -0.00 & -0.00 & n.s. & $67.0 \pm 0.4$  \\
\, + \gradcGAgF & $der \to die$ & 0.01 & 0.12 & 0.33 & *** & \textbf{0.12} & \textbf{0.23} & \textbf{***} & 0.21 & 0.29 & *** & -0.00 & -0.00 & n.s. & $67.5 \pm 0.4$ \\
\, + \gradcNDgM & $der \to dem$ & 0.01 & \textbf{0.16} & \textbf{0.32} & \textbf{***} & 0.00 & 0.09 & *** & 0.03 & 0.18 & *** & -0.00 & -0.00 & n.s. & $68.4 \pm 0.4$ \\
\, + \gradcDgFN & $der \to dem$ & 0.01 & 0.18 & 0.17 & *** & 0.06 & 0.05 & *** & \textbf{1.28} & \textbf{0.28} & \textbf{***} & 0.00 & 0.04 & *** &  $67.4 \pm 0.4$  \\
\, + \gradcNGgM & $der \to des$ & 0.01 & \textbf{0.60} & \textbf{0.29} & \textbf{***} & 0.44 & 0.30 & *** & 0.03 & 0.09 & *** & 0.00 & 0.08 & *** & $67.0 \pm 0.4$ \\
\, + \gradcGgFN & $der \to des$ & 0.01 & 0.25 & 0.18 & *** & \textbf{2.01} & \textbf{0.36} & \textbf{***} & 0.07 & 0.07 & *** & 0.00 & 0.06 & *** & $68.1 \pm 0.4$  \\

\midrule

\gpttwo & -- & 0.0 & -- & -- & -- & -- & -- & -- & -- & -- & -- & -- & -- & -- & $45.4 \pm 0.4$ \\
\, + \gradcNgMF & $der \to die$ & 0.5 & \textbf{0.31} & \textbf{0.08} & \textbf{***} & 0.15 & 0.05 & ** & 0.25 & 0.08 & *** & -0.00 & -0.03 & *** & $45.3 \pm 0.4$ \\
\, + \gradcADgF & $der \to die$ & 0.5 & 0.20 & 0.04 & * & 0.25 & 0.09 & *** & \textbf{0.24} & \textbf{0.07} & \textbf{***} & 0.00 & 0.06 & *** & $45.3 \pm 0.4$ \\
\, + \gradcGAgF & $der \to die$ & 0.5 & 1.08 & 0.34 & *** & \textbf{1.27} & \textbf{0.46} & \textbf{***} & 1.31 & 0.49 & *** & 0.00 & 0.01 & n.s. & $45.3 \pm 0.4$ \\
\, + \gradcNDgM & $der \to dem$ & 0.1 & \textbf{0.48} & \textbf{0.48} & \textbf{***} & 0.10 & 0.20 & *** & 0.06 & 0.04 & ** & 0.00 & 0.09 & *** & $45.3 \pm 0.4$ \\ 
\, + \gradcDgFN & $der \to dem$ & 1.0 & 0.21 & 0.59 & *** & 0.03 & 0.14 & *** & \textbf{0.07} & \textbf{0.11} & \textbf{***} & 0.00 & 0.06 & *** & $45.3 \pm 0.4$ \\
\, + \gradcNGgM & $der \to des$ & 0.1 & \textbf{0.47} & \textbf{0.27} & \textbf{***} & 0.49 & 0.11 & *** & 0.53 & 0.19 & *** & 0.00 & 0.02 & * & $45.3 \pm 0.4$ \\
\, + \gradcGgFN & $der \to des$ & 0.5 & 0.43 & 0.23 & *** & \textbf{1.39} & \textbf{0.27} & \textbf{***} & 0.61 & 0.20 & *** & 0.00 & 0.02 & ** & $45.3 \pm 0.4$	   \\

\midrule

\llama & -- & 0.0 & -- & -- & -- & -- & -- & -- & -- & -- & -- & -- & -- & -- & --\\
\, + \gradcNgMF & $der \to die$ & 0.2 & \textbf{0.03} & \textbf{0.10} & \textbf{***} & 0.00 & 0.04 & *** & 0.02 & 0.06 & *** & 0.00 & 0.12 & *** & -- \\
\, + \gradcADgF & $der \to die$ & 0.2 & 0.03 & 0.10 & *** & 0.00 & 0.04 & *** & \textbf{0.03} & \textbf{0.07} & \textbf{***} & 0.00 & 0.11 & *** & --\\
\, + \gradcGAgF & $der \to die$ & 0.2 & 0.04 & 0.10 & *** & \textbf{0.00} & \textbf{0.04} & \textbf{***} & 0.03 & 0.07 & *** & 0.00 & 0.11 & *** & --\\
\, + \gradcNDgM & $der \to dem$ & 0.2 & \textbf{0.01} & \textbf{0.09} & \textbf{***} & 0.00 & 0.07 & *** & 0.02 & 0.09 & *** & 0.00 & 0.06 & *** & -- \\
\, + \gradcDgFN & $der \to dem$ & 0.5 & 0.03 & 0.09 & *** & 0.00 & 0.07 & *** & \textbf{0.09} & \textbf{0.08} & \textbf{***} & 0.01 & 0.11 & *** & --\\
\, + \gradcNGgM & $der \to des$ & 0.2 & \textbf{0.00} & \textbf{0.01} & \textbf{n.s.} & 0.03 & 0.08 & *** & 0.00 & 0.03 & * & 0.00 & 0.05 & *** & -- \\
\, + \gradcGgFN & $ der \to des $ & 0.2 & 0.01 & 0.25 & *** & \textbf{0.09} & \textbf{0.15} & \textbf{***} & 0.02 & 0.12 & *** & 0.00 & 0.07 & *** & --\\
         \bottomrule
    \end{tabular}
    \caption{\gradiend-modified models for non-\bert\ models: mean change in target-article probability $\Delta\mathbb{P}$ scaled by 100, effect size (Cohen's $d$), and significance as *** $p<.001$, ** $p<.01$, * $p<.05$ (n.s. otherwise). Bold marks datasets being part of \gradiend\ gender-case cells. \acrshort{supergleber} score is scaled by 100, and only reported for small models ($<$1B).}
    \label{tab:prob-changes-other-models}
\end{table*}

\begin{figure*}[!p]
    \centering
    \includegraphics[width=\linewidth]{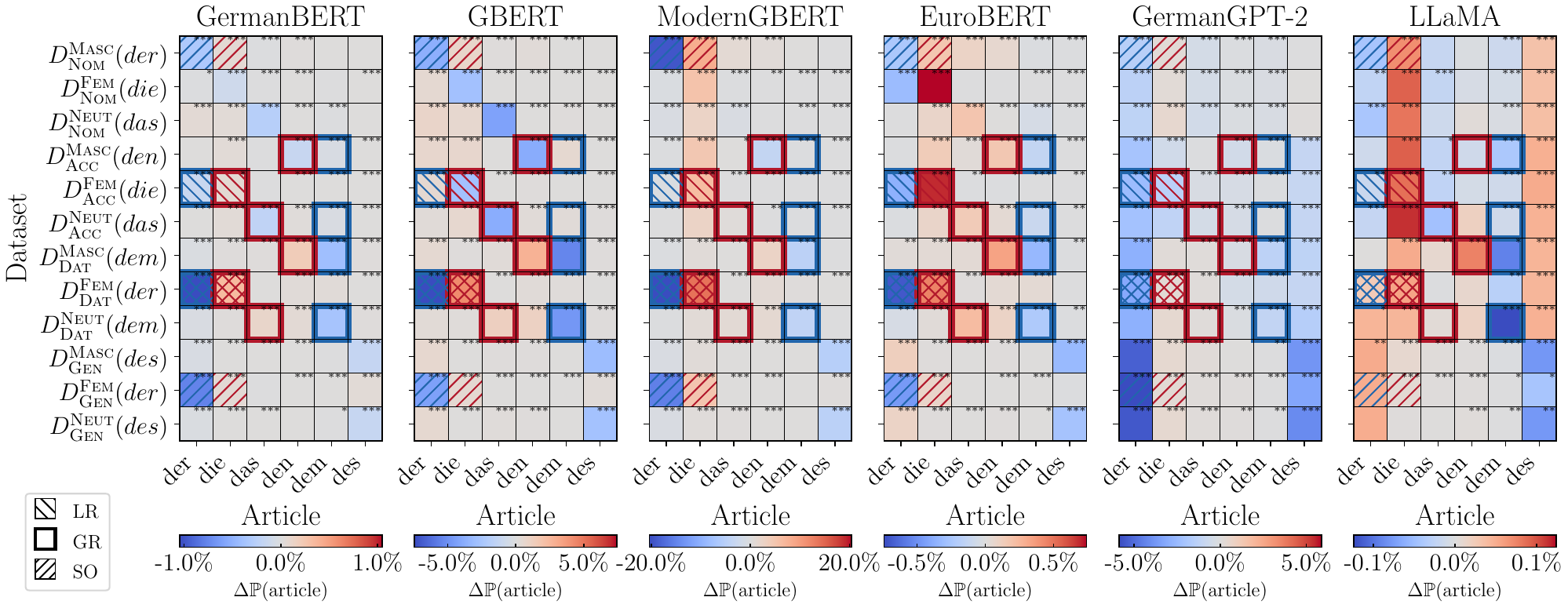}
  \caption{Mean probability change of articles between \gradiend-modified and base model for \gradcADgF\ $der\,{\to}\,die$. 
   Stars mark statistical significance after Benjamini-Hochberg FDR correction applied per model. 
    Marked cells are expectations for LR, GR, and SO (Figure~\ref{fig:running-example-cases}). 
    }
    \label{fig:heatmap-AD_F}
\end{figure*}

\begin{figure*}[!p]
    \centering
    \includegraphics[width=\linewidth]{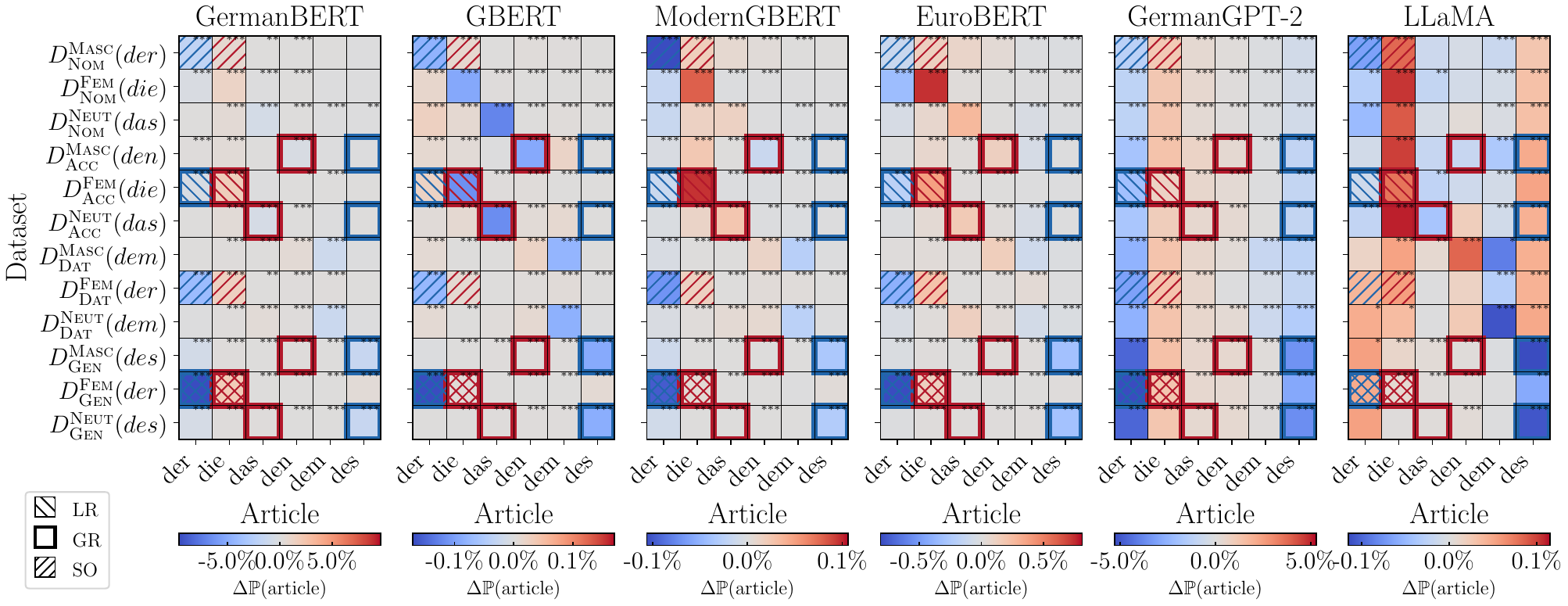}
  \caption{Mean probability change of articles between \gradiend-modified and base model for \gradcGAgF\ $der\,{\to}\,die$. 
   Stars mark statistical significance after Benjamini-Hochberg FDR correction applied per model. 
    Marked cells are expectations for LR, GR, and SO (Figure~\ref{fig:running-example-cases}). 
    }
    \label{fig:heatmap-GA_F}
\end{figure*}

\begin{figure*}[!p]
    \centering
    \includegraphics[width=\linewidth]{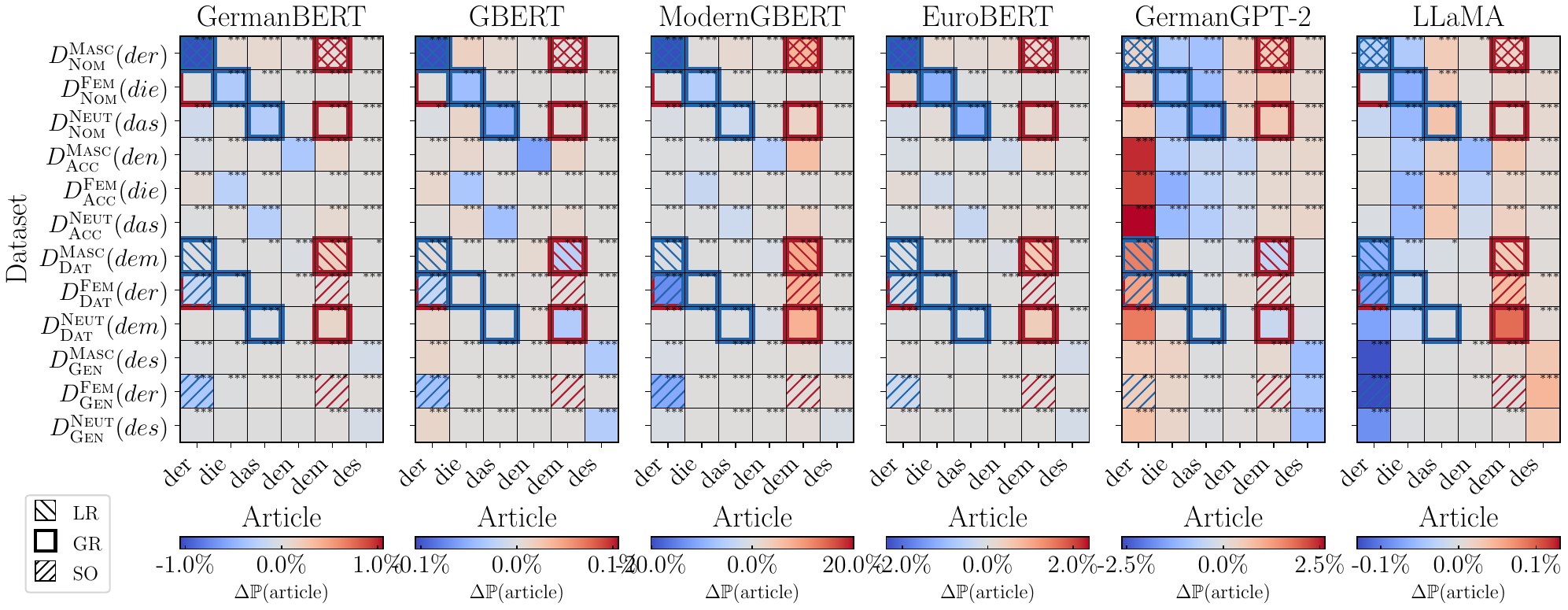}
  \caption{Mean probability change of articles between \gradiend-modified and base model for \gradcNDgM\ $der\,{\to}\,dem$. 
   Stars mark statistical significance after Benjamini-Hochberg FDR correction applied per model. 
    Marked cells are expectations for LR, GR, and SO (Figure~\ref{fig:running-example-cases}). 
    }
    \label{fig:heatmap-ND_M}
\end{figure*}

\begin{figure*}[!p]
    \centering
    \includegraphics[width=\linewidth]{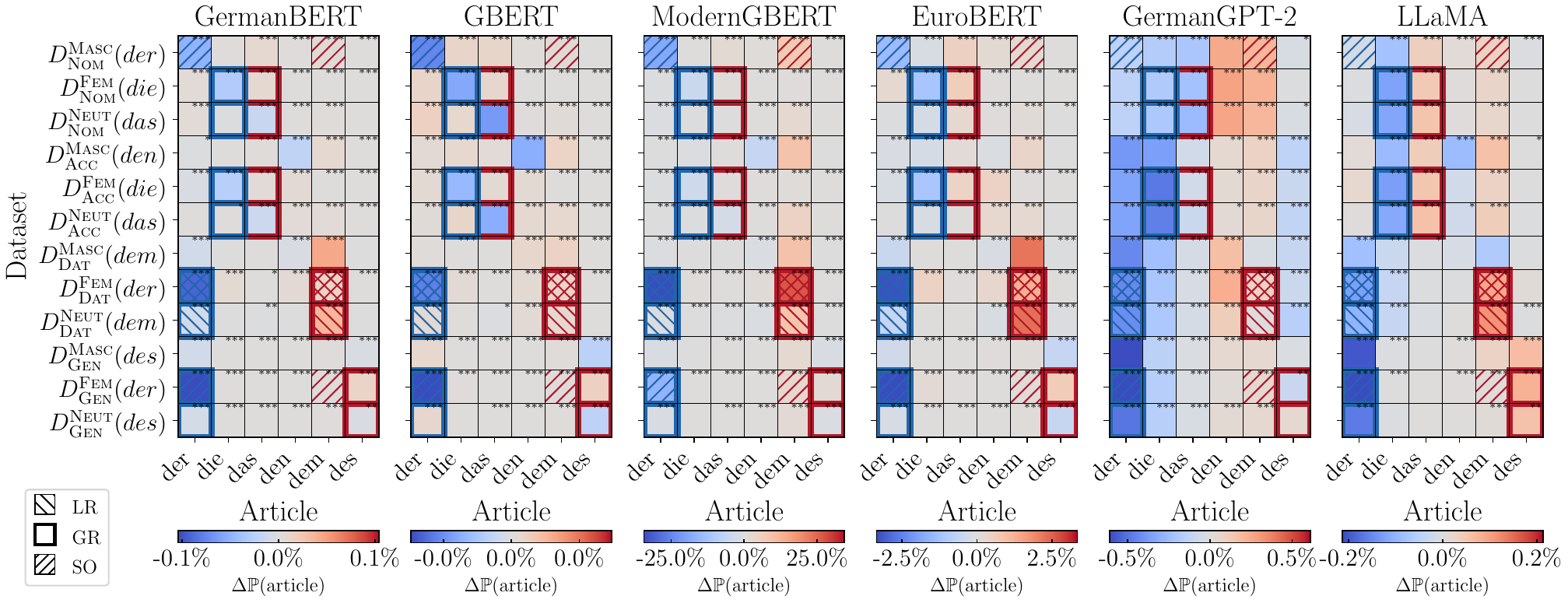}
  \caption{Mean probability change of articles between \gradiend-modified and base model for \gradcDgFN\ $der\,{\to}\,dem$. 
   Stars mark statistical significance after Benjamini-Hochberg FDR correction applied per model. 
    Marked cells are expectations for LR, GR, and SO (Figure~\ref{fig:running-example-cases}). 
    }
    \label{fig:heatmap-D_FN}
\end{figure*}

\begin{figure*}[!p]
    \centering
    \includegraphics[width=\linewidth]{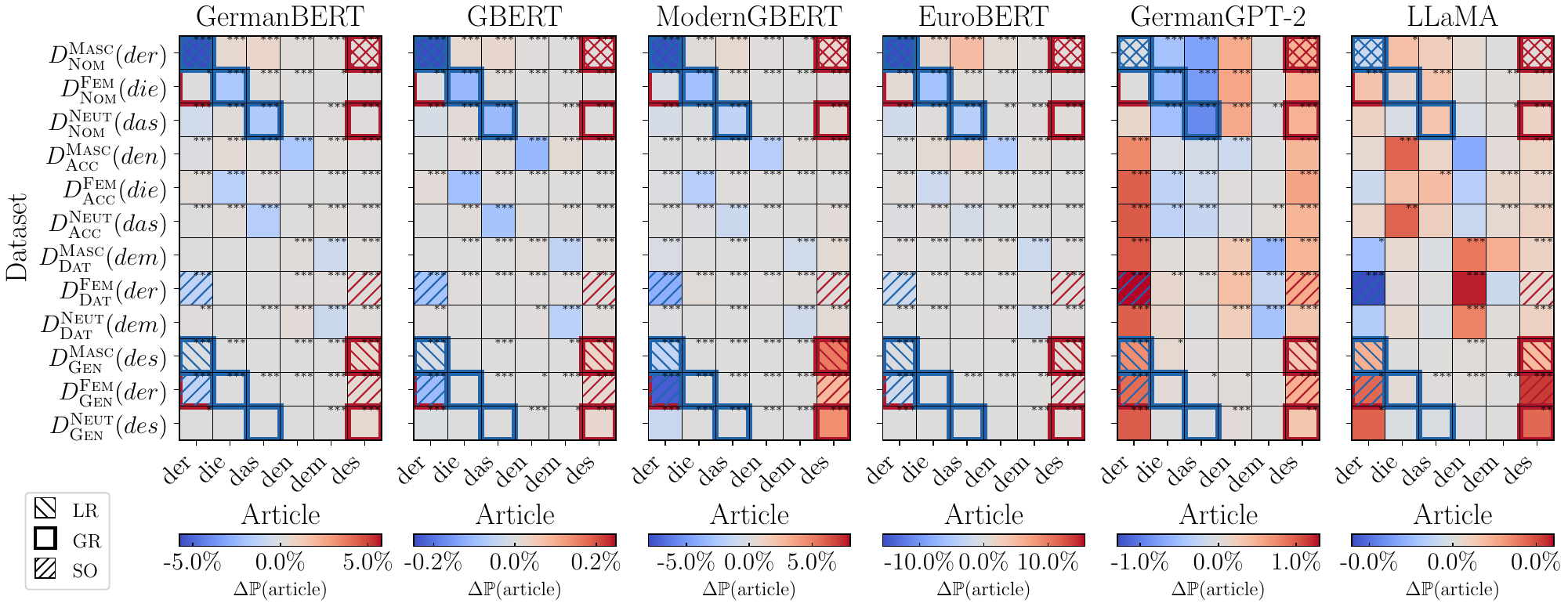}
  \caption{Mean probability change of articles between \gradiend-modified and base model for \gradcNGgM\ $der\,{\to}\,des$. 
   Stars mark statistical significance after Benjamini-Hochberg FDR correction applied per model. 
    Marked cells are expectations for LR, GR, and SO (Figure~\ref{fig:running-example-cases}). 
    }
    \label{fig:heatmap-NG_M}
\end{figure*}

\begin{figure*}[!p]
    \centering
    \includegraphics[width=\linewidth]{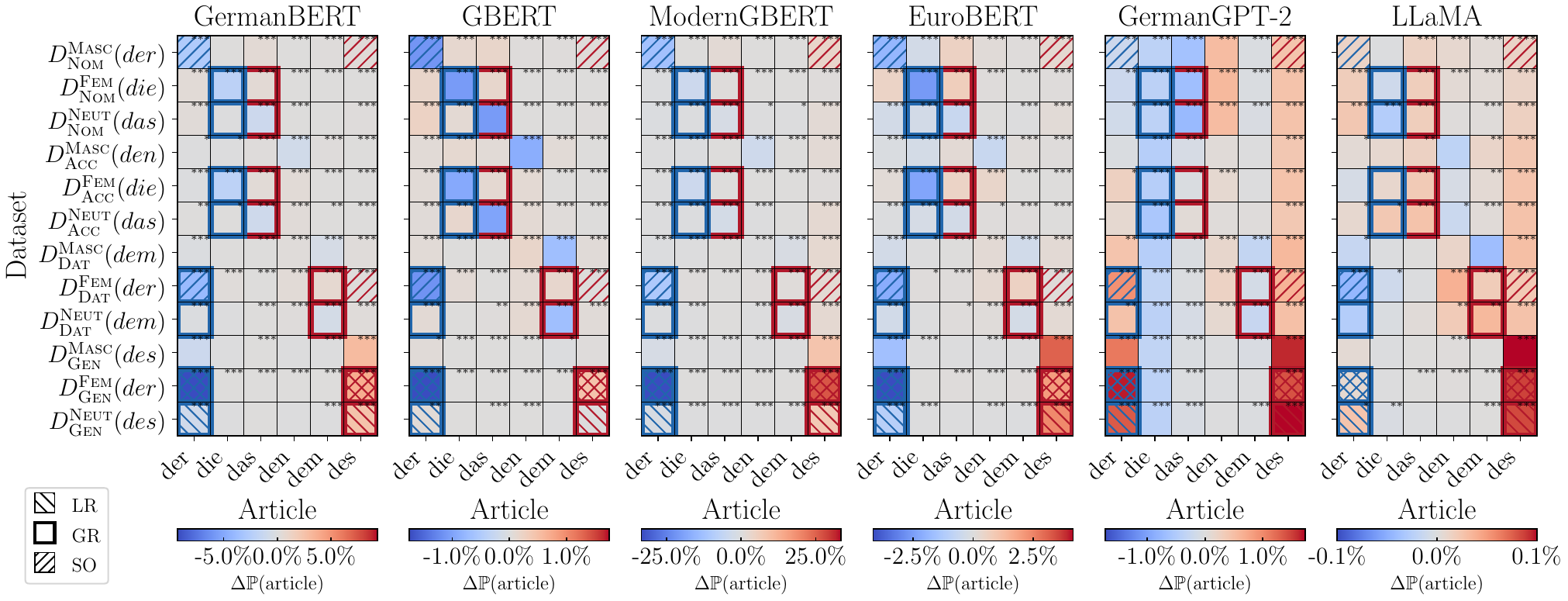}
  \caption{Mean probability change of articles between \gradiend-modified and base model for \gradcGgFN\ $der\,{\to}\,des$. 
   Stars mark statistical significance after Benjamini-Hochberg FDR correction applied per model. 
    Marked cells are expectations for LR, GR, and SO (Figure~\ref{fig:running-example-cases}). 
    }
    \label{fig:heatmap-G_FN}
\end{figure*}

Table~\ref{tab:prob-changes-other-models} and Figures~\ref{fig:heatmap-AD_F}--\ref{fig:heatmap-G_FN} show results for other models and variants reported in the main part of the paper.
Due to computational constraints, we limit the \gls{supergleber} evaluation to models with fewer than 1B parameters, and therefore exclude \modernbert\ and \llama.

\subsection{Details on Probability Calculation}

For an entry $x$ from a fixed gender-case dataset (clear from context) with a single article mask, 
let $\mathbb{P}_m(art|x)$ denote the MLM/CLM probability that model $m$ assigns to the article $art\in\mathcal{A}$ at the masked position for model $m$ (treating tokens as equal up to casing and leading whitespace). 
We define the mean article probability $\mathbb{P}_m(art)$ as the dataset average of $\mathbb{P}_m(art|x)$ over all single mask entries of the dataset.
For two models $m_1, m_2$, we define the mean probability change as 
\[
\Delta \mathbb{P}_{m_1, m_2} (art) = \mathbb{P}_{m_1}(art) - \mathbb{P}_{m_2}(art).
\]
For this study, $m_1$ is the $\alpha^\star$ selected \gradiend-modified model (clear from context) and $m_2$ the base model, so we write $\Delta \mathbb{P}(art)$ or simply $\Delta \mathbb{P}$ when the article is clear from context.

\subsection{Selection of the Intervention Strength $\alpha$.}\label{app:alpha-selection}
We evaluate a discrete set of step sizes $\alpha > 0$ that induce \gradiend-modified models at $\alpha$ (indicated in Figure~\ref{fig:model-modification}).
Let $s_0$ denote the \gls{lms} of the unmodified base model on the grammar-neutral dataset \dataNEUT.
For encoder-only models, this score corresponds to masked-token accuracy, while for decoder-only models it corresponds to perplexity (lower is better).

We define a tolerance threshold $\tau = 0.99$ and select $\alpha^\star$ according to the following procedure.
Among all evaluated step sizes, we retain those as \emph{candidates} whose score satisfies the constraint $s(\alpha) \geq \tau \cdot s_0$ for accuracy-based metrics or
$s(\alpha) \leq s_0 / \tau$ for perplexity-based metrics.
This candidate range is shaded in Figure~\ref{fig:model-modification}. 
Among candidates, we select the $\alpha$, whose score is the largest: $\alpha^\star \,{=}\, \mathrm{argmax}~ \mathbb{P}_\alpha(target)$.

\subsection{SuperGLEBer}

We evaluate downstream language understanding using the \acrlong{supergleber} (\acrshort{supergleber}; \citealt{supergleberr}), a German NLP benchmark covering multiple classification and inference tasks.
We follow the standard implementation provided by \citet{supergleberr}.

We note that for some model-task combinations, the Named Entity Recognition (NER) components yield \emph{empty entity predictions}, resulting in zero precision and recall, a known degeneracy in sequence labeling.
We attribute this to a dependency incompatibility in our evaluation pipeline, since it occurs consistently for both the base model and the corresponding \gradiend-modified variants. 

We extend their evaluation code to support bootstrap-based uncertainty estimation \cite{davison1997bootstrap}.
Our procedure mirrors the approach used by \citet{gradiend} for GLUE \cite{glue} and SuperGLUE \cite{superglue}, computing 95\% confidence intervals via resampling.

\section{Top-$k$ Analysis}\label{app:topk}

\begin{figure*}[!t]
    \centering

 \begin{subfigure}[t]{\linewidth}
    \includegraphics[width=0.18\linewidth]{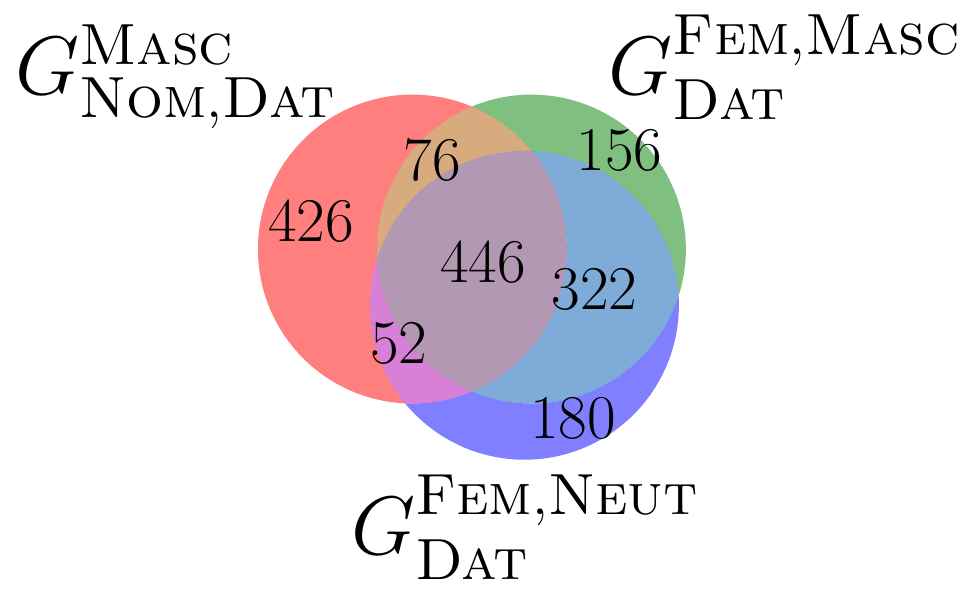}
    \includegraphics[width=0.18\linewidth]{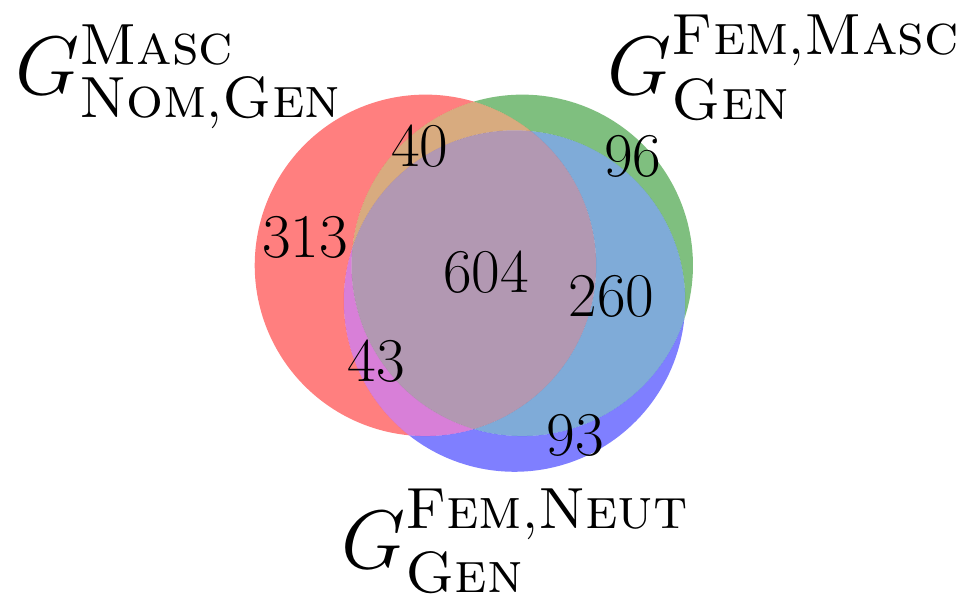}
    \includegraphics[width=0.14\linewidth]{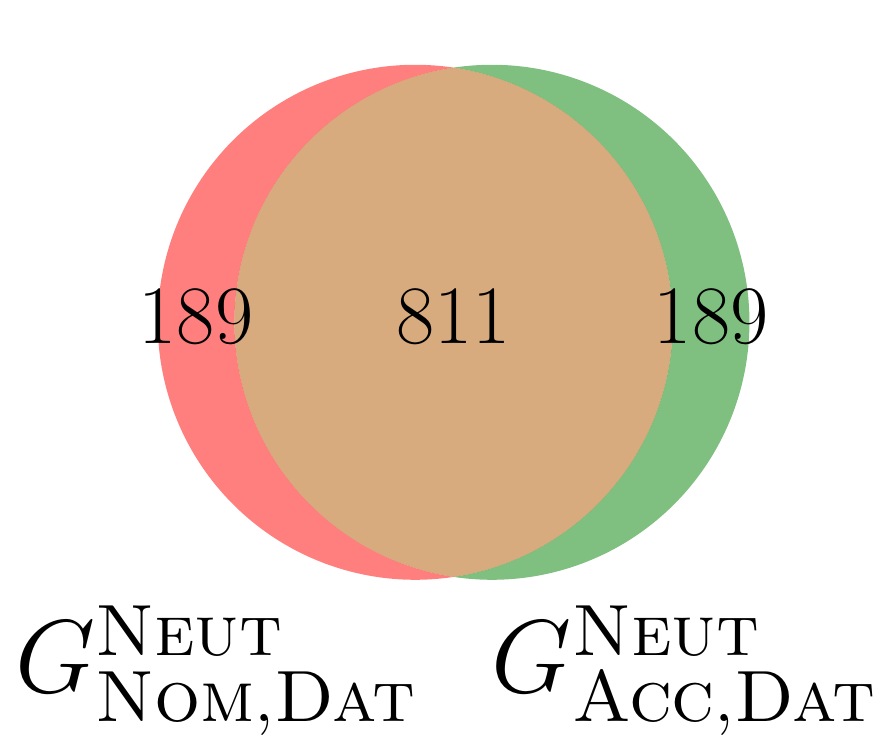}
    \includegraphics[width=0.14\linewidth]{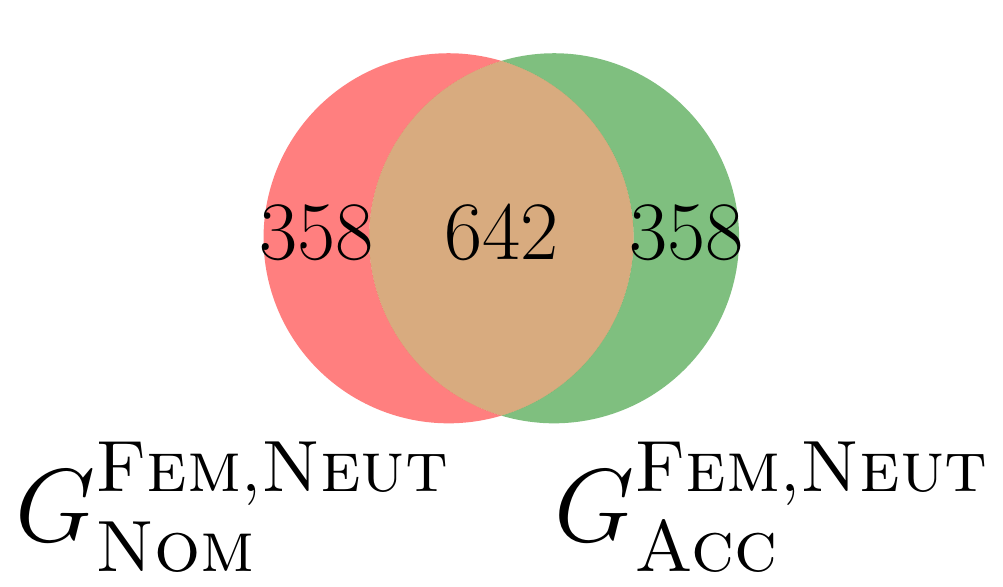}
    \includegraphics[width=0.14\linewidth]{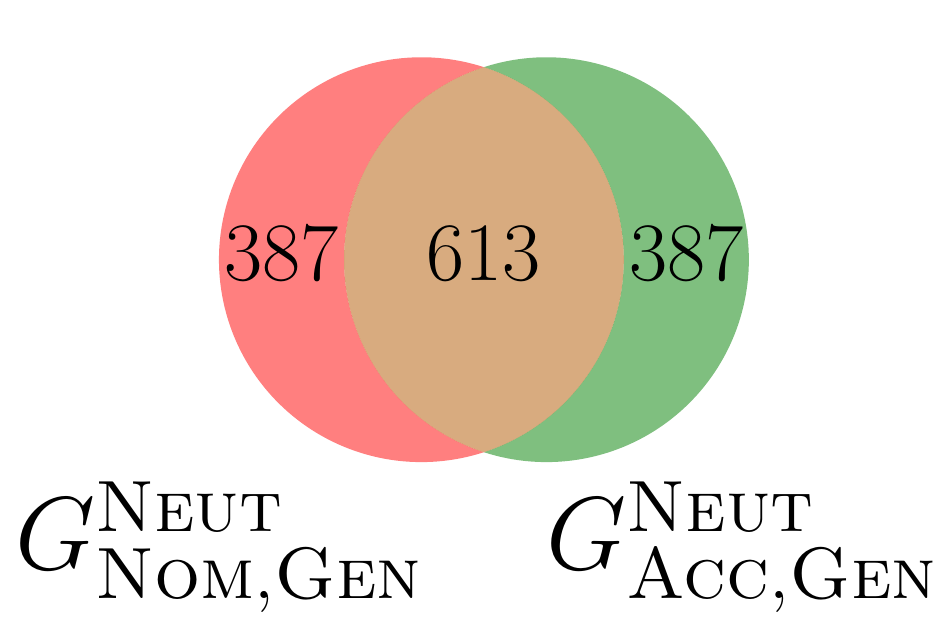}
    \revbg{
  \includegraphics[width=0.14\linewidth]{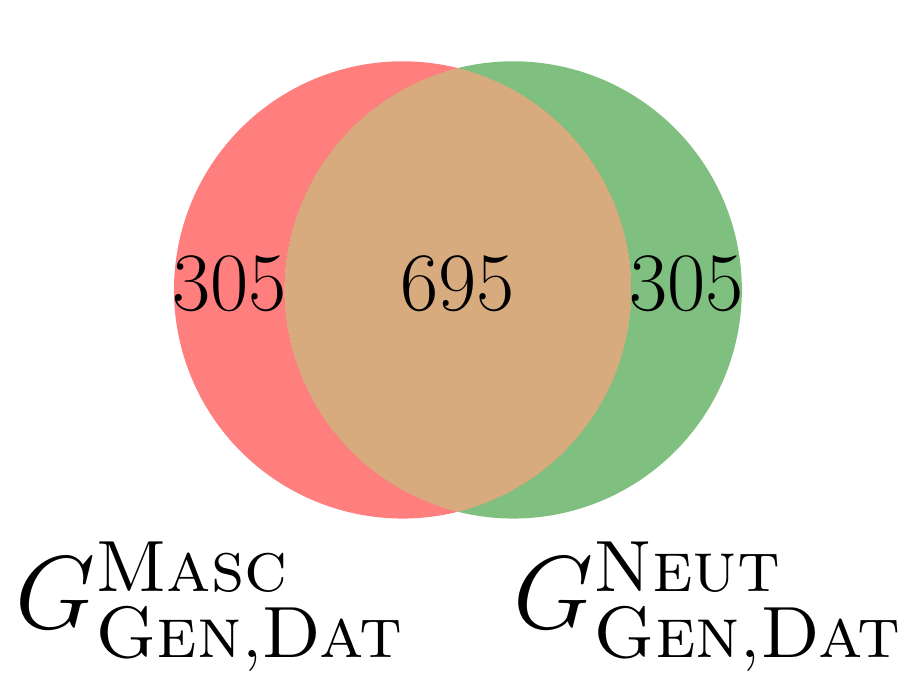}
}
    \caption{\gbert.}
\end{subfigure}

 \begin{subfigure}[t]{\linewidth}
    \includegraphics[width=0.18\linewidth]{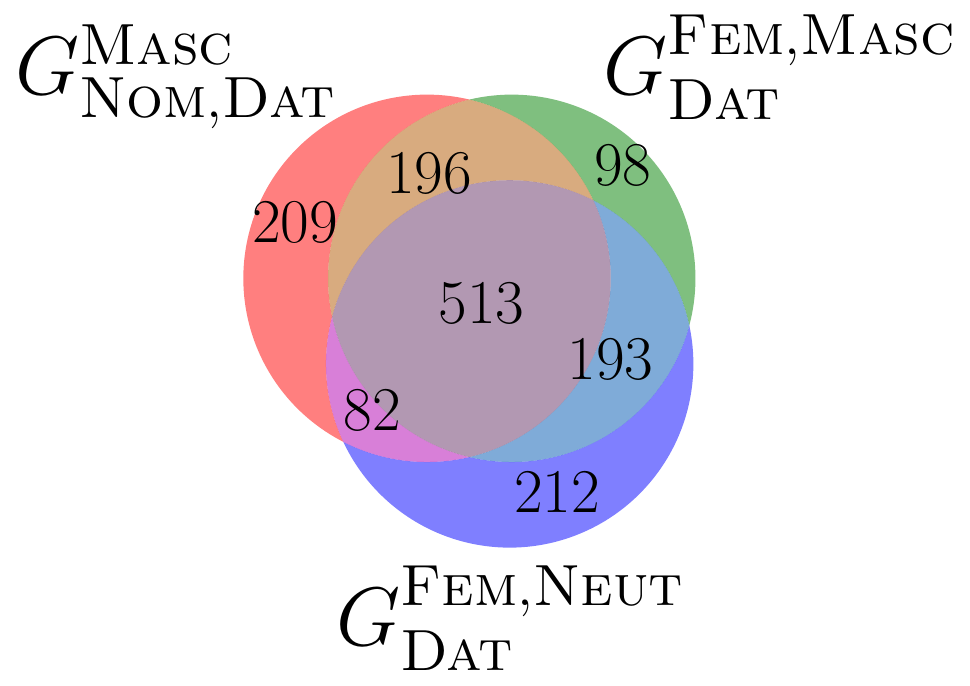}
    \includegraphics[width=0.18\linewidth]{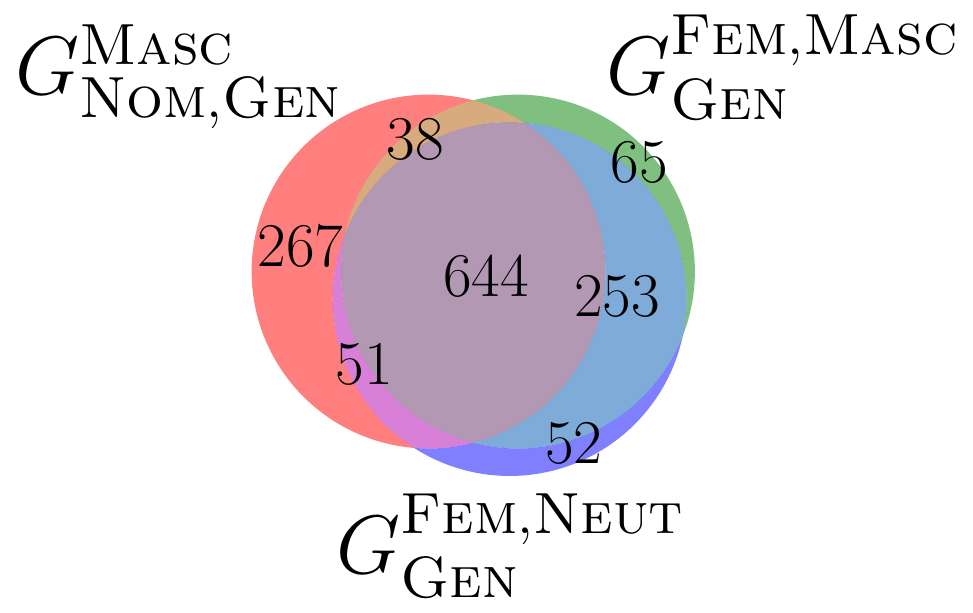}
    \includegraphics[width=0.14\linewidth]{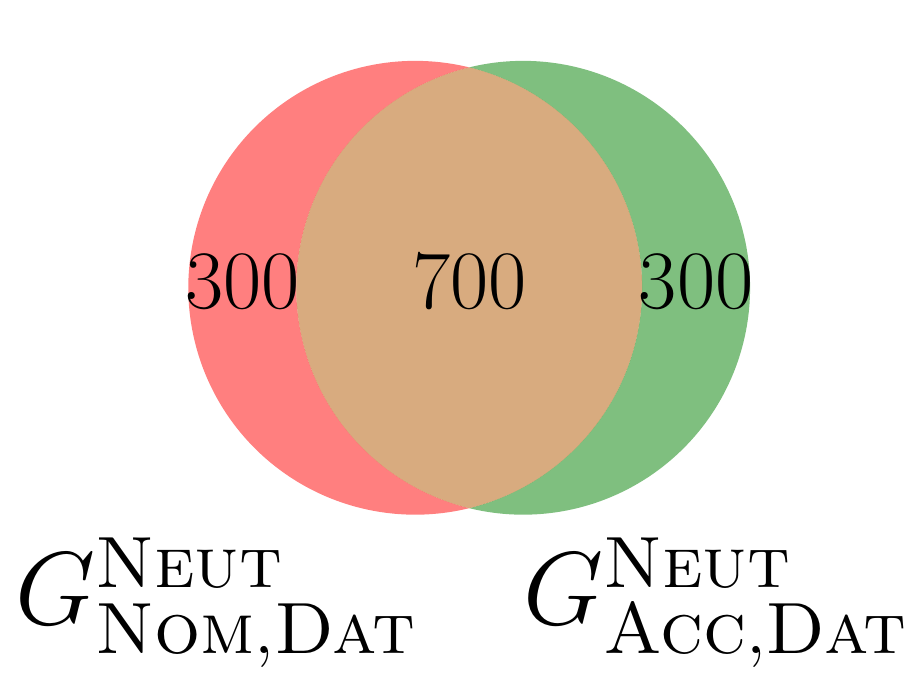}
    \includegraphics[width=0.14\linewidth]{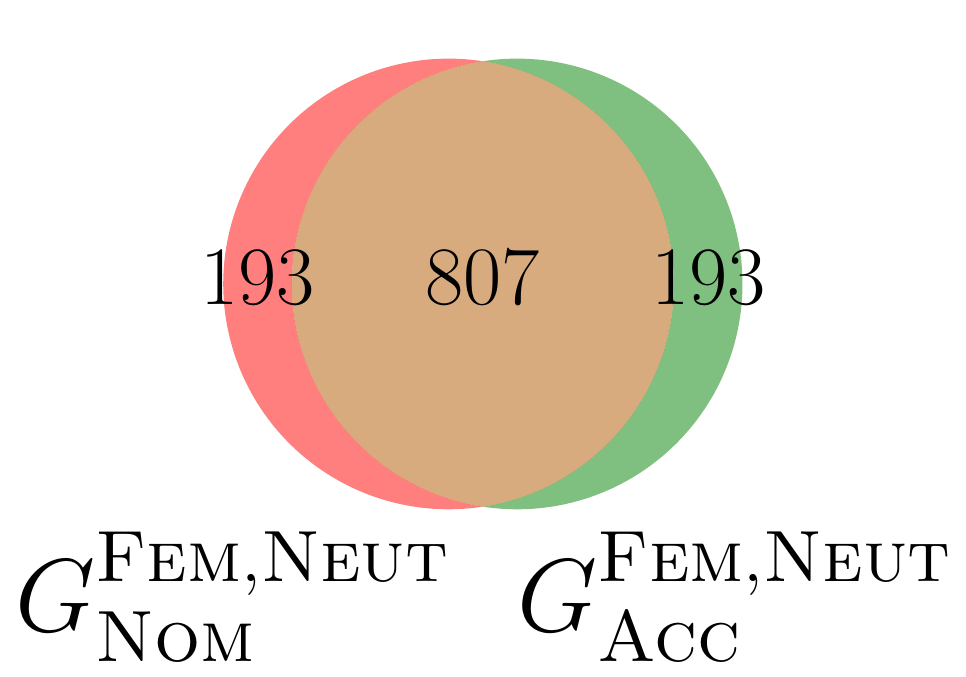}
    \includegraphics[width=0.14\linewidth]{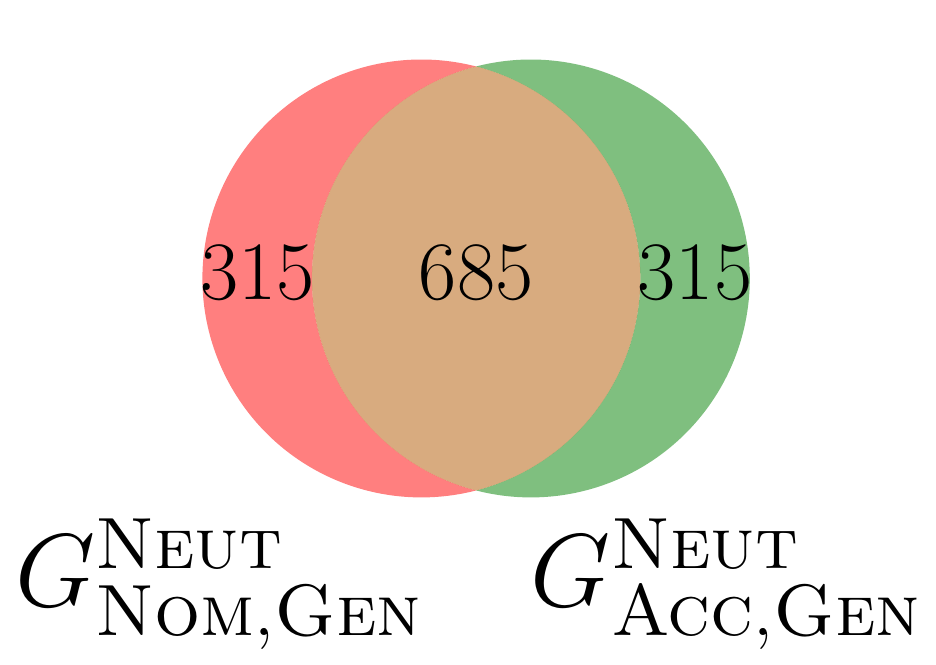}
    \revbg{
  \includegraphics[width=0.14\linewidth]{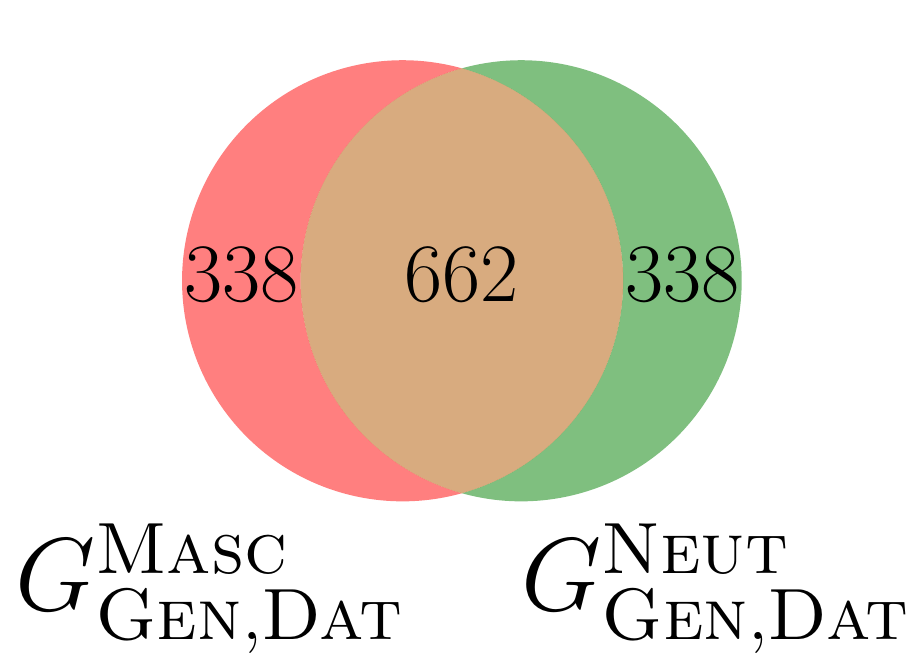}
}
    \caption{\modernbert.}
\end{subfigure}

 \begin{subfigure}[t]{\linewidth}
    \includegraphics[width=0.18\linewidth]{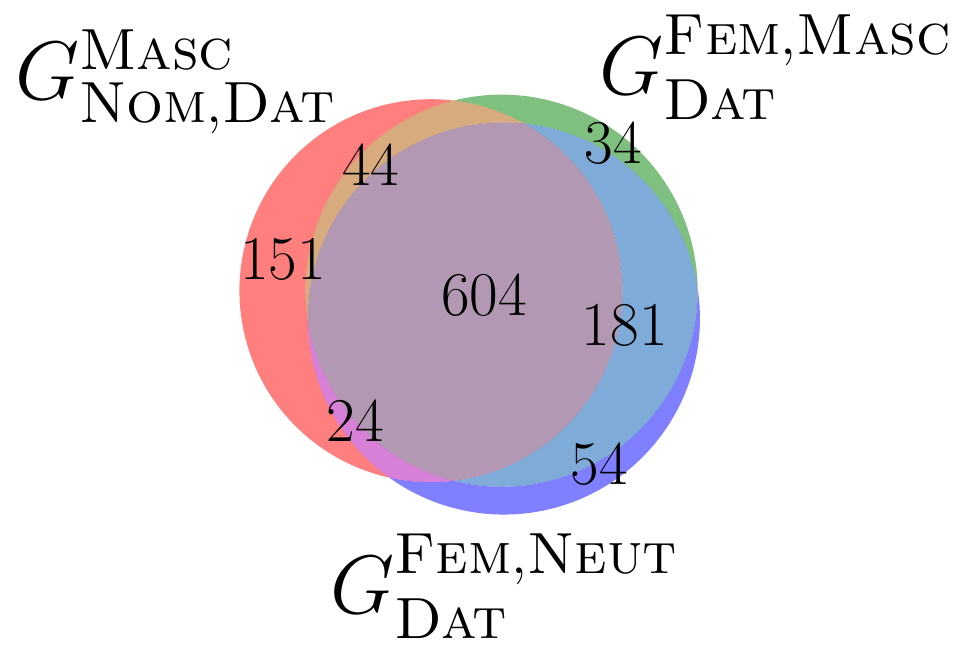}
    \includegraphics[width=0.18\linewidth]{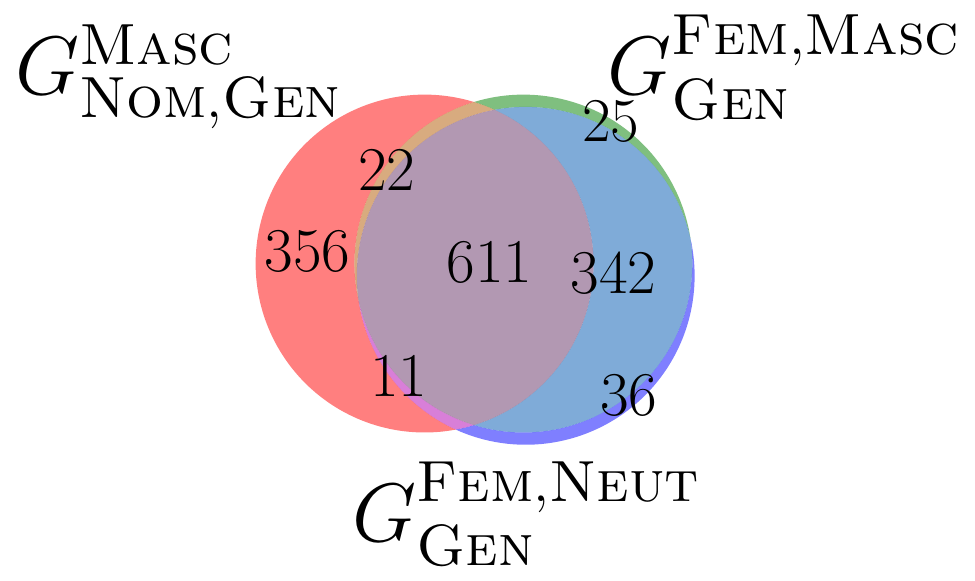}
    \includegraphics[width=0.14\linewidth]{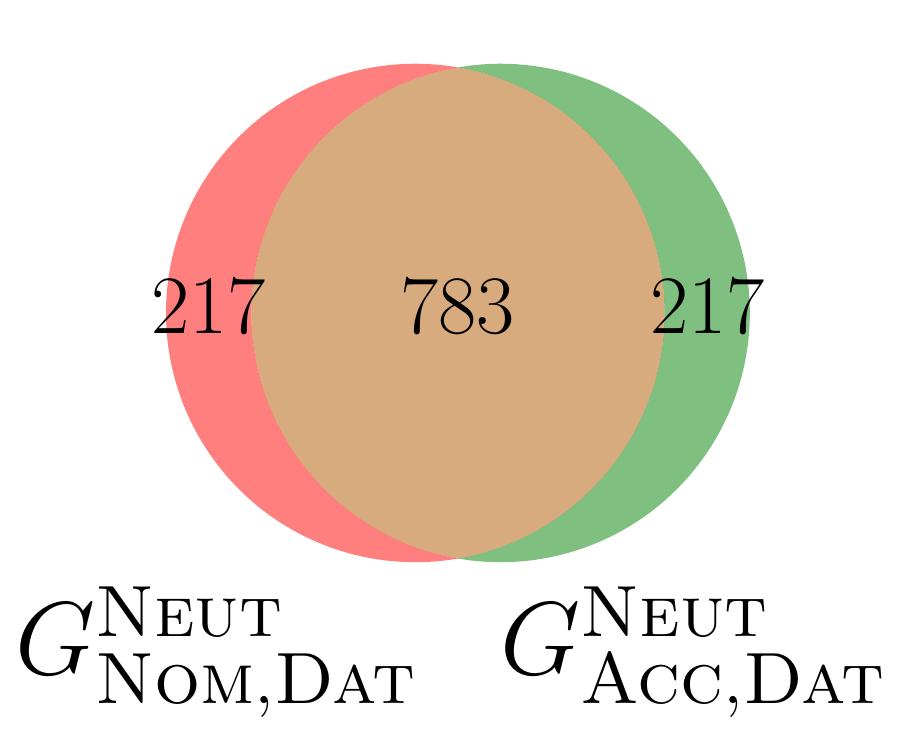}
    \includegraphics[width=0.14\linewidth]{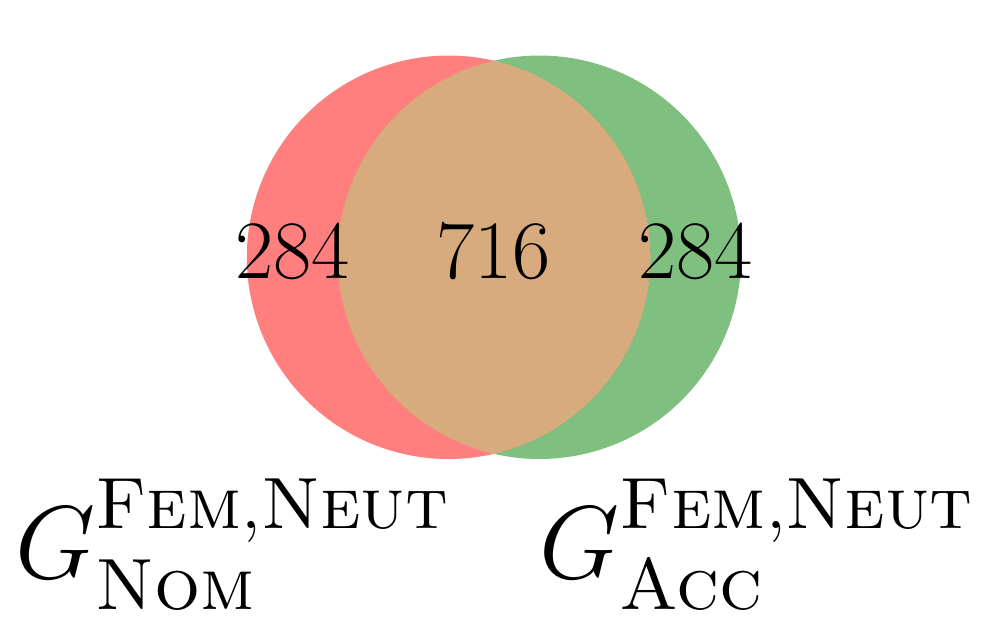}
    \includegraphics[width=0.14\linewidth]{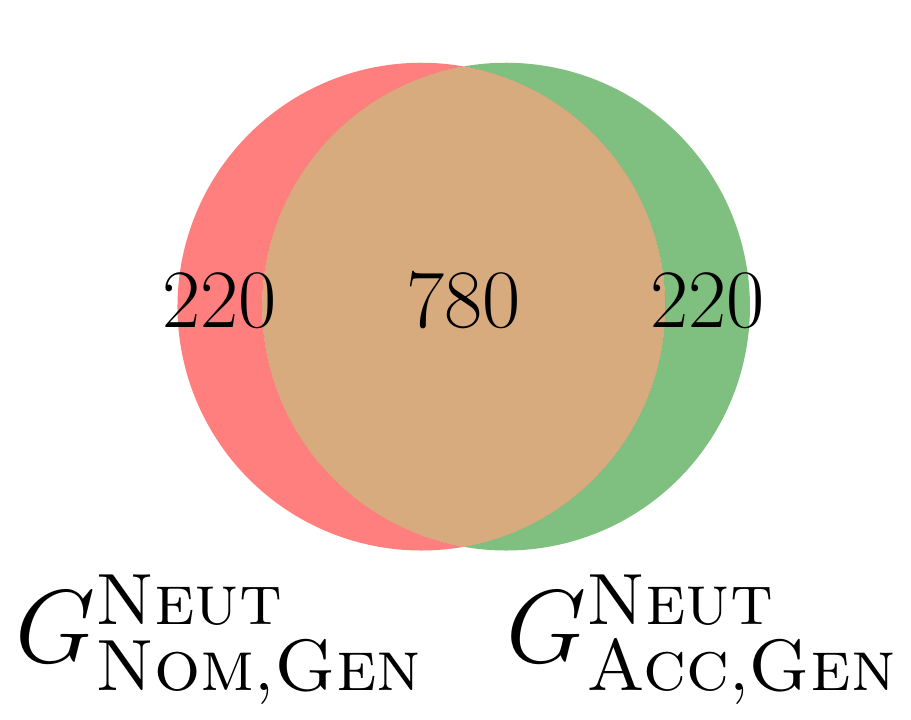}
    \revbg{
  \includegraphics[width=0.14\linewidth]{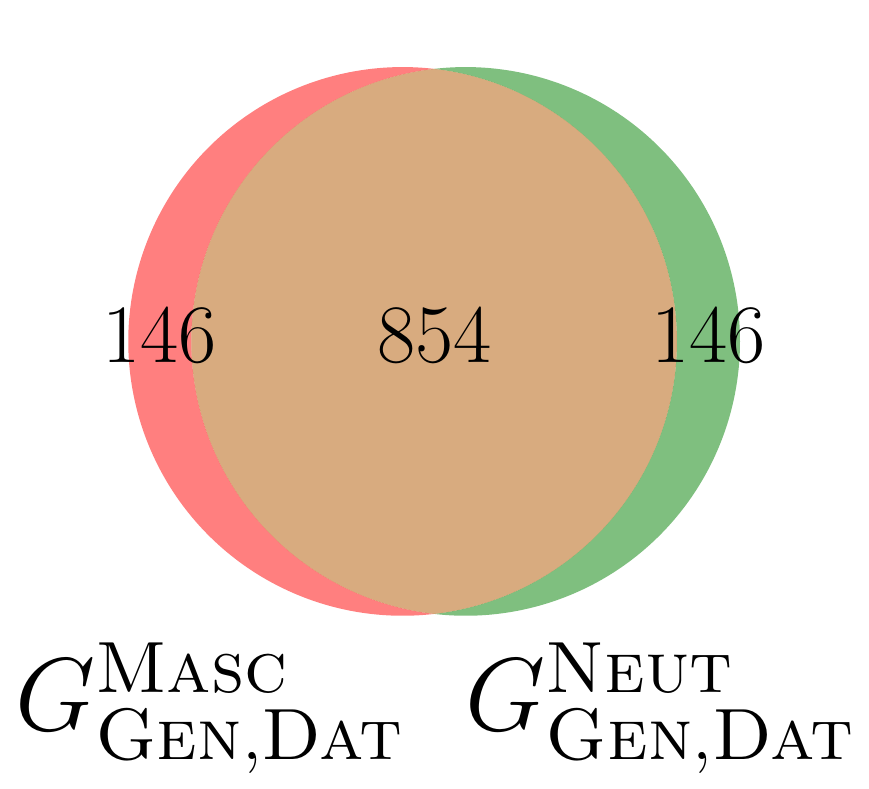}
}
    \caption{\eurobert.}
\end{subfigure}

 \begin{subfigure}[t]{\linewidth}
    \includegraphics[width=0.18\linewidth]{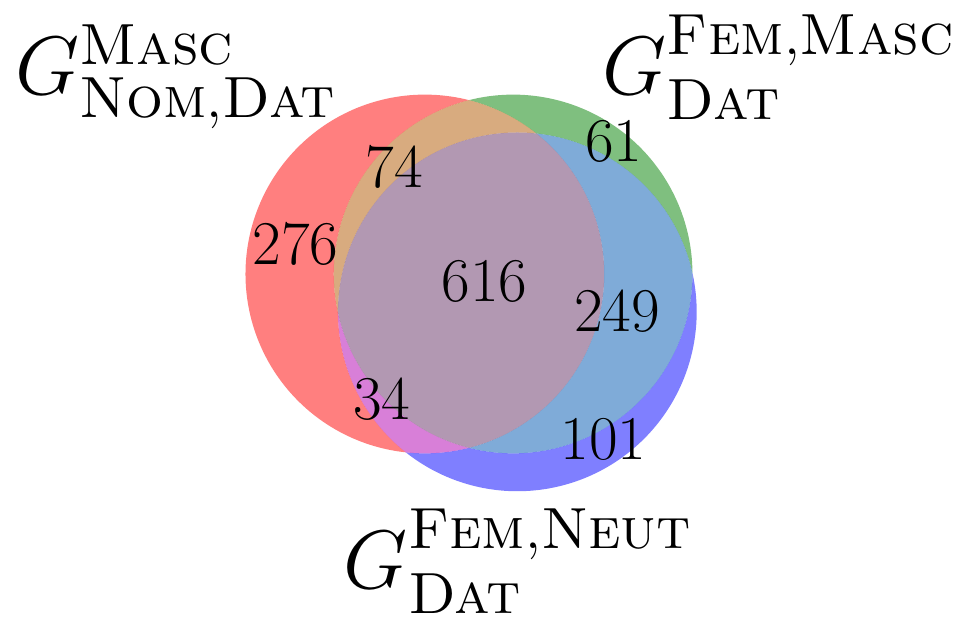}
    \includegraphics[width=0.18\linewidth]{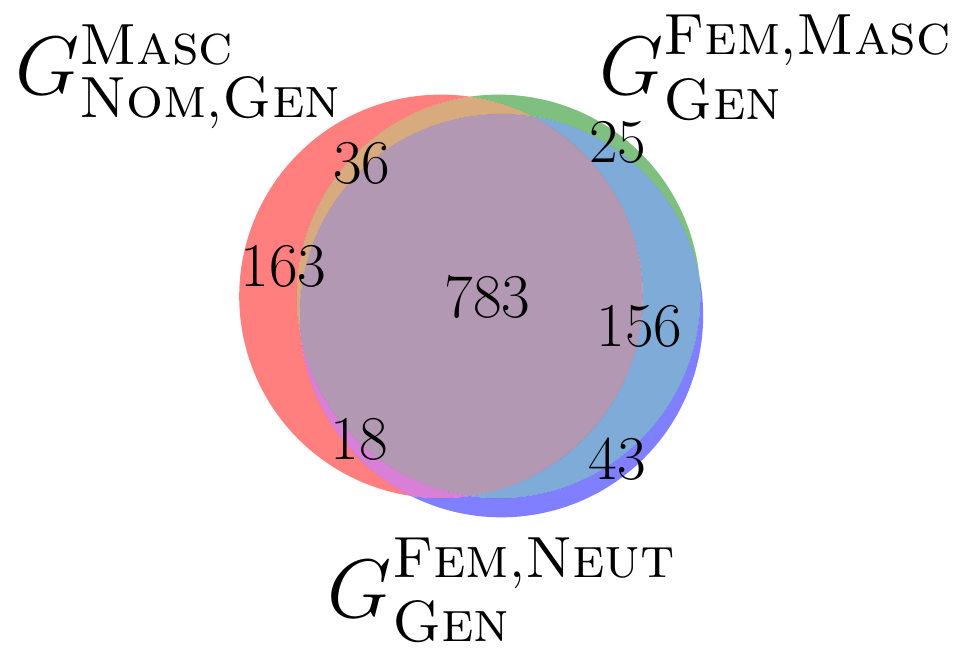}
    \includegraphics[width=0.14\linewidth]{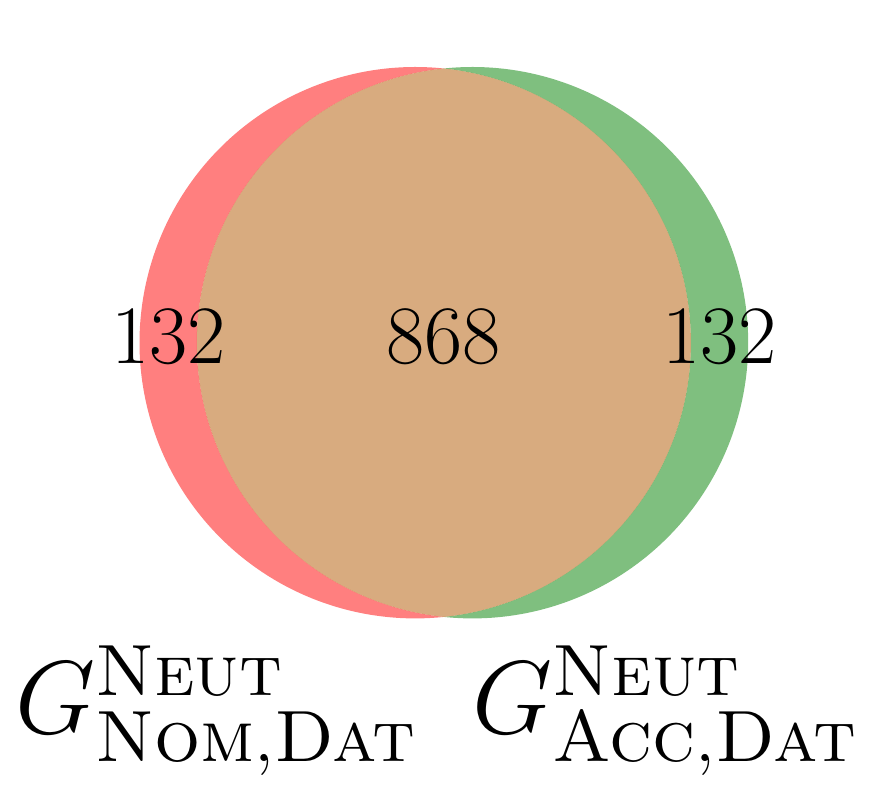}
    \includegraphics[width=0.14\linewidth]{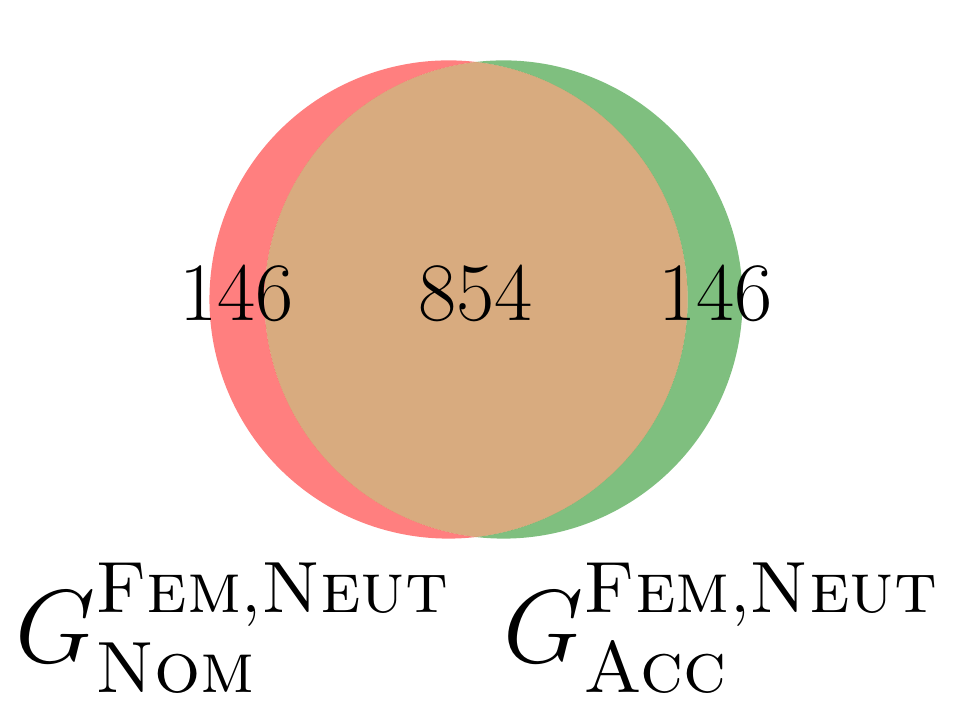}
    \includegraphics[width=0.14\linewidth]{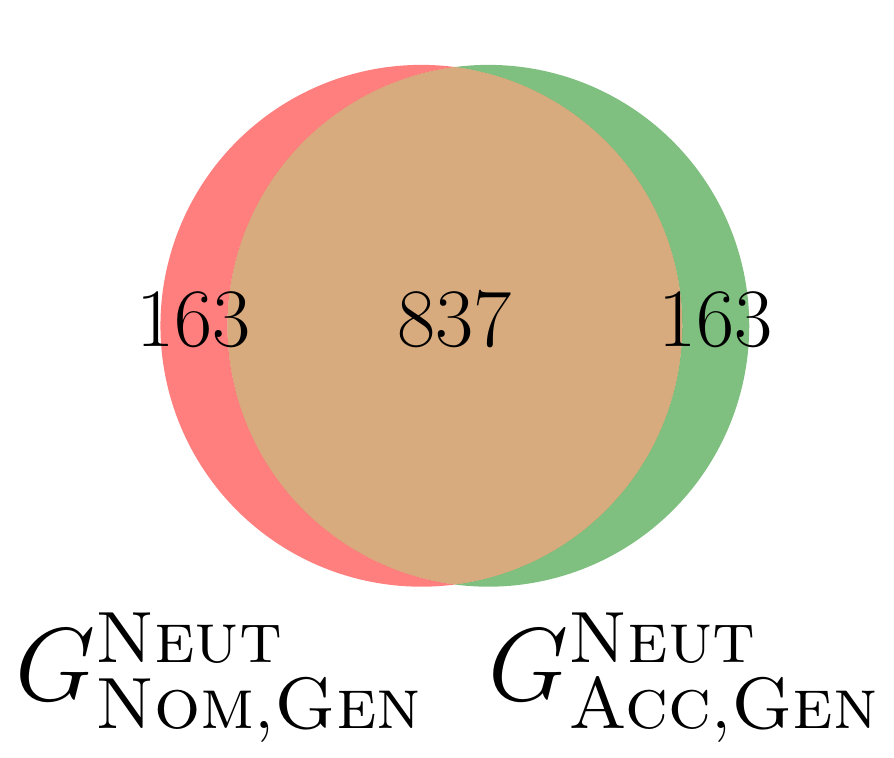}
    \revbg{
  \includegraphics[width=0.14\linewidth]{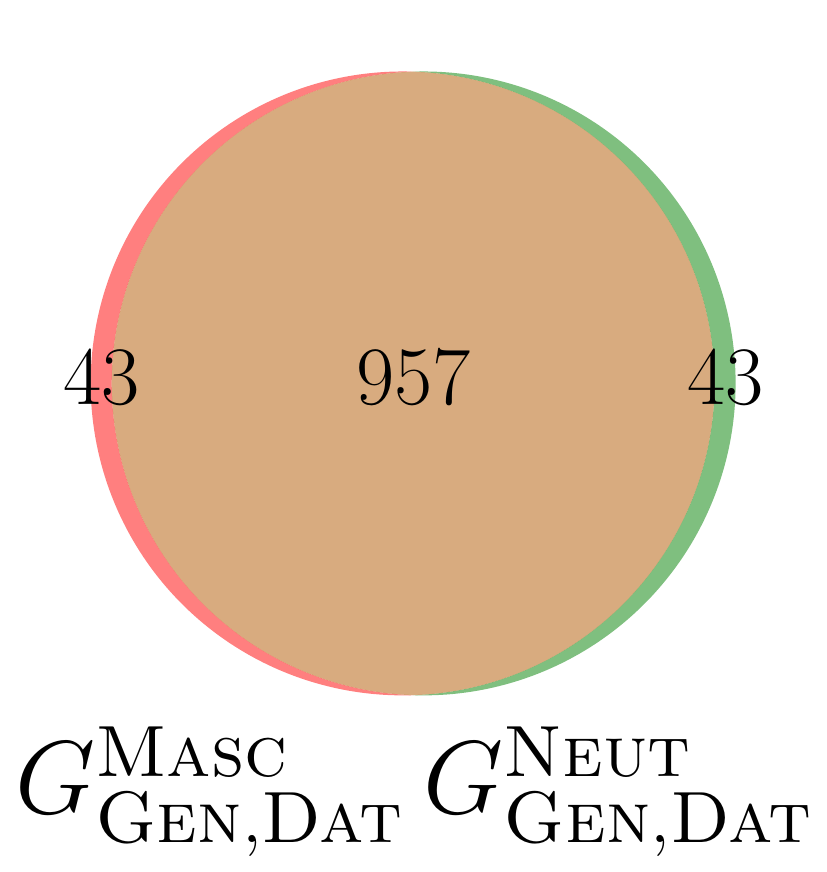}
}
    \caption{\gpttwo.}
\end{subfigure}

 \begin{subfigure}[t]{\linewidth}
    \includegraphics[width=0.18\linewidth]{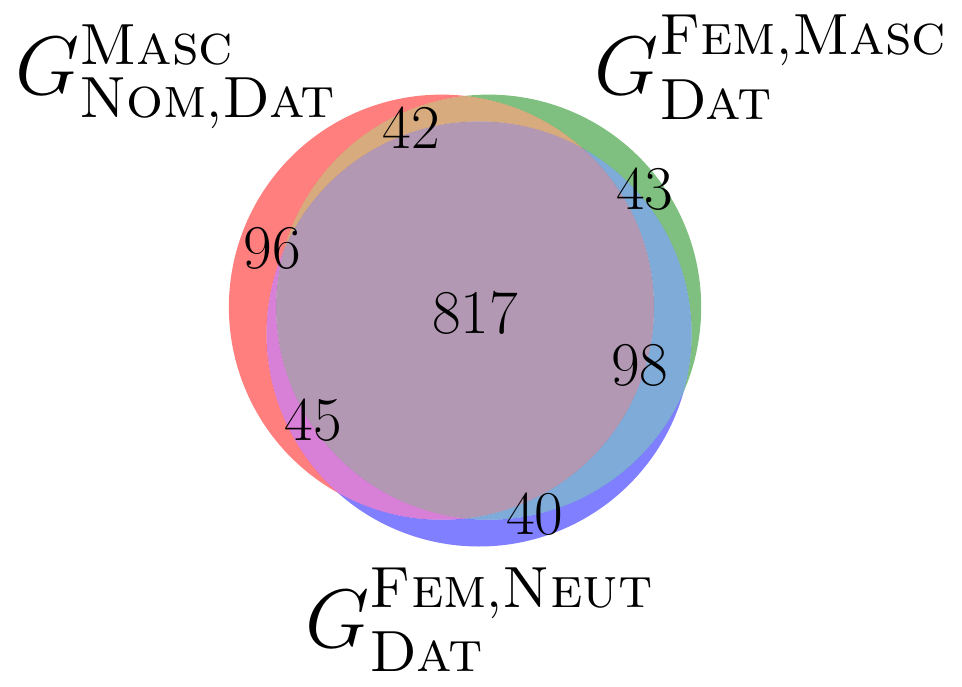}
    \includegraphics[width=0.18\linewidth]{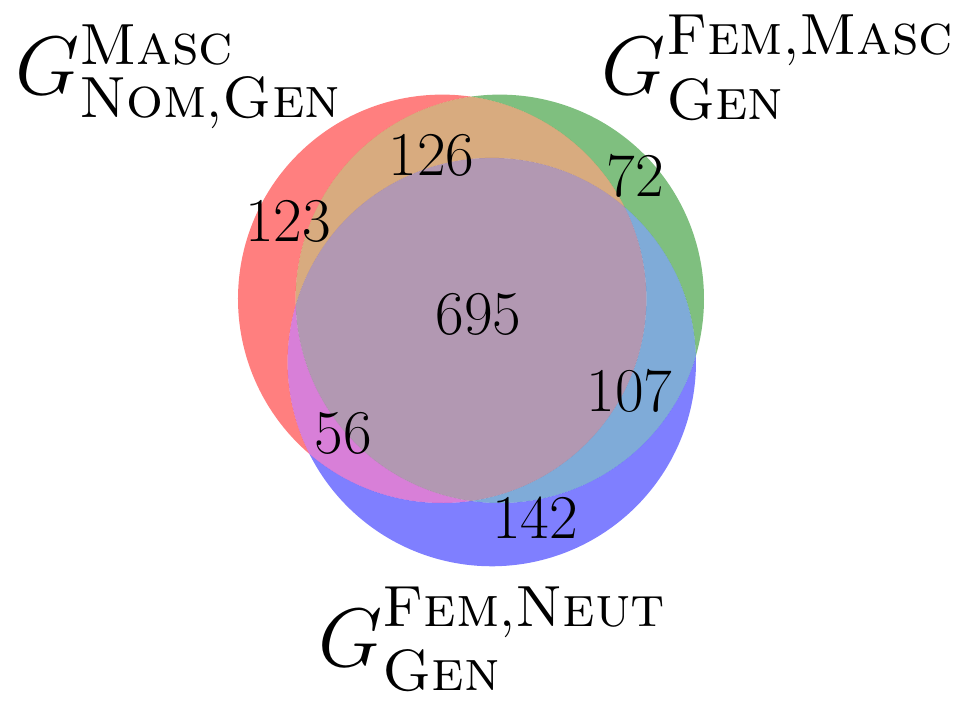}
    \includegraphics[width=0.14\linewidth]{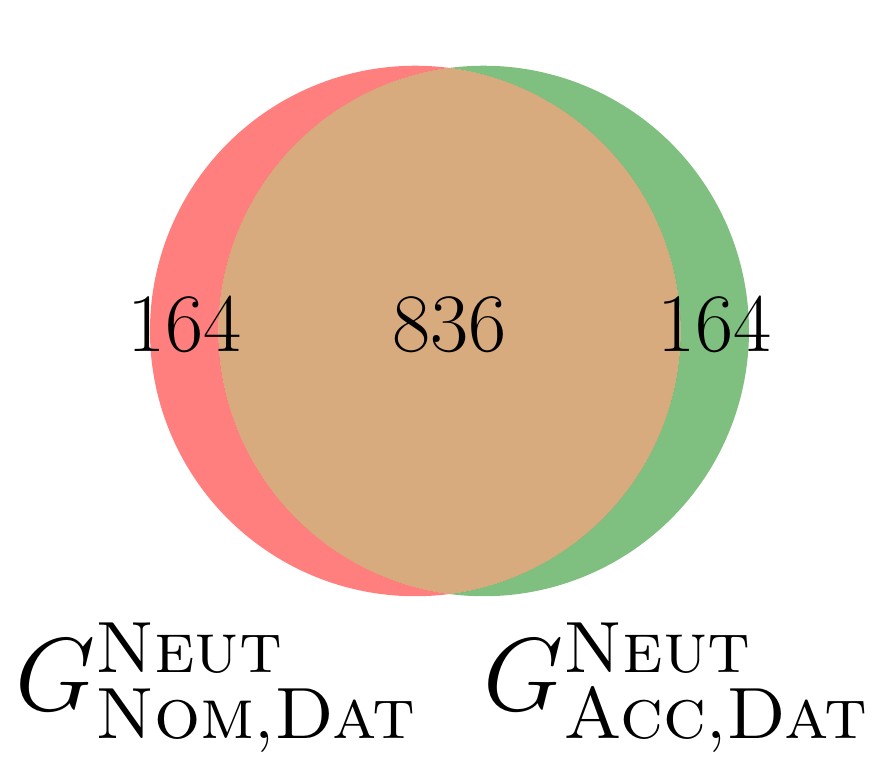}
    \includegraphics[width=0.14\linewidth]{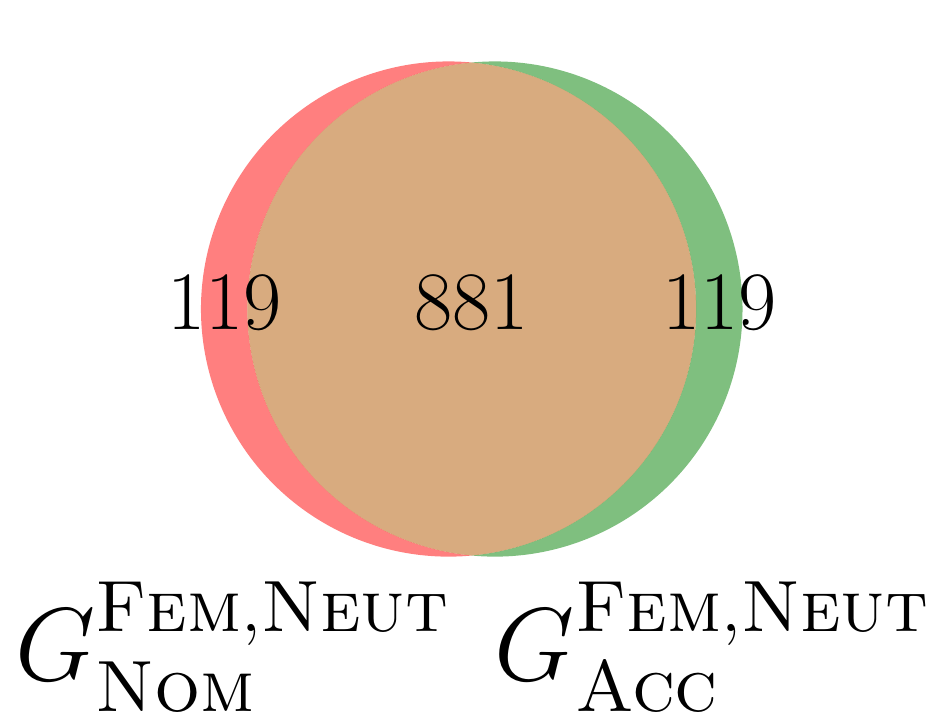}
    \includegraphics[width=0.14\linewidth]{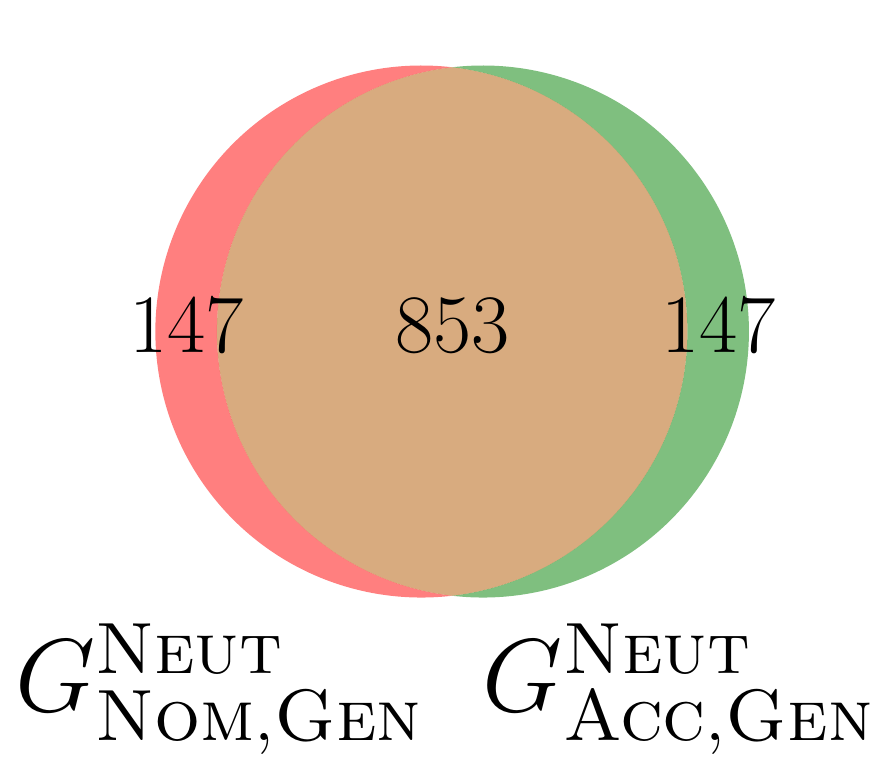}
    \revbg{
  \includegraphics[width=0.14\linewidth]{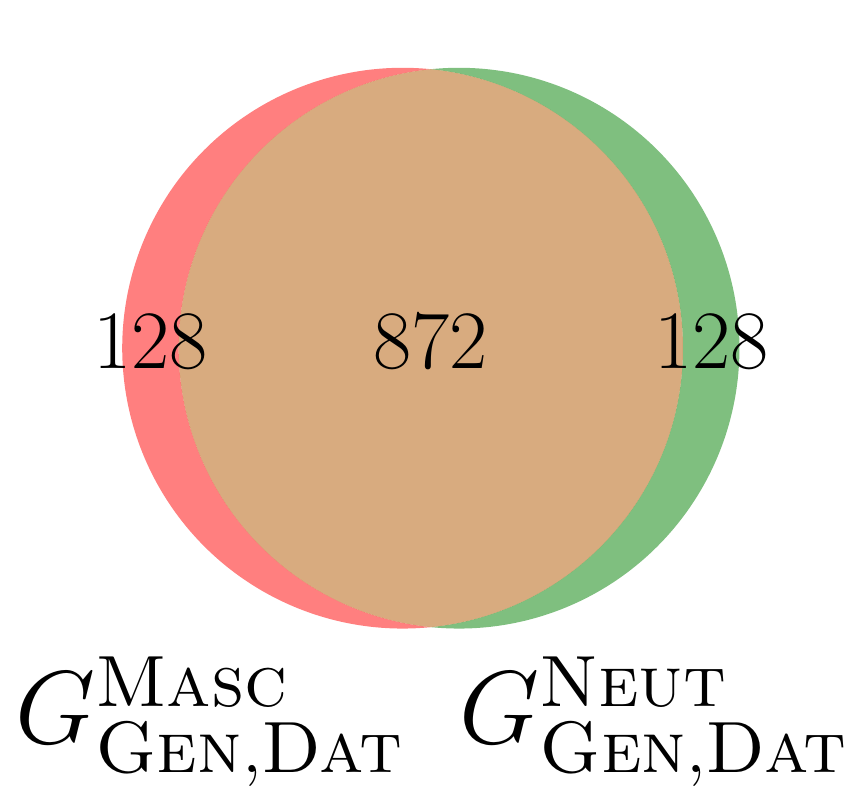}
}
    \caption{\llama.}
\end{subfigure}
    
    \caption{Top-$1,000$ weight overlaps across different \glspl{gradiend} for non-\bert-models.}
    \label{fig:venn-other-models-weight}
\end{figure*}
\begin{figure*}
    \centering
    \includegraphics[width=\linewidth]{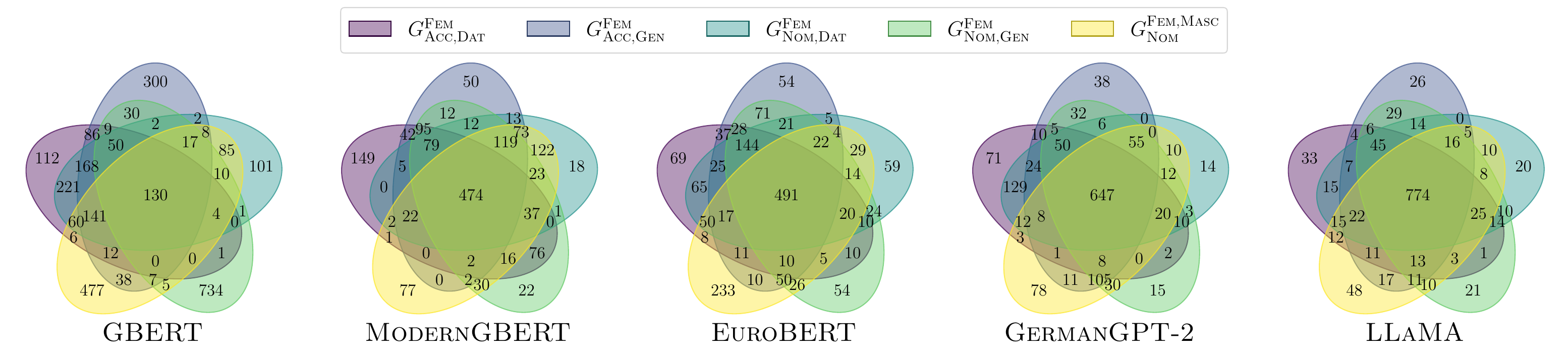}

    \caption{Top-$1,000$ weight overlaps for non-\bert\ models of article group $der\!\leftrightarrow\!die$.}
    \label{fig:venn-der-die-other-models-weight}
\end{figure*}
\begin{figure*}
    \centering
    \includegraphics[width=\linewidth]{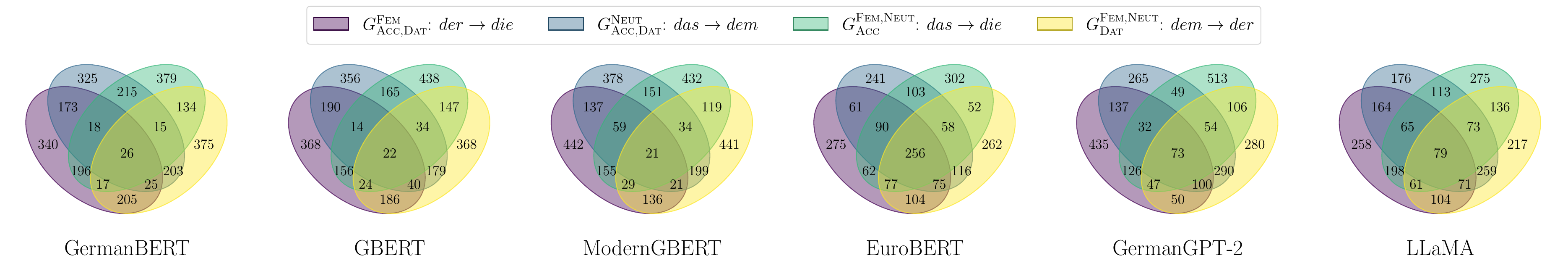}
    \caption{Top-$1,000$ weight overlaps for non-\bert\ models of the control group.}
    \label{fig:venn-control-other-models-weight}
\end{figure*}

Figures~\ref{fig:venn-other-models-weight}--\ref{fig:venn-control-other-models-weight} show the Venn diagrams of non-\bert-models similar to Figures~\ref{fig:venn-bert-weights}--\ref{fig:venn-control-group-bert-weight}.

Our overlap analysis depends on the choice of $k$.
Choosing $k$ too small makes overlaps sensitive to ranking noise and a few extreme weights, whereas choosing $k$ too large gradually includes many weakly-informative weights, making overlap less diagnostic.

\begin{figure*}[!t]
    \centering
    \includegraphics[width=\linewidth]{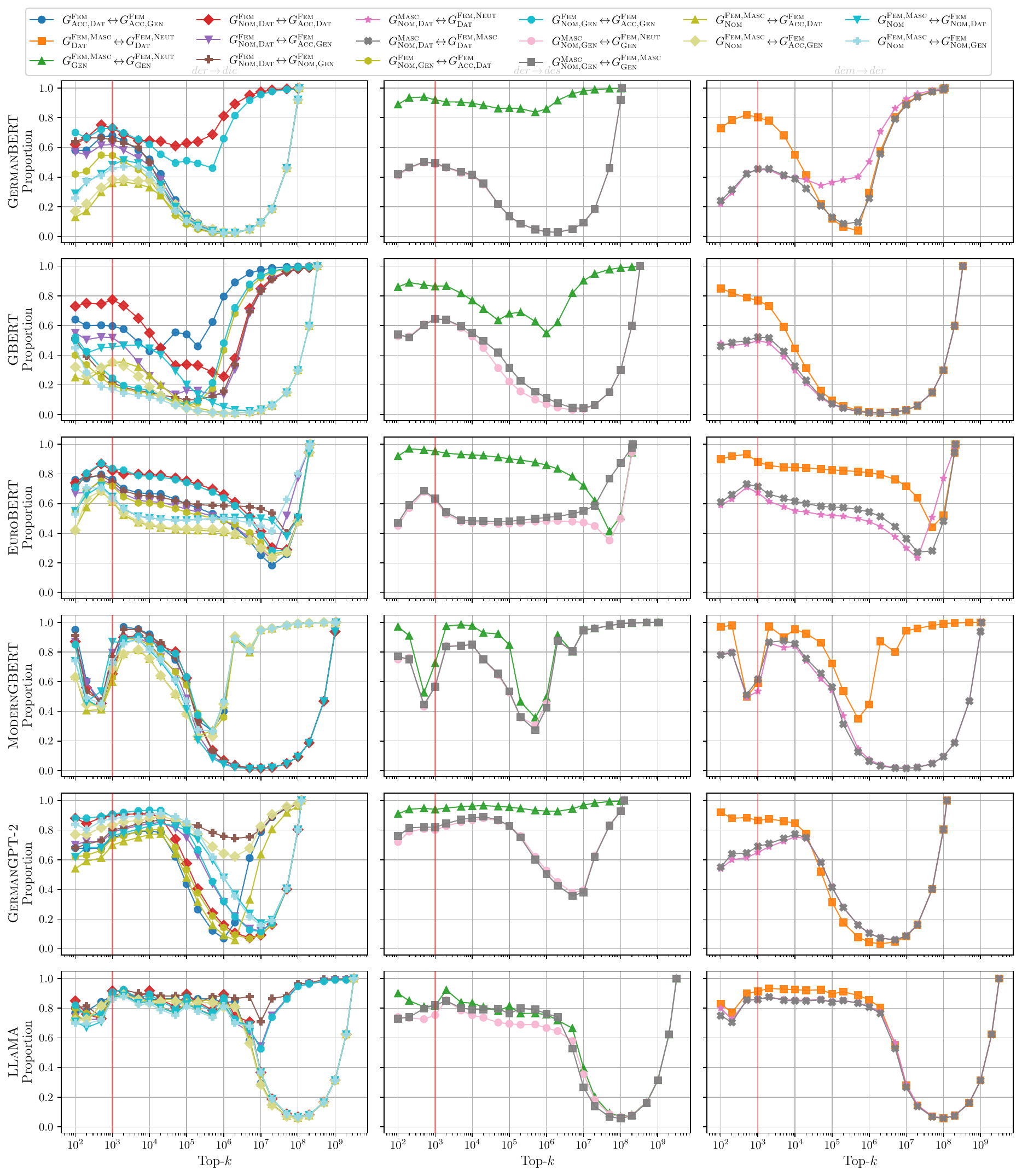}
\caption{Pairwise Top-$k$ weight-intersection proportions across values of $k$ for the control group for the two-dimensional article groups.}
    \label{fig:top-k-ablation-grid}
\end{figure*}

\begin{figure}[!t]
    \centering
    \includegraphics[width=\linewidth]{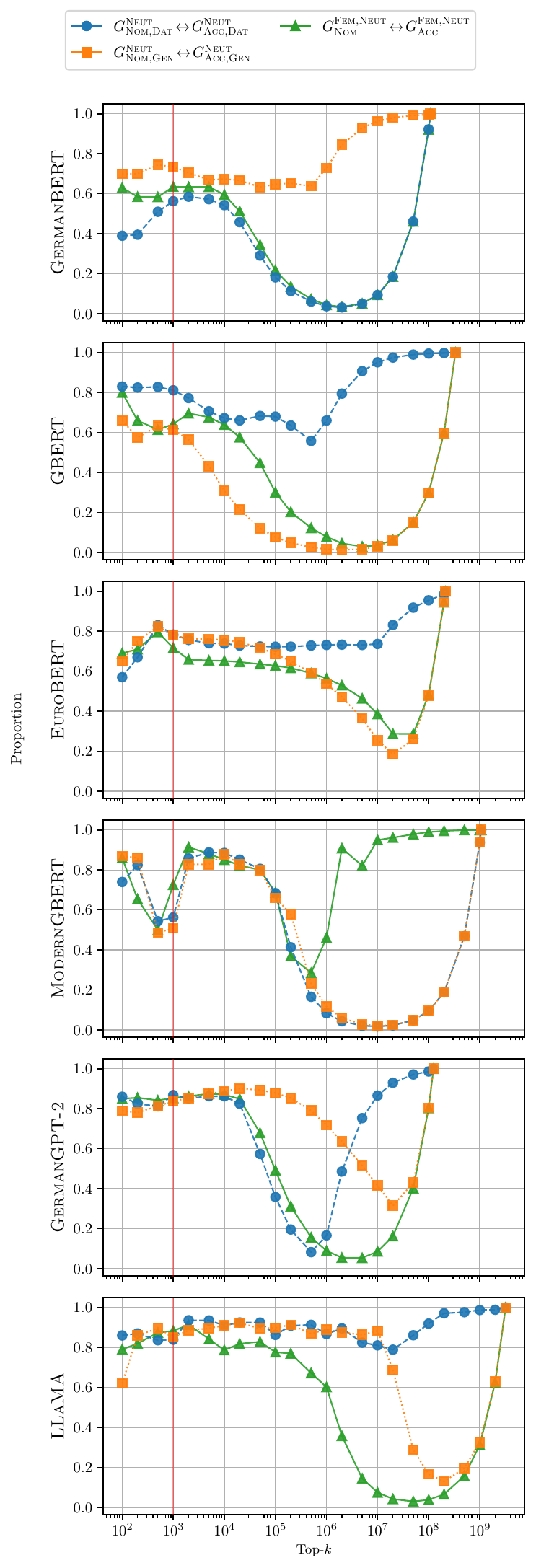}
\caption{Pairwise Top-$k$ weight-intersection proportions across values of $k$ for the one-dimensional article groups.}
    \label{fig:top-k-ablation-per_model}
\end{figure}

\begin{figure}[!t]
    \centering
    \includegraphics[width=\linewidth]{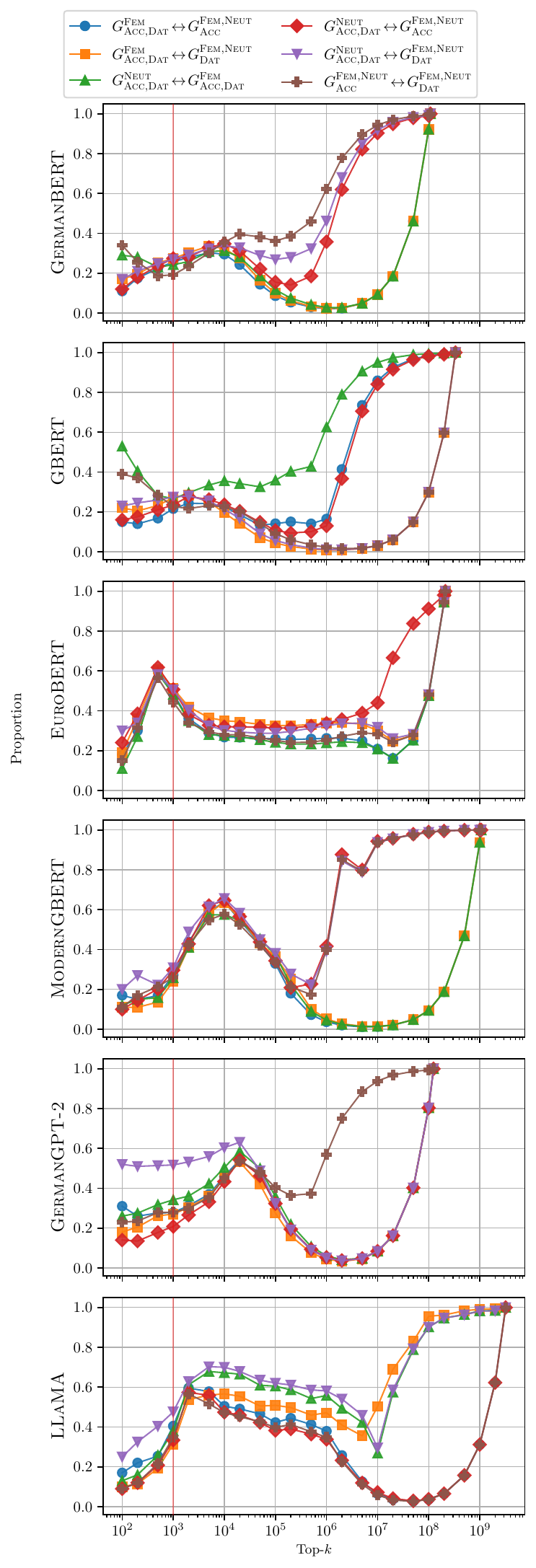}
\caption{Pairwise Top-$k$ weight-intersection proportions across values of $k$ for the control group.}
    \label{fig:top-k-ablation-control}
\end{figure}

To motivate our choice, we ablate $k$ and plot pairwise Top-$k$ overlap within each article group as a function of $k$ (Figures~\ref{fig:top-k-ablation-grid}--\ref{fig:top-k-ablation-control}).
Across model/group combinations, the curves vary in detail, but many exhibit a recurring three-range pattern:
\begin{enumerate}[label=(\roman*)]
    \item \textbf{Transition-dominated range (small/intermediate $k$):} overlap is comparatively high and relatively stable, indicating that the Top-$k$ sets are dominated by weights that are consistently ranked highly across variants within the same article group.
    \item \textbf{Reduction range (intermediate/large $k$):} overlap often decreases as increasing $k$ begins to include additional weights whose ranks are less consistent across variants, reducing the intersection proportion.
    \item \textbf{Trivial convergence range (very large $k$):} overlap increases again as selectivity vanishes; in the limit, overlap approaches $100\%$ when $k$ becomes large relative to the parameter count.
\end{enumerate}
While this trend is not uniform for all models and variants, $k{=}1000$ yields a stable and comparable operating point across article groups, focusing on a small but informative subset of weights. 
We choose $k{=}1000$ because it lies in the first, transition-dominated regime for most model-group combinations in Figures~\ref{fig:top-k-ablation-grid} and \ref{fig:top-k-ablation-per_model}, before the dilution-driven decrease becomes prominent.
Importantly, the intersection proportions of the article groups (Figures~\ref{fig:top-k-ablation-grid} and \ref{fig:top-k-ablation-per_model}) are consistently above the control group proportions (Figure~\ref{fig:top-k-ablation-control}) for  $k\le10,000$, indicating a generalization of our main claim based on Table~\ref{tab:overlap} across a wide range of small $k$.




\end{document}